\documentclass[11pt,bezier,amstex]{article}

\usepackage[in]{fullpage}
\usepackage{authblk}
\usepackage[utf8]{inputenc} 
\usepackage[T1]{fontenc}    
\usepackage{times}  
\usepackage{helvet} 
\usepackage{courier}  
\usepackage[hyphens]{url}  
\usepackage{graphicx} 
\usepackage{subfigure}
\usepackage{thmtools, thm-restate}
\usepackage{tabularx}
\usepackage{hhline}
\usepackage{makecell}
\usepackage{multirow}
\usepackage{multicol}
\usepackage{textcomp}

\usepackage[switch]{lineno}  %
\usepackage[hidelinks,colorlinks=true,linkcolor=blue,citecolor=blue]{hyperref}
\usepackage{cancel}
\usepackage{microtype}      

\usepackage{amsmath}
\usepackage{amssymb,amsopn,algorithm,algorithmic,theorem,float,bbm,bm,enumerate,color,multirow,gensymb}

\usepackage{natbib}

\allowdisplaybreaks[4]
\usepackage[dvipsnames]{xcolor}

\newcommand{\beq}{\begin{equation}}
\newcommand{\eeq}{\end{equation}}

\newcommand{\g}{\mathbf{g}}

\newcommand{\cN}{{\cal N}}

\newcommand{\vertiii}[1]{{\left\vert\kern-0.25ex\left\vert\kern-0.25ex\left\vert #1
    \right\vert\kern-0.25ex\right\vert\kern-0.25ex\right\vert}}

\newcommand {\commentout}[1] {}

\def\ints{{{\rm Z} \kern -.35em {\rm Z} }}  
\def\smallints{{{\rm Z} \kern -.3em {\rm Z} }}  
\def\pints{{{\rm I} \kern -.15em {\rm N} }}      
\newcommand{\reals}{\mathbb R}

\def\cplx{{{\rm I} \kern -.45em {\rm C} }}       
\def\l2{\rm {\mathcal L}^{2}(\reals)}            

\newcommand{\be}{\begin{eqnarray}}
\newcommand{\ee}{\end{eqnarray}}
\newcommand{\bea}{\begin{eqnarray}}
\newcommand{\eea}{\end{eqnarray}}
\newcommand{\beaa}{\begin{eqnarray*}}
\newcommand{\eeaa}{\end{eqnarray*}}
\newcommand{\bnad}{\begin{nad}}
\newcommand{\enad}{\end{nad}}

\title{Experiments with Rich Regime Training for Deep Learning}
\date{}
\author[1]{Xinyan~Li}
\author[2]{Arindam~Banerjee}

\affil[1]{Department of Computer Science \& Engineering, 
University of Minnesota, Twin Cities}
\affil[2]{Department of Computer Science,
University of Illinois Urbana-Champaign}
\affil[ ]{Emails: \texttt{\{lixx1166@umn.edu, arindamb@illinois.edu\}}}

\begin{document}

\maketitle

\begin{abstract}

In spite of advances in understanding lazy training, recent work attributes the practical success of deep learning to the rich regime with complex inductive bias. In this paper, we study rich regime training empirically with benchmark datasets, and find that while most parameters are lazy, there is always a small number of active parameters which change quite a bit during training. We show that re-initializing (resetting to their initial random values) the active parameters leads to worse generalization. Further, we show that most of the active parameters are in the bottom layers, close to the input, especially as the networks become wider. Based on such observations, we study static Layer-Wise Sparse (LWS) SGD, which only updates some subsets of layers. We find that only updating the top and bottom layers have good generalization and, as expected, only updating the top layers yields a fast algorithm. Inspired by this, we investigate probabilistic LWS-SGD, which mostly updates the top layers and occasionally updates the full network. We show that probabilistic LWS-SGD matches the generalization performance of vanilla SGD and the back-propagation time can be 2-5 times more efficient.

\end{abstract}

\section{Introduction}
Despite the remarkable success of deep networks in many domains, such as computer
vision \citep{Szegedy_2015_CVPR,he2016deep,krizhevsky2017imagenet}, speech recognition \citep{amodei2016deep,weninger2014discriminatively,oord2016wavenet}, and natural language processing \citep{sutskever2014sequence,wu2016google,vaswani2017attention}, their inductive bias and associated generalization performance is still not well understood. Recent years have seen considerable interest and advances on understanding the infinite width limit of the neural networks~\citep{jacot2018neural, arora2019exact,huang2020deep,lee2020wide}. In such infinite width limit, the learning dynamics of wide neural networks can be simplified by a linear model obtained from the first-order Taylor expansion around its initial parameters \citep{jacot2018neural,chizat2018note,lee2020wide}. Further, such a setting can be viewed as a kernel model, and
the training algorithms can be shown to have an implicit bias of picking the minimum norm solution corresponding to the Hilbert space that is closest to the initialization~\citep{arora2019finegrained,mei2019meanfield,chizat2019}. Such a setting is often referred to as the {\em lazy regime} (or {\em kernel regime}), and recent 
years have shown that a model can operate in the lazy regime based on suitably large scaling~\citep{chizat2018note,chizat2019,wogl2019}. In contrast, without such width or scaling going to infinity, deep models have been shown to have much richer inductive biases, which cannot be represented as a norm derived from a Hilbert space~\citep{gunasekar2017implicit,gunasekar2018conv,li2018learning,savarese2019infinite,wogl2019}. Such a setting is referred to as the {\em rich regime}, and key recent advances have been made on understanding how a model moves from the lazy regime to the rich regime~\citep{chizat2019,wogl2019,moroshko2020implicit}.

\begin{figure}[t]
    \centering
    \subfigure[VGG-5, MNIST ]{\includegraphics[width=0.45\columnwidth]{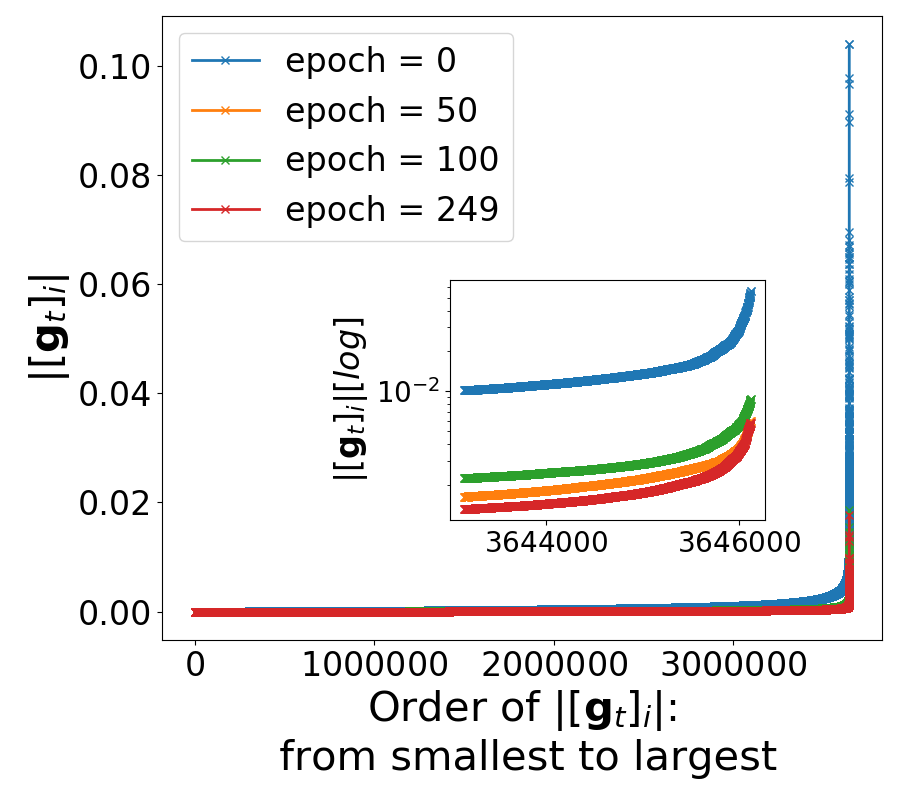}}
    \hspace{5mm}
    \subfigure[ResNet-18, CIFAR-10]{\includegraphics[width=0.45\columnwidth]{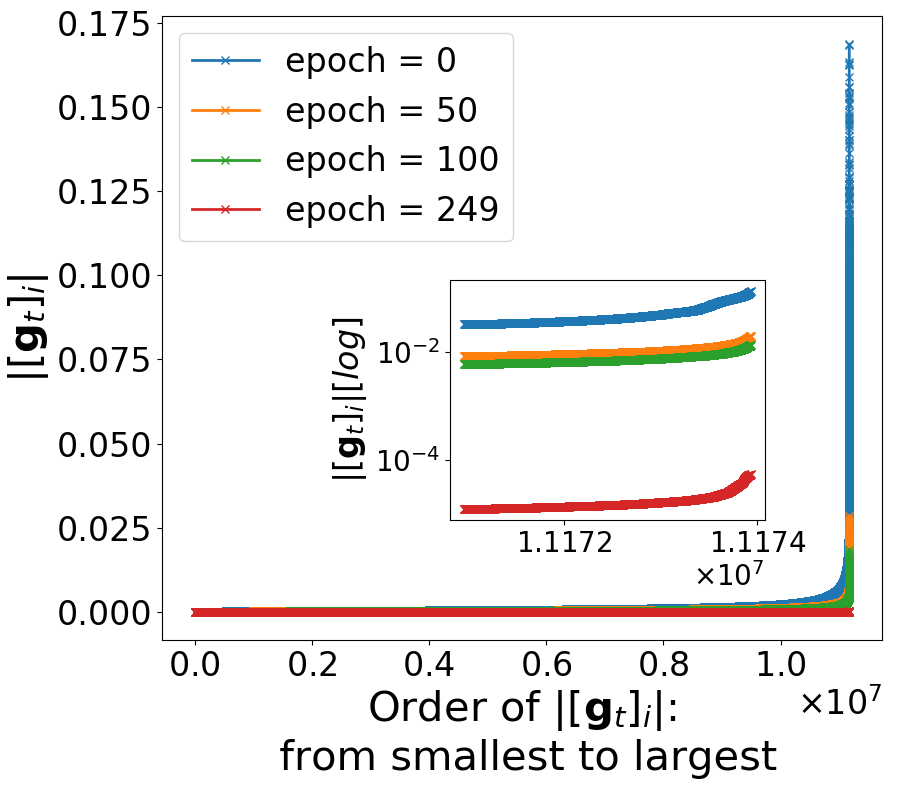}}
    \caption{Sorted absolute values of stochastic gradient components $|[\g_t]_i|$ for (a) VGG-5 ($p=3,646,154$) trained on MNIST and (b) ResNet-18 ($p=11,173,962$) trained on CIFAR-10. $\g_t$ is the $p$-dimensional stochastic gradient computed from mini-batch and $[\g_t]_i$ denotes the $i$-th component of $\g_t$ for $i \in [p]$. The inset plot provides a zoom-in view of what happened at ``elbow''. Druing training, the stochastic gradients are highly skewed such that most gradient components are near zero and only a small portion are significant. Such ``sparse'' structure of gradients can be observed across different over-parameterized deep networks and various datasets.}
     \label{fig:gt_decay}
\end{figure}

In this paper, we empirically study a variety of behaviors in the rich regime with focus on commonly used deep nets and benchmark datasets: MNIST and CIFAR-10. Our goal is to empirically complement the recent work on the rich regime~\citep{chizat2019,wogl2019} where the focus was on the theory and the experiments were on relatively simple models. 
While many factors, such as width, initialization (scaling), step size, batch size, and depth, can potentially help the transition into the rich regime, we primarily focus on varying the width and to some extent the depth to do a detailed empirical study of generalization and optimization in the rich regime. Our empirical results show what happens in typical deep nets used in practice in the rich regime, and will hopefully provide fodder for new theoretical advances on understanding inductive bias.

Our first set of experiments (Section~\ref{sec:generalization}) sorts individual parameters after training from active to lazy, for networks with different widths, and studies the effect of re-initializing the most active vs.~the most lazy sets of parameters to their original random values. How active a parameter is measured by the absolute distance from initialization to convergence, to keep things simple. The results show that re-initializing even 0.1\% of the most active parameters leads to significant drop in generalization performance, and re-initializing 1-10\% of the most active parameters leads to the most significant drop in performance. Oddly, re-initializing even more of the active parameters, say 50\%, leads to an improvement in generalization performance over the 1-10\% performance for wider networks. In the limit of 100\% re-initialization, i.e., random networks, wider networks have better generalization, but it does not reach the generalization performance of the trained networks. In fact, even a width 10 trained network outperforms a width 100,000 random network.

Our second set of experiments (Section~\ref{sec:dynamics}) investigates how the active parameters are spread over the layers. Existing work~\citep{gunasekar2018conv,wogl2019,gunasekar2017implicit} has shown that the model should have an implicit bias towards sparsity. Recent empirical observation also suggests that, during training, the stochastic gradients of over-parameterized deep networks decay very fast and only a small portion is significant \citep{gurari2018gradient,zhou2020bypassing} (see Figure \ref{fig:gt_decay}). However, it is unclear if the active parameters will be spread across the entire network or be concentrated in certain layers. Our experiments show that the active parameters for Conv-Nets are primary in the bottom layers, closest to the input, and the concentration of the active parameters in the bottom layers increase with increase in width. For ReLU-Nets, the behavior in the bottom layer is similar, but the top layer also has some of the active parameters. In all case, the middle layers hardly have any active parameters, especially for wide networks.

The observations above on the spread of active parameters across layers lead to the natural question: what happens if we train the network with a layer-wise sparse (LWS) SGD algorithm which updates only subsets of layers? Of specific interest, based on our observations, is a LWS-SGD algorithm which only update the top and bottom layers. In Section~\ref{sec:layer_sparse}, we consider LWS-SGD applied to different subsets of layers, and compare their generalization performance and running time. We find that just training the very top and very bottom layer works as well as training the full network in terms of generalization, and such training is indeed quite sparse layer-wise for models such as VGG-5 and VGG-11, which we use for our experiments. However, training just the top and bottom layers is only mildly faster than training the full network since the backward pass of back-propagation has to all the way down to the bottom. Training just the top (few) layer(s) is much faster, but its generalization performance is not as good. Based on this, we finally explore a few probabilistic LWS-SGD algorithms including one which trains the top layer in most epochs (with probability $(1-\rho)$) and occasionally trains the full network (with probability $(\rho)$). Such an algorithm matches up the performance of training the full network in every epoch, especially when initialized with pre-training, and the back-propagation phase of such an algorithm is 2-5 times faster than training the full network.

{\bf Paper organization.} We briefly review related work in Section \ref{sec:related}.  We discuss our experimental setup and notation in Section \ref{sec:exp}. We study the effect of active parameters on generalization in Section \ref{sec:generalization}. Section \ref{sec:dynamics} studies the layer-wise distribution of active parameters with varying width. Variants of LWS-SGD are discussed in Section \ref{sec:layer_sparse}, and we conclude in Section \ref{sec:conclude}. We provide key experimental details and primary results in the main paper and defer results on 
additional network architectures, dataset, and layer combinations used in LWS-SGD to the supplementary material.

\label{sec:intro}

\section{Related Work}
{\bf Lazy (kernel) and rich regime.} Recent study \citep{du2018a,du2018b,zou2018stochastic,allenzhu2019convergence} has shown that over-parameterized neural networks with sufficient large width can converge to zero training loss at a linear convergence rate and their parameters stay close to the initialization during training. While such work usually requires the network width to be larger than a high degree polynomial of the training sample size $n$, the inverse of the target 
error $\epsilon$, and the inverse of the failure probability $\delta$, \citet{ji2020polylogarithmic,chen2020much} have shown that, with suitable assumptions on data separability, polylogarithmic width is sufficient for ReLU networks. 
Jacot et al. \citep{jacot2018neural} characterized the behavior of an infinitely wide fully-connected neural network trained with gradient descent by the Neural Tangent Kernel (NTK) which essentially relies on the linearization of the network around its initialization. The NTK has later been extended to convolutional neural networks \citep{arora2019exact,yang2019scaling}, residual neural networks \citep{huang2020deep}, and recurrent neural networks \citep{alemohammad2020recurrent}. 

\citet{chizat2019} have argued that such laziness during training is due to an implicit choice of the scale of the initialization. For example, at the infinite-width limit $m \to \infty$, the NTK \citep{jacot2018neural,arora2019exact} and other work on two-layer networks with random initialization \citep{du2018b,li2018learning} consider the scale of $1/\sqrt{m}$, whereas the study of the mean-field limit of neural networks with one hidden layer 
\citep{mei2018mean,rotskoff2018parameters,sirignano2020mean} leads to a choice of the scaling equals to $1/m$. Such initialization scaling can control the transition between the ``kernel regime'' and the ``rich regime'' \citep{wogl2019,moroshko2020implicit}. \citet{fort2020deep} have showed that, in practice, non-linear deep nets usually travel far enough from their initialization, violating the assumption made by NTK, thus are more likely to act in the rich regime. 



{\bf Implicit bias.} Existing work has shown that over-parameterized neural networks trained with gradient-based methods can reach zero training error with implicit bias towards some form of sparsity, including, but not limited to, the minimum $\ell_1$ norm solution \citep{wogl2019} in multi-layer homogeneous network 
the minimum nuclear norm solution \citep{gunasekar2017implicit} in over-parametrized matrix factorization which guarantees low rank matrix recovery [\citep{li2018regularization}, sparsity in the frequency domain \citep{gunasekar2018conv}, the low-rank solution in deep matrix factorization \citep{arora2019implicit,razin2020implicit}, and the minimum variation norm solution \citep{chizat2020implicit} for wide two-layer networks trained with logistic loss.




{\bf Which layer(s) helps or hurts the training.} \citet{Lan2019} studied which parameters are “helpful” or “hurtful” using the Loss Change Allocation (LCA) score which measures the per-parameter, per-iteration changes to the overall loss. They found the first and last layers consistently hurt training due to a positive total LCA. 
\citet{raghu2017} found the layers in neural networks converging in a bottom-up manner and proposed ``Freeze Training'' to sequentially freeze lower layers after a certain number of epochs. \citet{zhby2019} also studied the behavior of individual layers through the concept of re-initialization and re-randomization robustness. They showed the layers in a deep network are not homogeneous 
and found the bottom layers are sensitive while the top layers are robust to re-initialization. We enrich the concept of re-initialization in \citep{zhby2019} to study the behaviors of deep nets trained in the rich regime. 


{\bf Stochastic depth.} Our work on LWS-SGD also broadly relates to a line of work specifically targets the residual neural network architecture, such as stochastic depth \citep{huang2016} and probabilistic gates \citep{herrmann2018deep}, which automatically discard layers by bypassing them with the identity function and reduce training time substantially. Unlike stochastic depth which bypassing a subset of layers, our LWS-SGD uses the initial (random) values of frozen layers and treats them as random projections.

\label{sec:related}

\section{Experimental Setup}
\label{sec:exp}
We consider fully-connected neural networks with ReLU activation (ReLU-Net),  shallow convolutional neural networks (Conv-Net), and the family of Visual Geometry Group (VGG) networks \citep{vgg} which are also convolutional neural networks. Experiments are conducted on two benchmark datasets: a subset of MNIST \citep{mnist} ($n=10,000$) where equal number of samples has been selected from each class and CIFAR-10 \citep{cifar10}. In particular, we train VGG-5 on MNIST and VGG-11 on CIFAR-10. Since the study purely focuses on the convolutional layer and the fully-connected layer, we exclude the batch normalization and the dropout layers existed in VGG-5 and VGG-11. All models are initialized with Xavier initialization \citep{xavier}, such that the parameters between layer $l_i$ and $l_{i+1}$ are sampled from a Gaussian distribution $\cN(0,\sigma^2\mathbb{I})$ with $\sigma^2 \sim 1/(|l_i|+|l_{i+1}|)$ which is inverse proportional to width. 

For experiments conducted on MNIST, we train our model using Adam with constant learning rate of 0.1. For those models trained on CIFAR-10, we use SGD accelerated with Nesterov momentum equals to 0.9 \citep{sutskever2013importance} and employ a weight decay of 0.0005. The initial learning rate is set to be 0.01 and reduced by half at every 30 epochs. For both cases, we use mini-batch size of 128 and let the optimizer run for a fixed number of epochs (100 epochs for MNIST and 180 epochs for CIFAR-10) to minimize the cross-entropy loss. All experiments have been run on NVidia Tesla K40m GPUs, and been repeated 5 times.

{\bf Notations.} Considering a neural network with width $w$ and depth $d$, we use $\theta \in \mathbb{R}^p$ to denote the model parameters. For ReLU-Nets, $w$ represents the number of hidden units at each layer, as for Conv-Nets, $w$ is the number of channels. A network with depth $d$ has in total $(d+1)$ layers with $d$ of them are hidden layers. We use $l_i$ to denote the $i^{th}$ layer of a network, with $l_0$ represents the input layer and $l_{d+1}$ represents the output. We call layers closed to the input as bottom layers and those closed to the output as top layers.

\section{Active Parameters and Generalization}
Let $\theta^T$ be the parameters of a neural network at the final epoch $T$, and $\theta^0$ be the parameters at initialization. To keep things simple, we compute the absolute distance each parameter $\theta_i$ moves from initialization to convergence $|\theta_i^T-\theta_i^0|$, and determine the active subspace by considering the $\varepsilon \%$ of the largest movements. Let $\tilde{\theta}^T=[\theta^T_{(1)},\dots,\theta^T_{(p)}]$ be the ordered parameters of the network sorted by $|\theta^T-\theta^0|$ in descending order, we examine the following two post-training re-initialization scenarios:

{\bf $\gamma$ Active-re-initialization.} Let $k_{\gamma}$ denotes the number of active coordinates corresponding to the first $\gamma\%$ of $\tilde{\theta}^T$. After training completes, we reset $[\theta^{T}_{(1)},\dots,\theta^{T}_{(k_{\gamma})}]$ to their initial values $[\theta^{0}_{(1)},\dots,\theta^{0}_{(k_{\gamma})}]$ and use $\tilde{\theta}^{T}_{\gamma}=[\theta^{0}_{(1)},\dots,\theta^{0}_{(k_{\gamma)}},\theta^{T}_{({k_{\gamma}}+1)},\dots,\theta^T_{(p)}]$ to denote the parameters after $\gamma$ Active-re-initialization.

{\bf $\varepsilon$ Lazy-re-initialization.} Let $k_{\varepsilon}$ be the number of active coordinates that corresponds to the first $\varepsilon\%$ of $\tilde{\theta}^T$. As training ends, the top $k_{\varepsilon}$ parameters stay intact and the remaining $(p-k_{\varepsilon})$ parameters $[\theta^{T}_{(k_{\varepsilon}+1)},\dots,\theta^{T}_{(p)}]$ are assigned to their initial values $[\theta^{0}_{({k_{\varepsilon}}+1)},\dots,\theta^{0}_{(p)}]$. We use $\tilde{\theta}^{T}_{\varepsilon}=[\theta^{T}_{(1)},\dots,\theta^{T}_{(k_{\varepsilon})},\theta^{0}_{({k_{\varepsilon}}+1)},\dots,\theta^{0}_{(p)}]$ to denote the parameters after $\varepsilon$ Lazy-re-initialization.

\begin{figure}[t]
    \centering
    \subfigure[Relu Network, d=1 (2 layers), MNIST, X-axis in log scale.]{
    \includegraphics[width=0.88\textwidth]{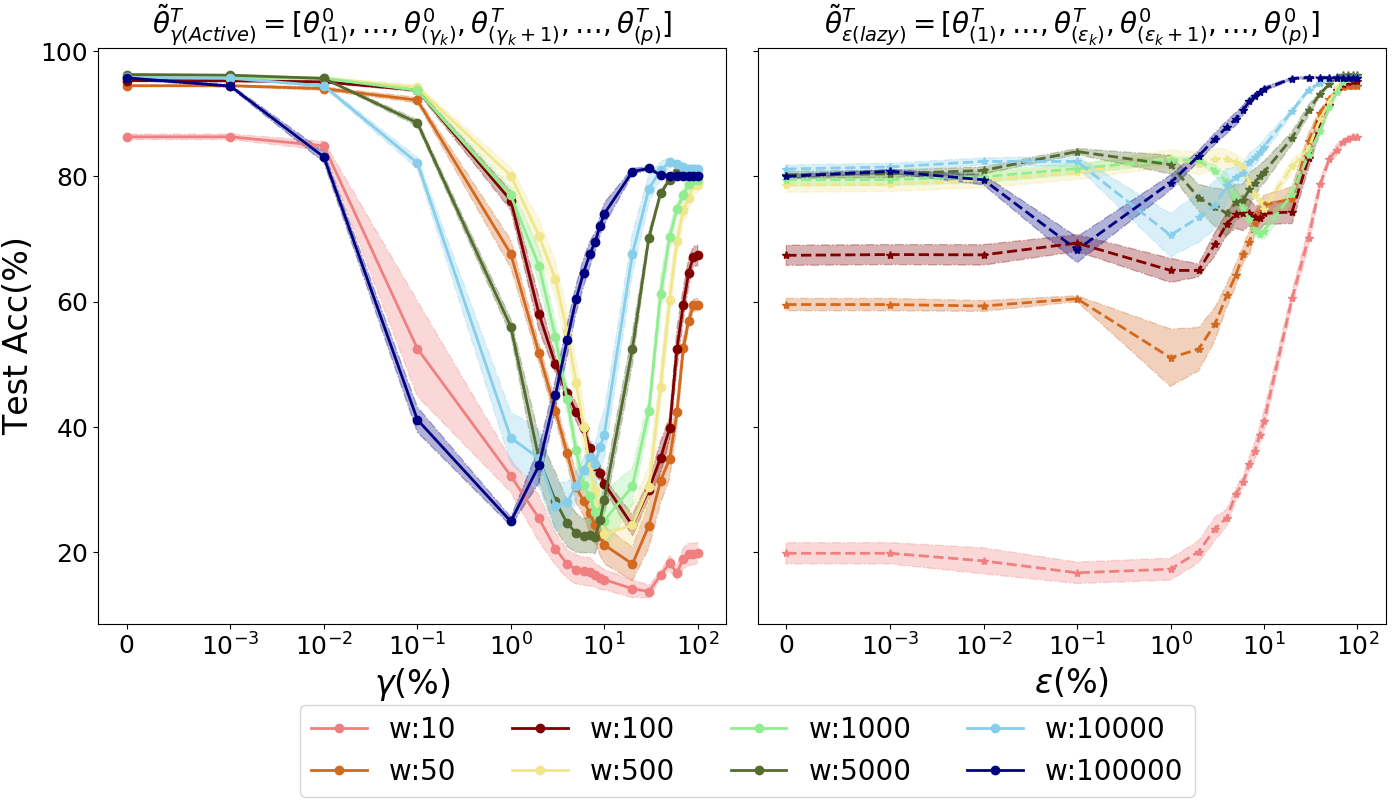}}
    \subfigure[Relu Network, d=1 (2 layers), MNIST, X-axis in linear scale.]{
    \includegraphics[width=0.88\textwidth]{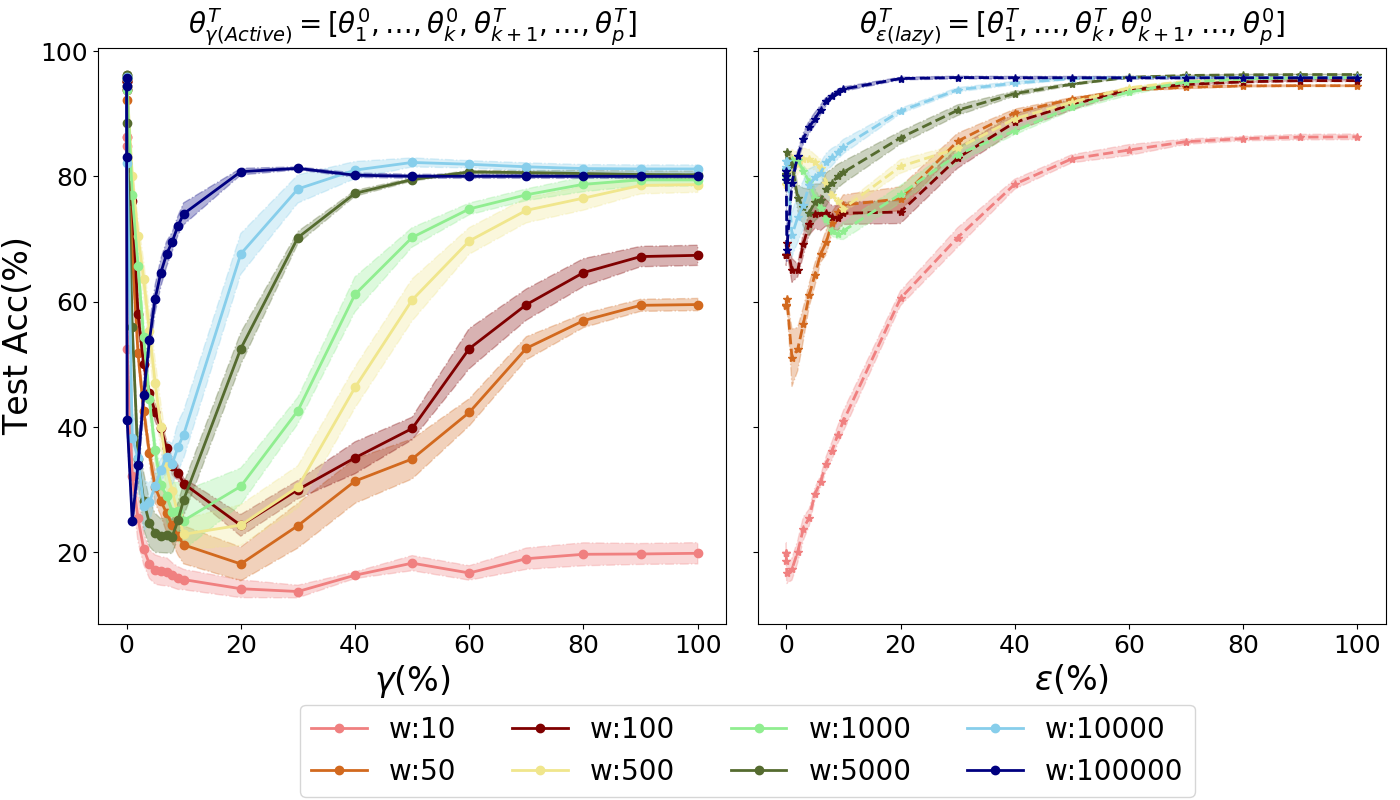}}    
    \caption{Test accuracy of 2-Layer ReLU-Nets ($d=1$) as a function of the choice of $\gamma$ and $\varepsilon$ after applying $\gamma$ Active-re-initialization (left) and $\varepsilon$ Lazy-re-initialization (right). (a) The x-axis is in log scale, and (b) the x-axis is in linear scale. As $\gamma$ increases, the test performance plunges at first then increases. The worse performance usually happens when around $10\%$ of the active parameters are re-initialized, indicating those active parameters may be of greater importance. As $\varepsilon$ increases, the test performance gradually improves. With $w=100,000$, keeping approximately $10\%$ of parameters at their trained value is sufficient to recover the performance of the fully trained network($\varepsilon= 100\%$).
    }
    \label{fig:relu_mnist_reinit_d1}
    \vspace{-5mm}
\end{figure}

Figure \ref{fig:relu_mnist_reinit_d1} show the generalization performance of ReLU-Nets as a function of the choice of $\gamma$ and $\varepsilon$ after applying $\gamma$ Active-re-initialization (left) and $\varepsilon$ Lazy-re-initialization (right). Different lines represent different widths. We increase the network width from $10$ to an extremely large value, e.g., $100,000$ for $d=1$ (2 layers) and $5,000$ for $d=4$ (5 layers), to approximate the network with infinite width.

\begin{figure}[t]
    \centering
    \subfigure[Relu Network, d=4 (5 layers), MNIST, X-axis in log scale.]{
    \includegraphics[width=0.88\textwidth]{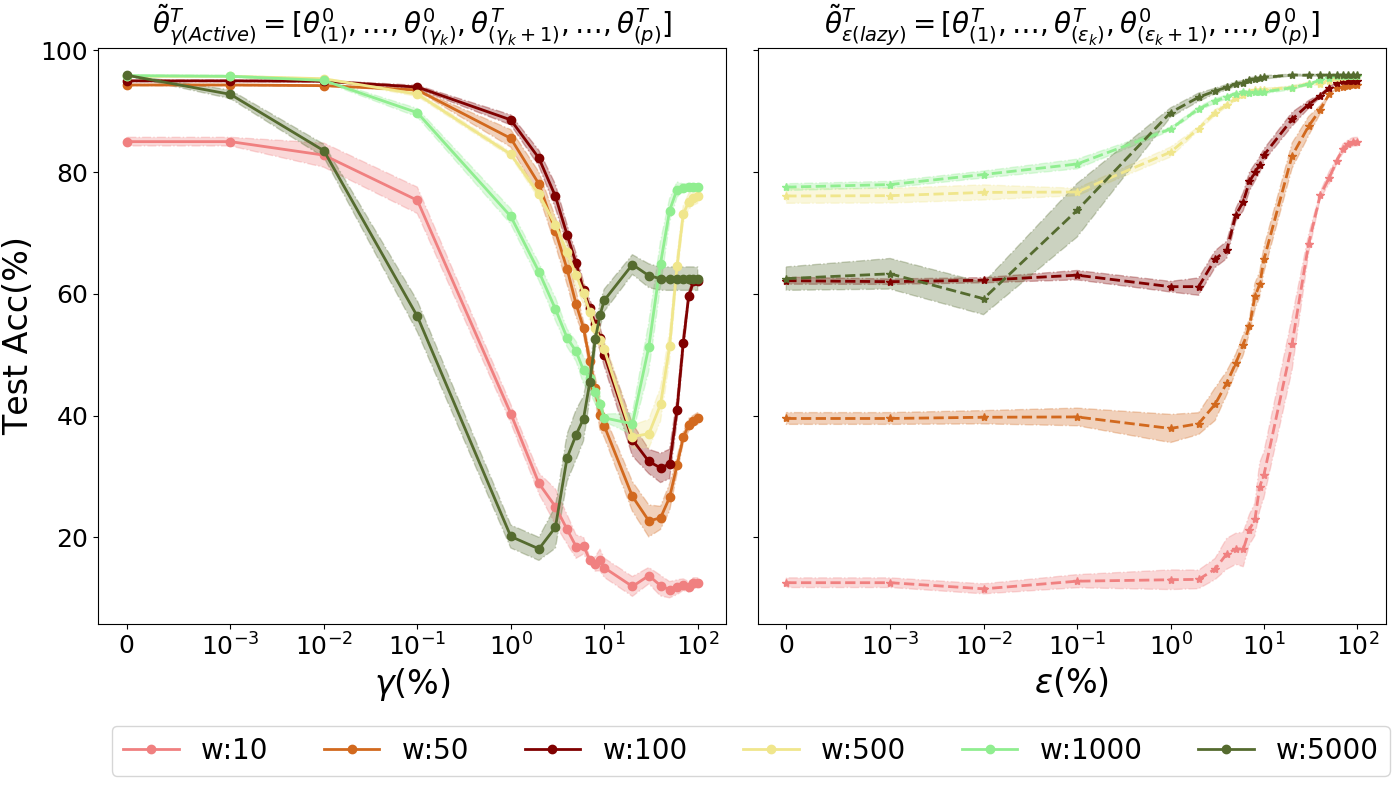}}
    \subfigure[Relu Network, d=4 (5 layers), MNIST, X-axis in linear scale.]{
    \includegraphics[width=0.88\textwidth]{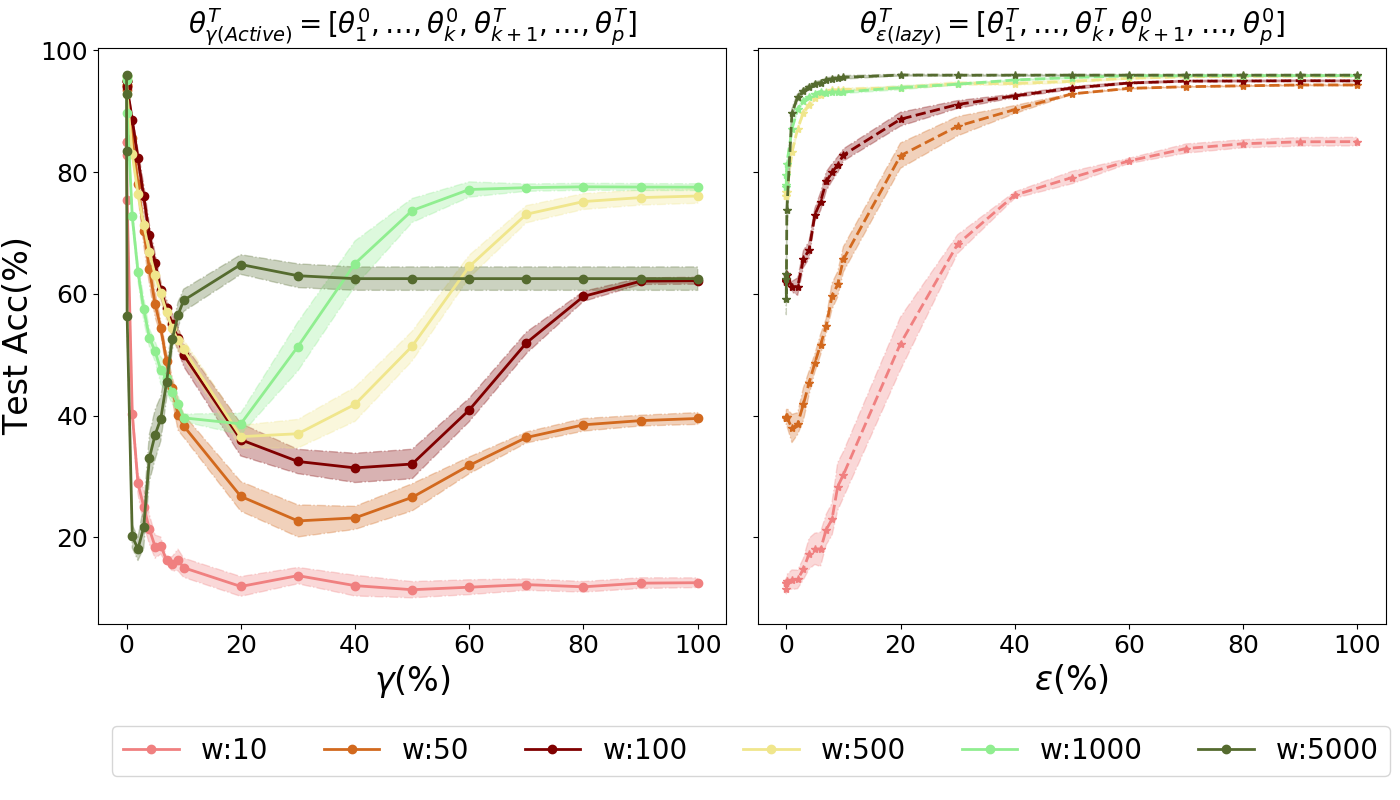}}    
    \caption{Test accuracy of 5-Layer ReLU-Nets ($d=4$) as a function of the choice of $\gamma$ and $\varepsilon$ after applying $\gamma$ Active-re-initialization (left) and $\varepsilon$ Lazy-re-initialization (right). (a) The x-axis is in log scale, and (b) the x-axis is in linear scale. As $\gamma$ increases, the test performance plunges at first then increases. As $\varepsilon$ increases, the test performance gradually improves. With $w=5,000$, keeping less than $10\%$ of parameters at their trained value is sufficient to recover the performance of the fully trained network($\varepsilon= 100\%$).
    }
    \label{fig:relu_mnist_reinit_d4}
    \vspace{-5mm}
\end{figure}

\begin{figure}[t]
    \hspace{-3mm}
    \centering
    \subfigure[Relu Network, d=1 (2 layers), MNIST.]{
    \includegraphics[width=0.48\textwidth]{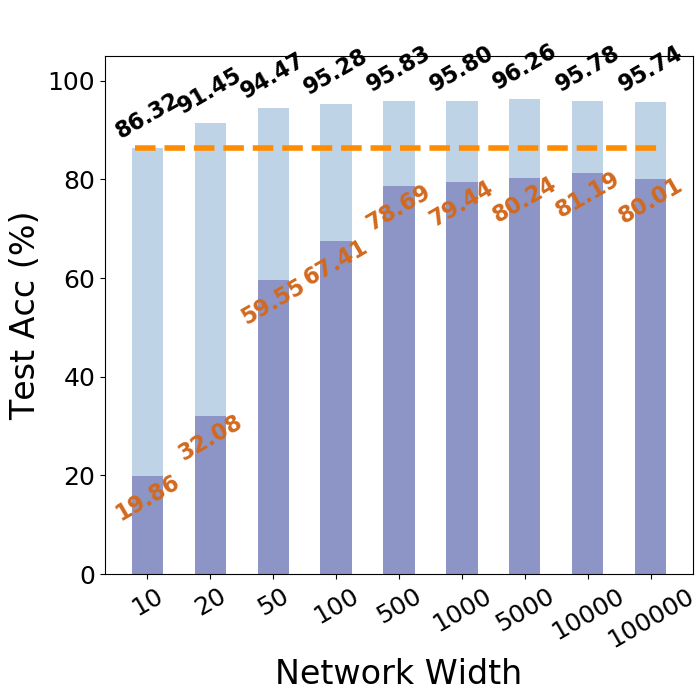}}
    \subfigure[Relu Network, d=4 (5 layers), MNIST.]{
    \includegraphics[width=0.48\textwidth]{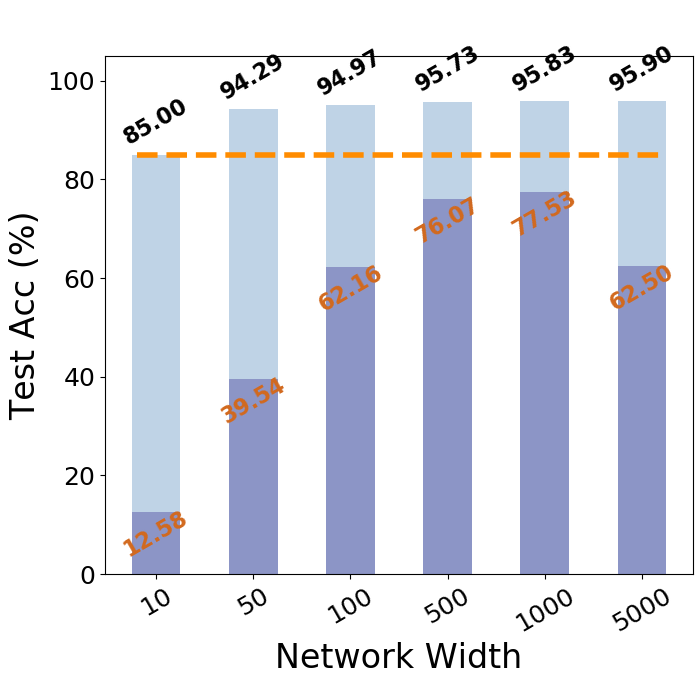}}
    \vspace{-3mm}
    \caption{Test accuracy at initialization and convergence for (a) 2-Layer ($d=1$) and (b) 5-Layer ($d=4$) ReLU-Nets trained on MNIST. The dark blue bar represents the test accuracy at initialization of which the value is shown in orange and the light blue bar shows the difference in performance between initialization and convergence. The final test accuracy is shown in black. Increasing width improves the generalization at random initialization. However, a trained network with $w=10$ can beat a random network with $w=100,000$.}
     \label{fig:relu_mnist_init_vs_final}
\end{figure}

When applying $\gamma$ Active-re-initialization, $\gamma=0\%$ corresponds to the trained parameters at convergence, and $\gamma=100\%$ represents the corresponding (random) parameters at initialization. As $\gamma$ increases, the generalization (test) performance plunges at first and then increases where the increase is more for wider networks (Figures~\ref{fig:relu_mnist_reinit_d1} and \ref{fig:relu_mnist_reinit_d4}). In particular, even re-initializing 0.1\% of the most active parameters adversely affects ReLU-Nets across all widths and most models have the worst performance with around 10\% of the active parameters are re-initialized. Interestingly, after hitting rock bottom, the generalization performance of models, especially ones with larger width, increases all the way to the point where all parameters have been re-initialized, i.e., random network. Wider random networks perform better, with width $w=100,000$ having the best performance for 2-layer ($d=1$) networks, illustrating the promise of lazy or kernel regime. However, the performance of random networks ($\gamma=100\%$) falls significantly short of that of the trained network ($\gamma=0\%$) across all widths (Figure~\ref{fig:relu_mnist_init_vs_final}). Moreover, after a point, increasing the width does not improve the

\clearpage

\clearpage

performance of the random network, e.g., for 2-layer networks, random networks with $w=1000$ through $w=100,000$ have the same performance. Further, a trained network with $w=10$ outperforms a random network with $w=100,000$.

We examine how many active parameters are necessary to maintain the generalization of an over-parameterized ReLU-Net through the study of $\varepsilon$ Lazy-re-initialization. As $\varepsilon$ varies from $0$ to $100\%$, i.e., more lazy parameters are allowed to take their trained values, the behavior is somewhat the opposite of re-initializing active parameters. For small values of $\varepsilon$, there is little impact on generalization and the performance is similar to that of the random network ($\varepsilon=0\%$), with wider networks having better performance. As $\varepsilon$ increases, at some point there is a decrease in generalization performance, especially for wider networks. Further increase in $\varepsilon$ leads to increase in generalization performance till it reaches the performance of the fully trained network ($\varepsilon=100\%$). Our experiments indicate a clear separation in performance between the rich regime and the lazy (kernel) regime, supporting and strengthening the concerns about the lazy regime in \citep{chizat2019}.

\label{sec:generalization}

\section{Layer-Wise Distribution: Active vs.~Lazy Parameters}

\begin{figure*}[t]
    \centering
    \subfigure[$d=4, w=100, \alpha=1\%$.]{
    \includegraphics[width=0.48\textwidth]{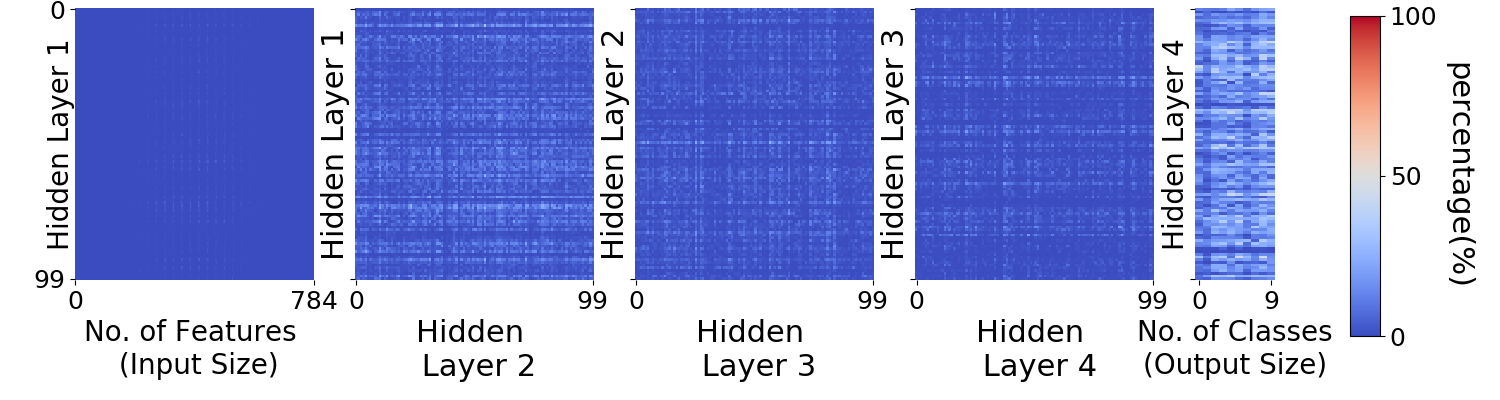}}
    \subfigure[$d=4, w=1000, \alpha=1\%$.]{
    \includegraphics[width=0.48\textwidth]{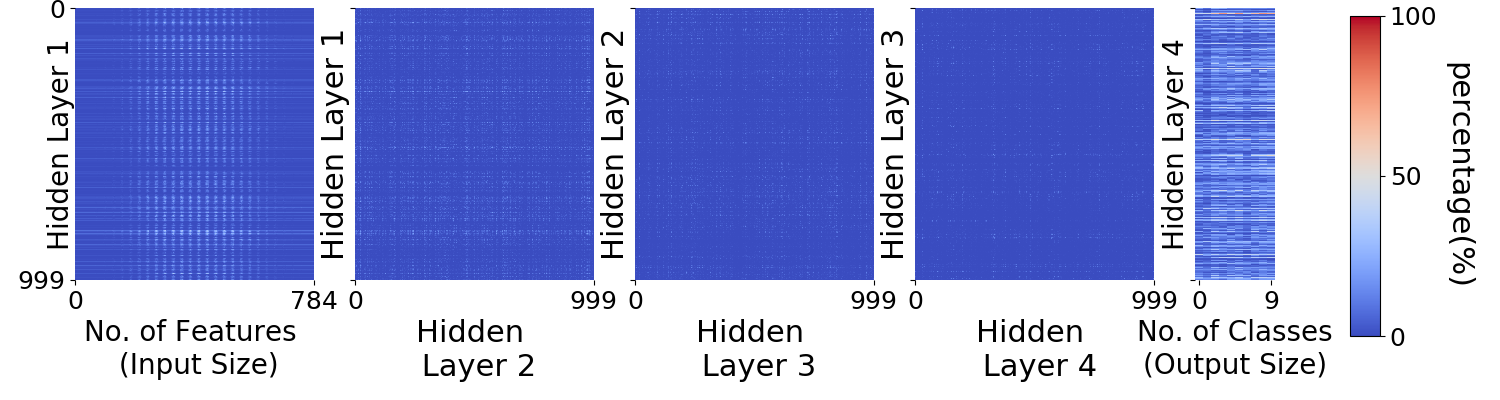}}
    \subfigure[$d=4, w=100, \alpha=10\%$.]{
    \includegraphics[width=0.48\textwidth]{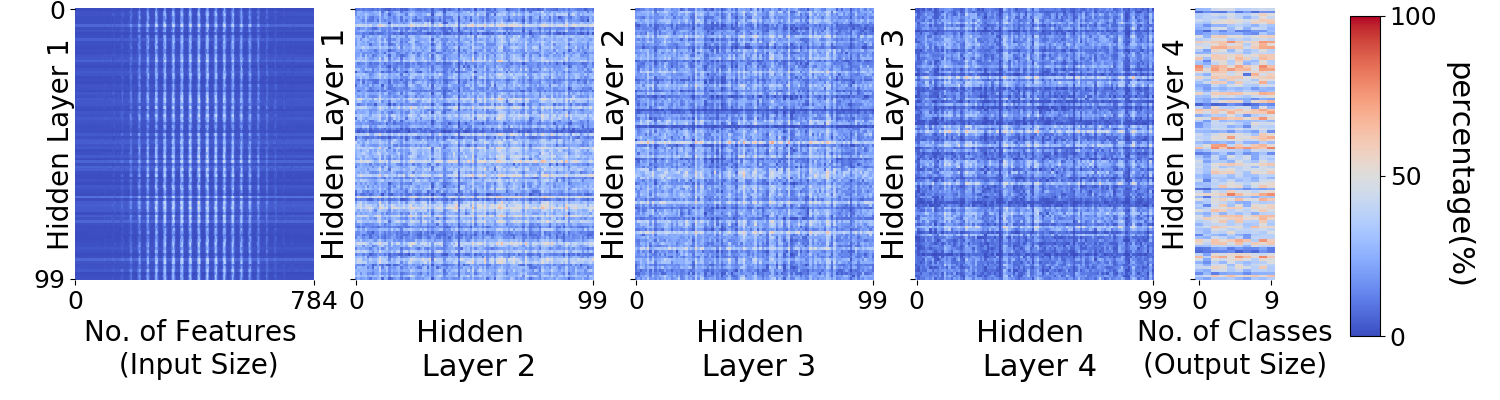}}
    \subfigure[$d=4, w=1000, \alpha=10\%$.]{
    \includegraphics[width=0.48\textwidth]{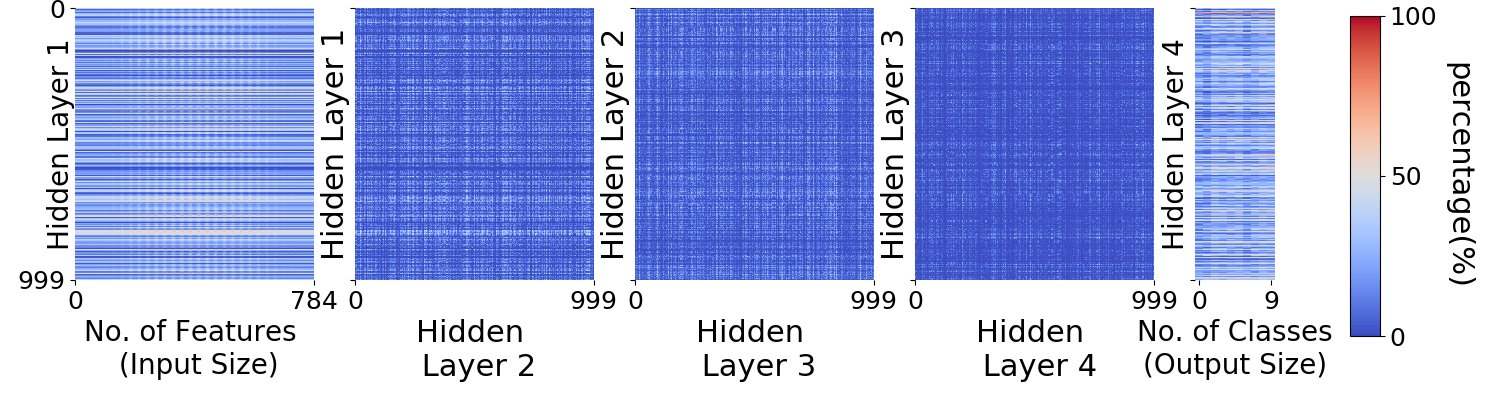}}
    \subfigure[$d=4, w=100, \alpha=30\%$.]{
    \includegraphics[width=0.48\textwidth]{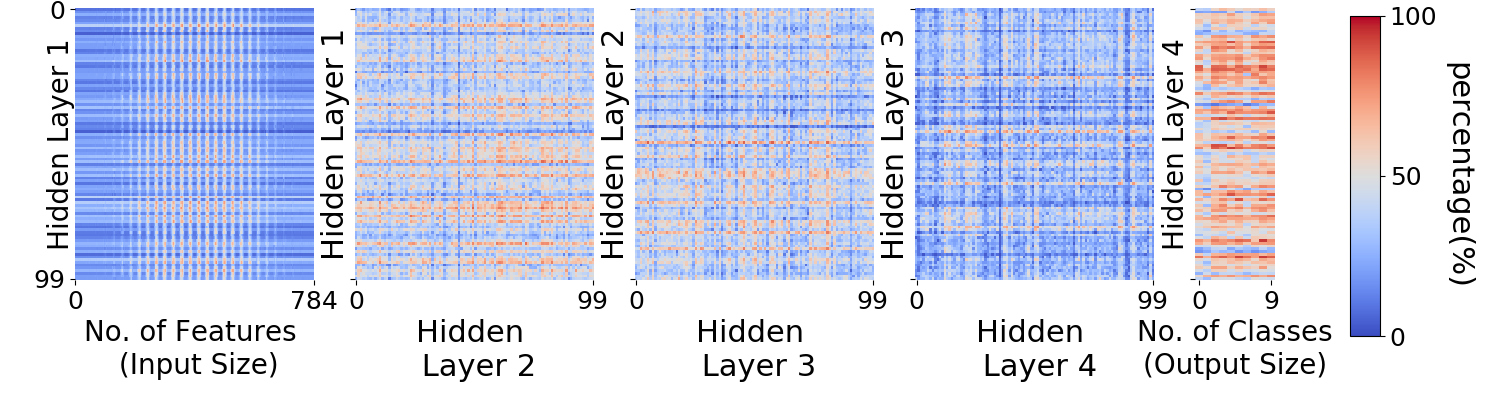}}
    \subfigure[$d=4, w=1000, \alpha=30\%$.]{
    \includegraphics[width=0.48\textwidth]{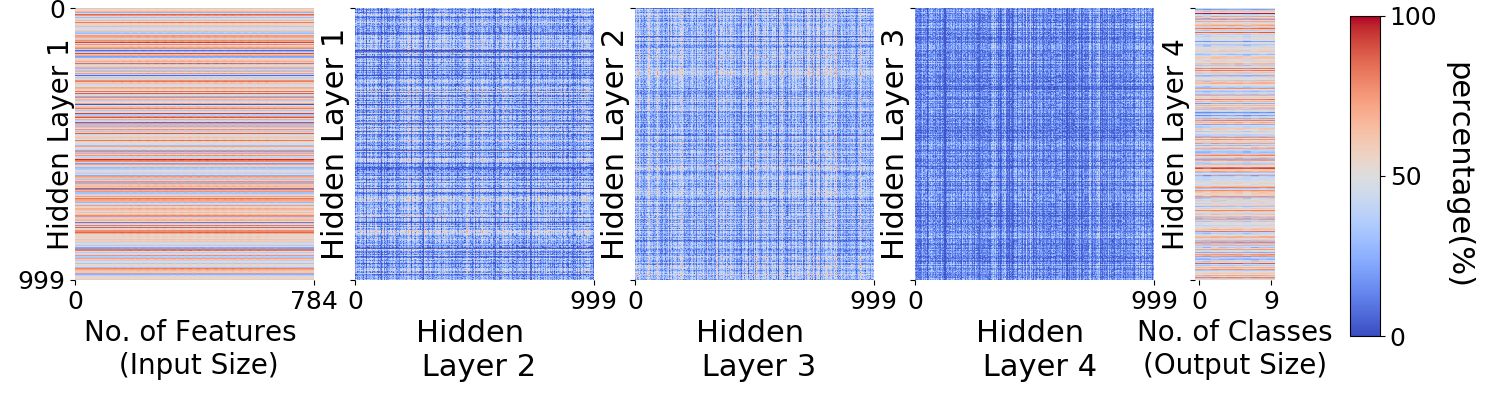}}
    \caption{Frequency of $\theta_i$ been selected as active during training with different choice of $\alpha$ for 5-Layer ReLU-Nets ($d=4$) trained on MNIST. The results are the average over 5 repetitive runs. Each rectangle consists of parameters connecting layer $l_i$ and $l_{i-1}$. Red indicates a high frequency, close to $100\%$ and blue means low frequency, close to $0$. When $w$ is small, active parameters are spread across all layers. As we increase the width, active parameters in a ReLU-Net are concentrated to both the very top and the very bottom layer.
    }
     \label{fig:relu_mnist_active}
\end{figure*}

\begin{figure*}[t]
    \centering
    \subfigure[$d=2, w=10, \alpha=1\%$.]{
    \includegraphics[width=0.48\textwidth]{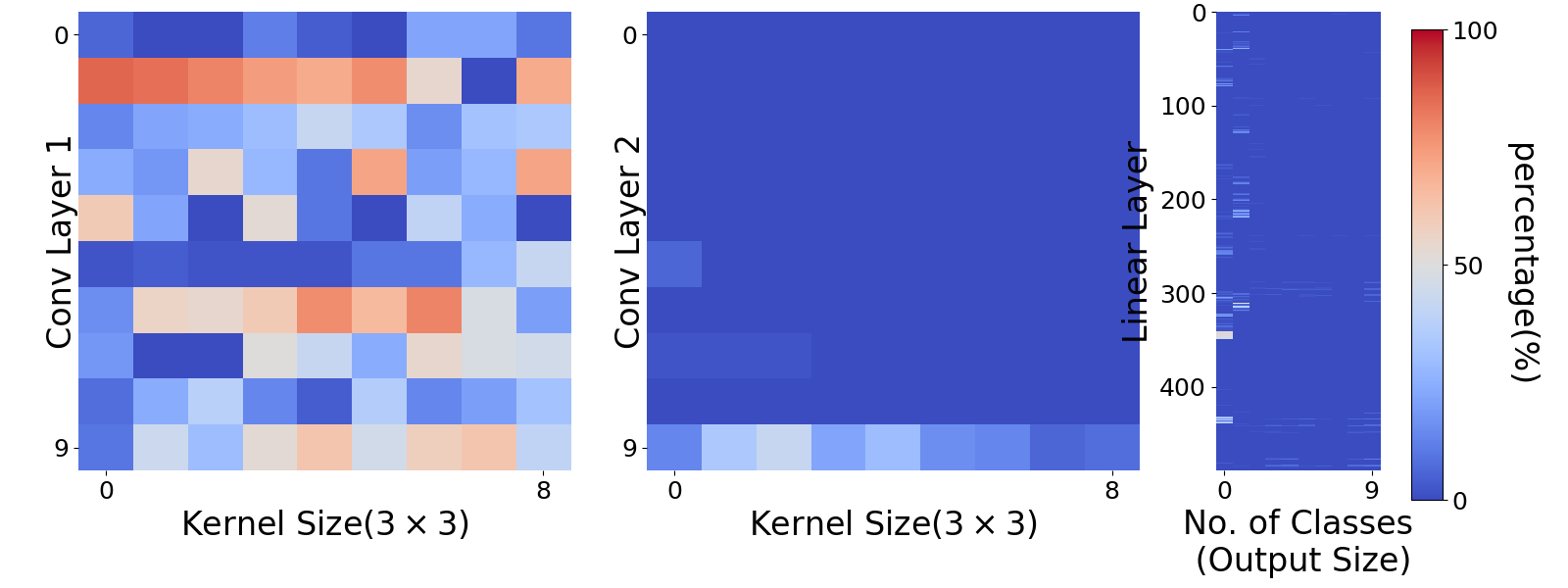}}
    \subfigure[$d=2, w=100, \alpha=1\%$.]{
    \includegraphics[width=0.48\textwidth]{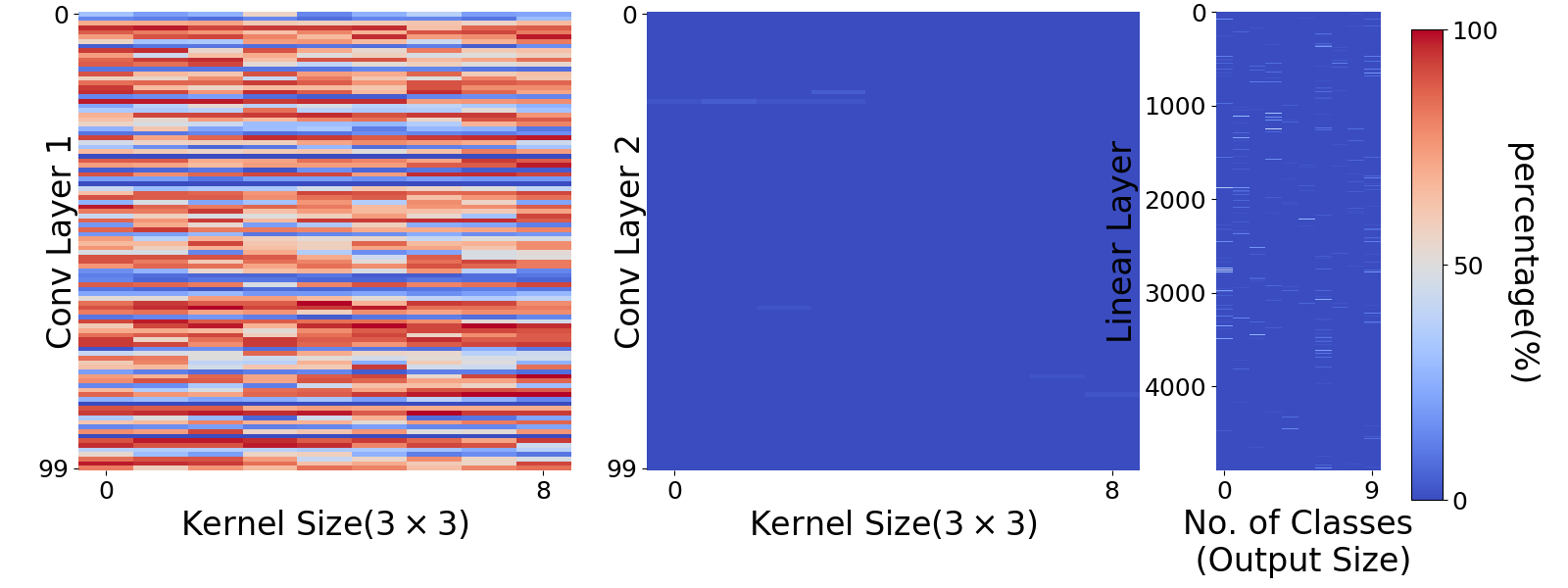}}
    \subfigure[$d=2, w=10, \alpha=10\%$.]{
    \includegraphics[width=0.48\textwidth]{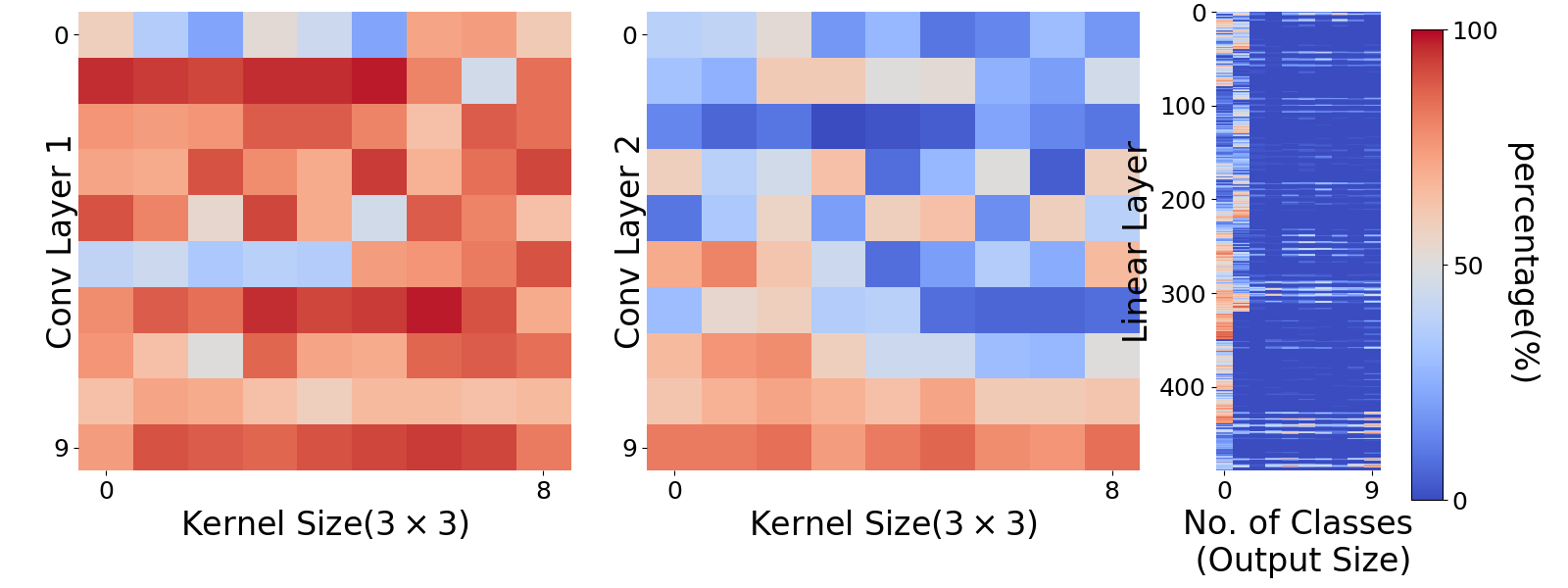}}
    \subfigure[$d=2, w=100, \alpha=10\%$.]{
    \includegraphics[width=0.48\textwidth]{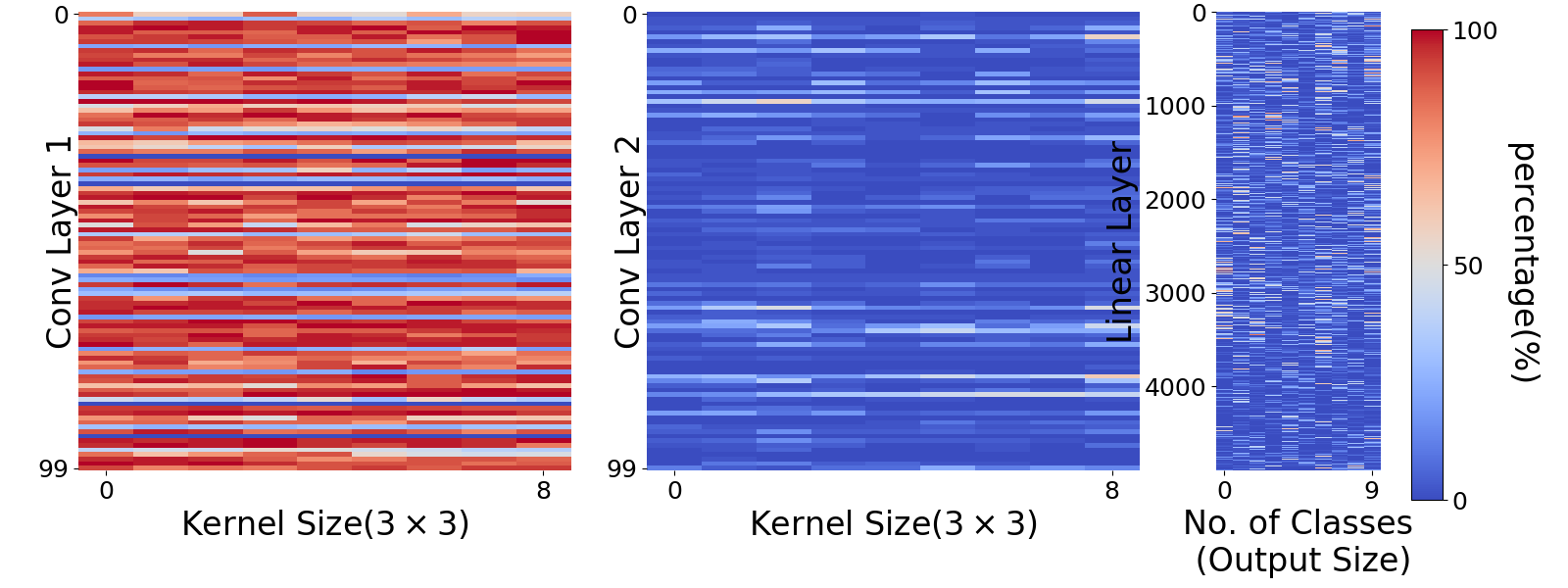}}
    \subfigure[$d=2, w=10, \alpha=30\%$.]{
    \includegraphics[width=0.48\textwidth]{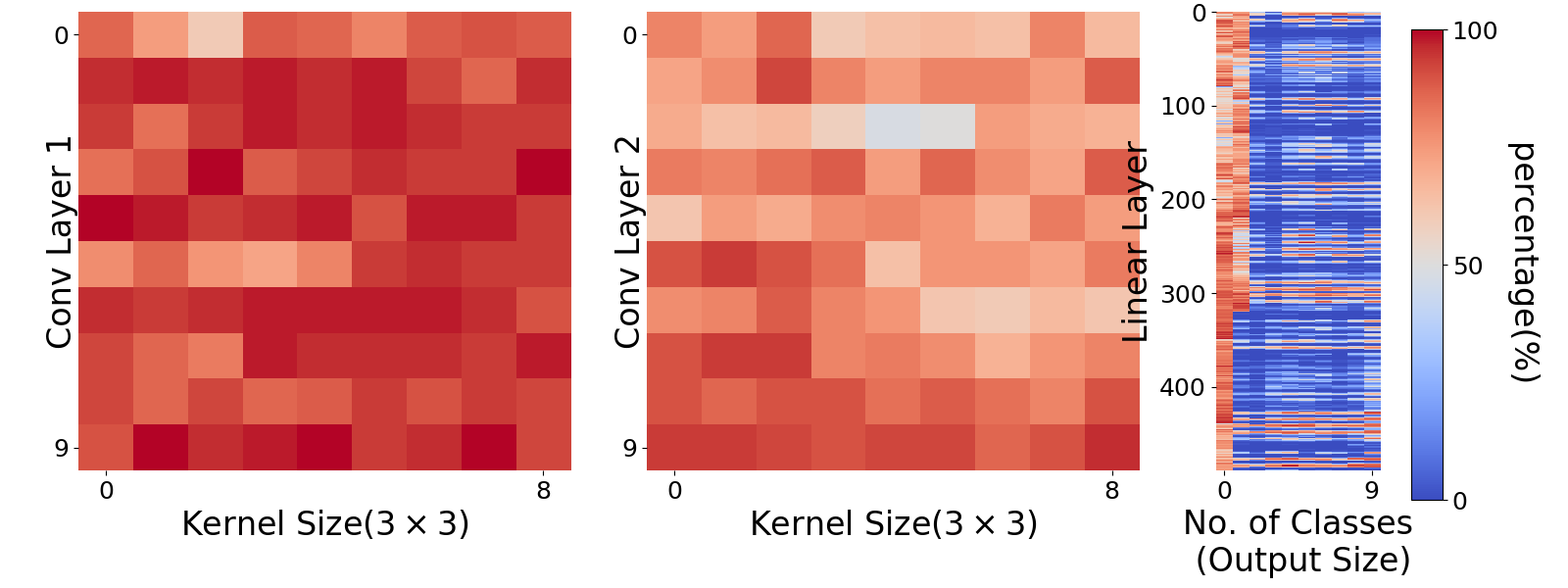}} 
    \subfigure[$d=2, w=100, \alpha=30\%$.]{
    \includegraphics[width=0.48\textwidth]{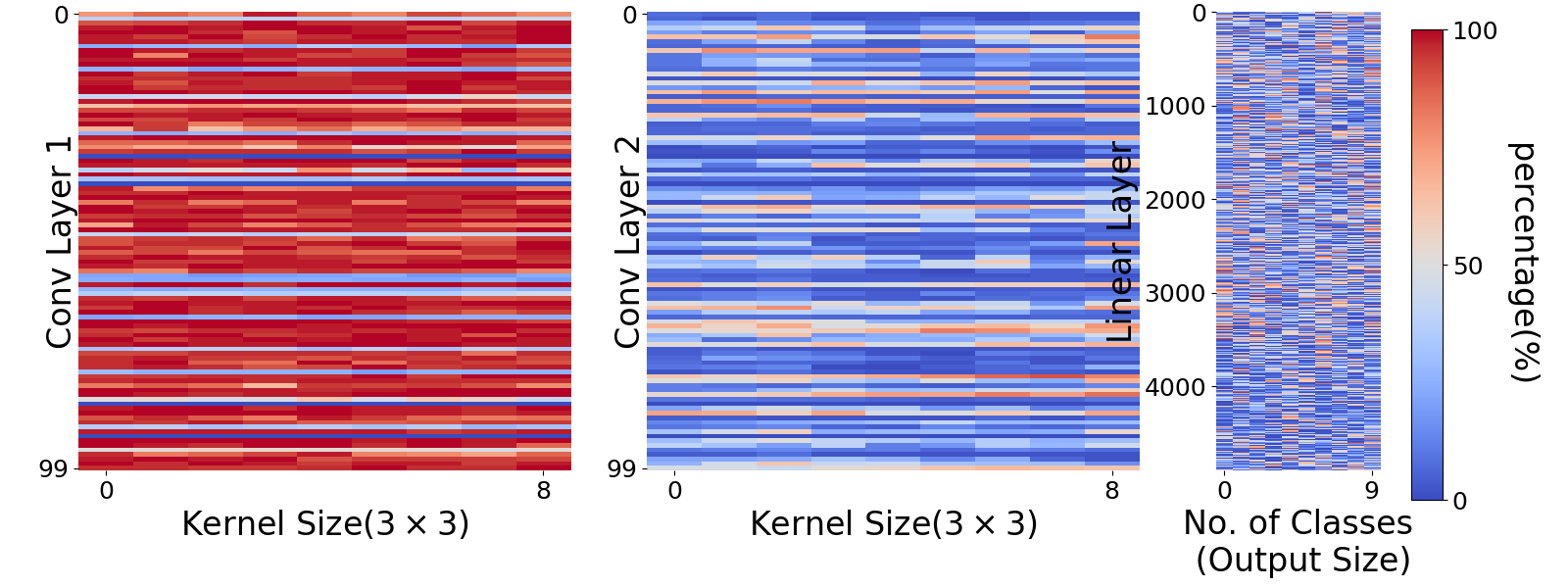}}
    \vspace{-2mm}
    \caption{Frequency of $\theta_i$ been selected as active during training for different choice of $\alpha$ for Conv-Nets ($d=2$) trained on MNIST. The results are the average over 5 repetitive runs.
    Each rectangle consists of parameters of a $3\times3$ kernel at layer $l_i$. Red indicates high frequency, close to $100\%$ and blue means low frequency, close to $0$. As we increase the width, the active parameters in a Conv-Net concentrate at the very bottom layer.
    }
    \vspace{-4mm}
     \label{fig:vgg2_mnist_active}
\end{figure*}

Next, we study how the active parameters are spread over the layers. Will they be spread across the entire network or be concentrated in certain layers? To seek an answer, we track how the active subspace evolves during training. At each epoch $t\in [T]$, we compute the full batch gradient $\g^{t}$ and sort the gradient by its absolute value $|\g^t|$. We consider the largest $\alpha\%$ elements in sorted gradient as active and count the number of times a parameter $\theta_i$ becomes active over $T$ training epochs. Figures~\ref{fig:relu_mnist_active} and \ref{fig:vgg2_mnist_active} show how often has a parameter $\theta_i$ for $i \in [1,\dots,p]$ been considered active during training with various $\alpha$ values (1\%, 10\%, 30\%) for ReLU-Nets and Conv-Nets respectively. We organize the parameters by their relative position, and each colored rectangle represents a weight matrix containing parameters connecting layer $l_i$ and layer $l_{i-1}$. While the darker blue means $\theta_i$ has rarely became active, we use darker red to indicate that $\theta_i$ has often been selected as active.

For both ReLU-Nets and Conv-Nets, width has a key impact on the distribution of active parameters. For small width, viz.~$w=100$ for ReLU-Nets (Figure~\ref{fig:relu_mnist_active}(a),(c), and (e)) and $w=10$ for Conv-Nets (Figure~\ref{fig:vgg2_mnist_active}(a),(c), and (e)), active parameters are spread across all layers. For larger width, the active parameters are concentrated in the bottom layer for both ReLU-Nets and Conv-Nets, and also in the top layer for ReLU-Nets (Figures~\ref{fig:relu_mnist_active} and \ref{fig:vgg2_mnist_active}, (b),(d), and (f)). Given our current understanding of the inductive bias of SGD-type algorithms in the rich regime, especially the sparsity~\citep{gunasekar2018conv,wogl2019}, it is not surprising that the gradients over epochs are sparse. Our experiments show that the active non-sparse components concentrate primarily on the bottom layer.

\label{sec:dynamics}

\section{Layer-Wise Sparse (LWS) SGD}
\label{sec:layer_sparse}
\begin{figure*}[t]
    \centering
    \subfigure[VGG-5, MNIST.]{
    \includegraphics[width=0.6\textwidth]{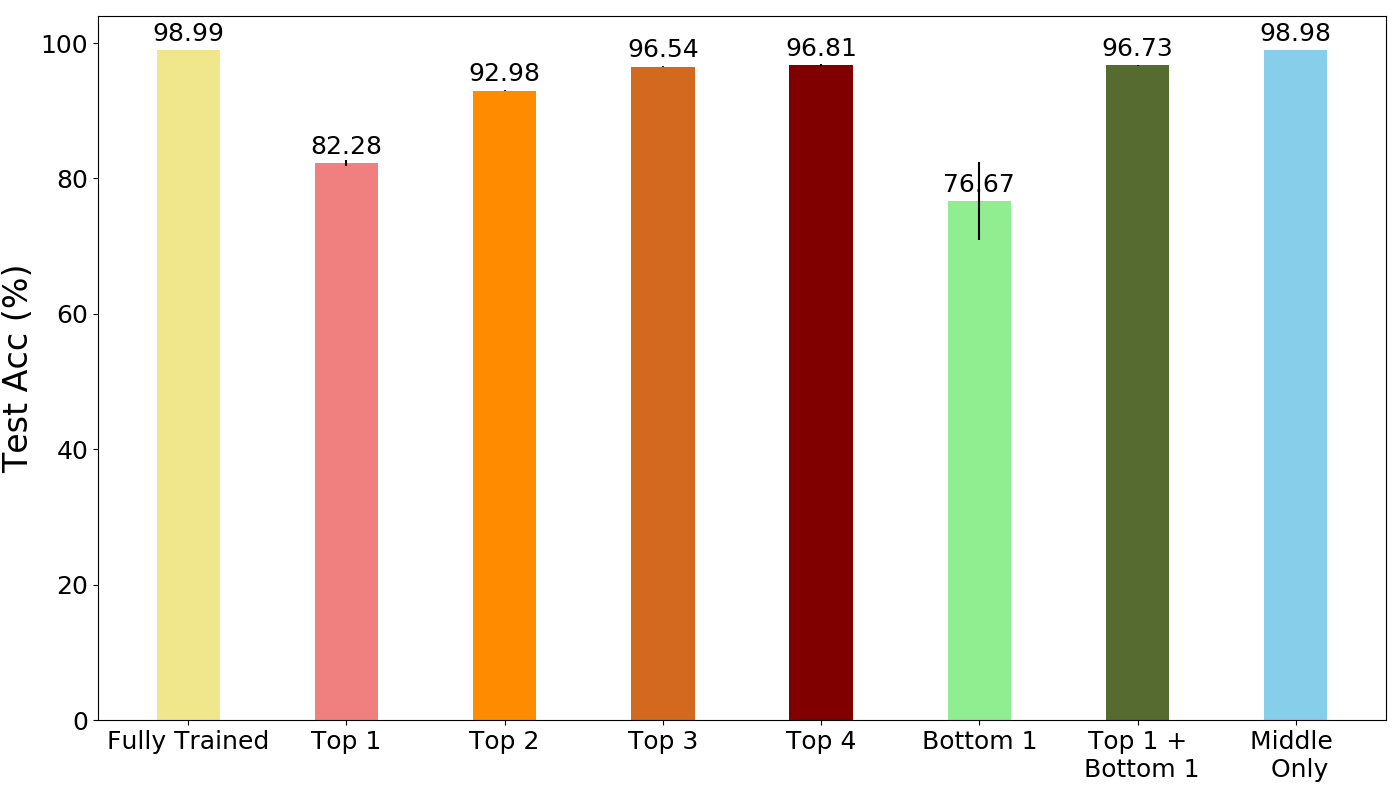}}
    \subfigure[VGG-11, CIFAR-10]{
    \includegraphics[width=0.6\textwidth]{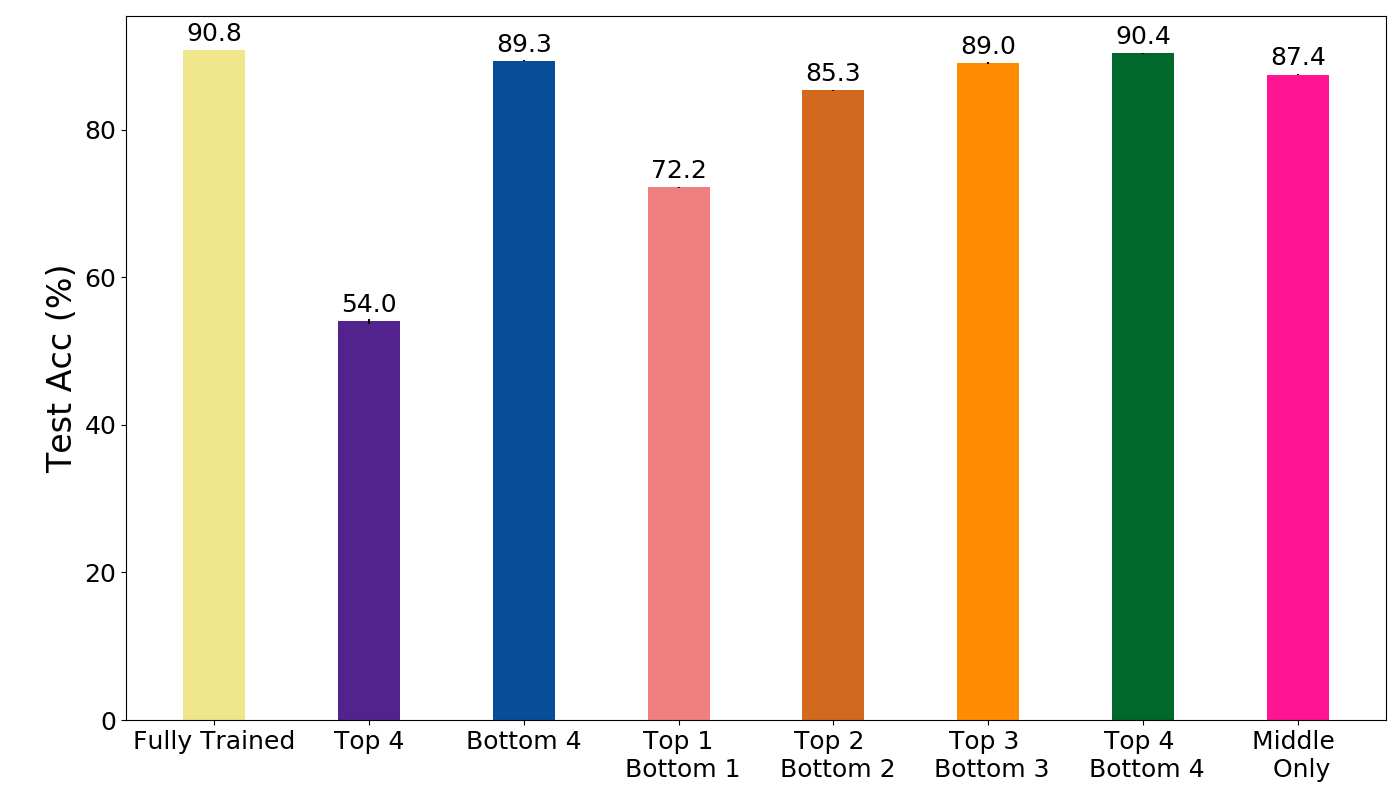}}
    \vspace{-2mm}
    \caption{Test accuracy ($\%$) of networks trained by the following static LWS-SGD variants: (a) Top $k$ only, (b) Bottom $q$ only, (c) Top $k$ $+$ Bottom $q$, and (d) Middle only. The parameters in other layers are frozen during the training. The bar shows the average test accuracy at convergence, computed over 5 repetitive runs, and the error bar shows the corresponding one standard error. In general, training an appropriate combination of top and bottom layers works the best with little or no damage on generalization.}
    \vspace{-3mm}
     \label{fig:vgg5_mnist_retrain}
\end{figure*}

\begin{figure*}[t]
    \centering
    \subfigure[VGG-5, MNIST, 100 Epochs.]
    {\includegraphics[width=0.6\textwidth]{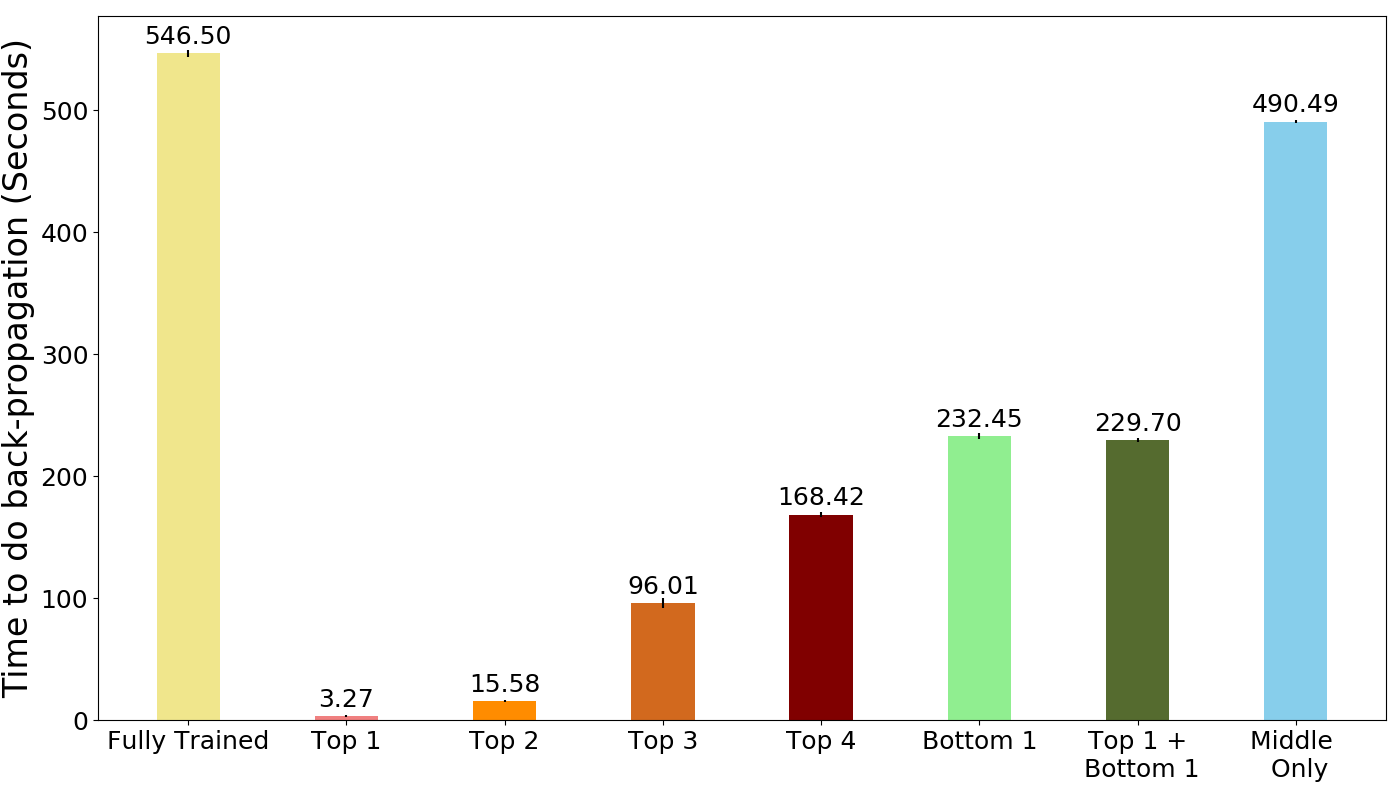}}
    \subfigure[VGG-11, CIFAR-10, 180 Epochs.]{
    \includegraphics[width=0.6\textwidth]{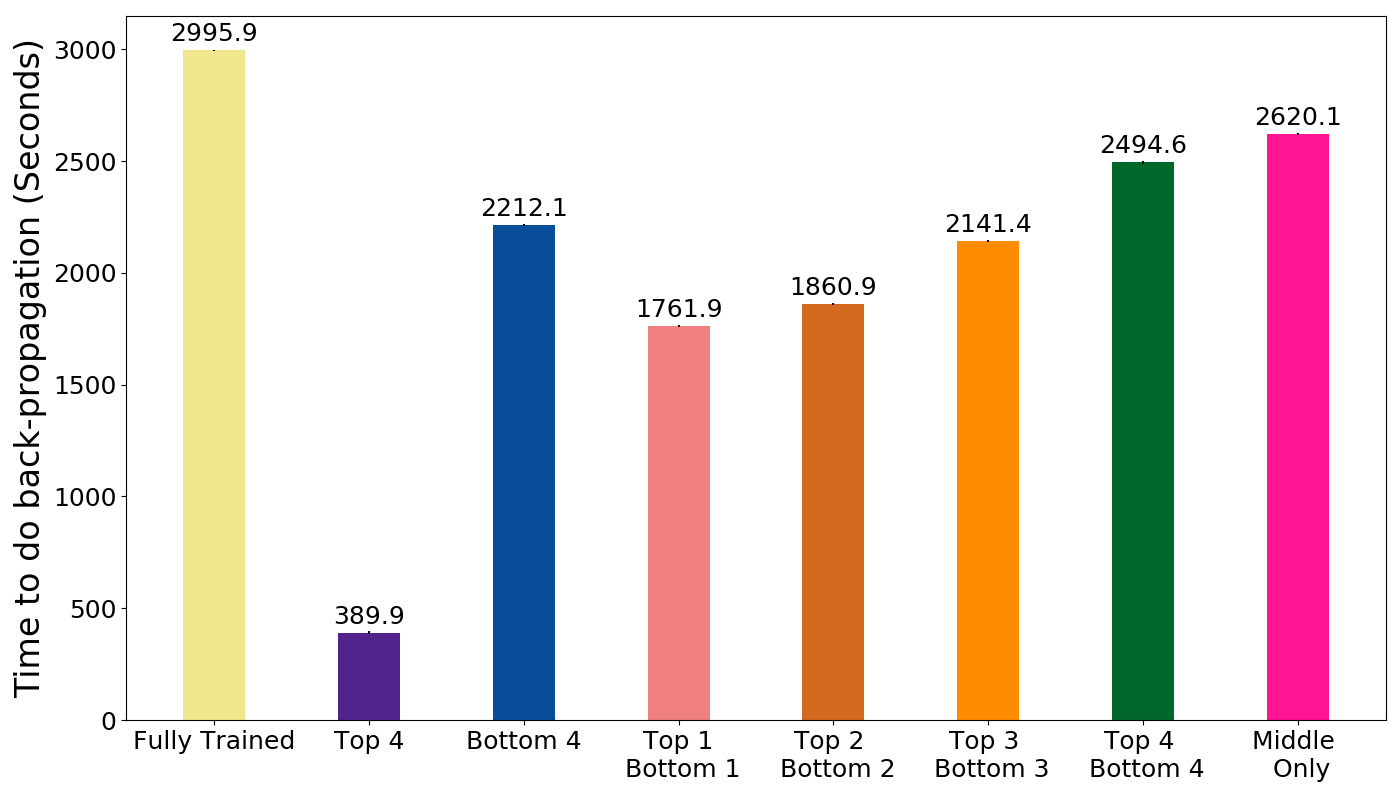}}
    \vspace{-2mm}
    \caption{Time (in seconds) to perform back-propagation for networks trained by the following static LWS-SGD variants: (a) Top $k$ only, (b) Bottom $q$ only, (c) Top $k$ $+$ Bottom $q$, and (d) Middle only. The bar shows the average time layer-sparse SGD required to perform back-propagation, computed over 5 repetitive runs, and the error bar shows the corresponding one standard error. Overall, training the top k layer(s) only is the most efficient approach.}
    \vspace{-4mm}
     \label{fig:vgg5_mnist_retrain_time}
\end{figure*}

Inspired by our observation that the bottom and top few layers are more active during training than the middle ones, we explore whether training can be focused only on these layers, more generally a subset of layers, without significantly sacrificing generalization. Towards this end, we study the generalization performance of a variety of Layer-Wise Sparse (LWS) SGD variants, each of which only updates a certain subset of layers during training. 

\vspace{-2mm}
\subsection{Static Layer-Wise Sparse SGD} 
We consider training layer-wise sparse models by considering parameters in (a) the top $k$ layer(s) only, (b) the bottom $q$ layer(s) only, (c) both the top $k$ and the bottom $q$ layers, and (d) the middle layers only which exclude the very top ($k=1$) and the very bottom layer ($q=1$). The parameters in the remaining layers are kept frozen at their initial values during the entire training process. Figure~\ref{fig:vgg5_mnist_retrain} compares their generalization performance with the fully trained model for (a) VGG-5 on MINST and (b) VGG-11 on CIFAR-10. 

Considering an easy problem, such as MNIST, training the top 1 and the bottom 1 layer at the same time can almost attain the generalization of the fully trained network. Training only the top $k$ layers for large enough $k$ can also achieve a reasonable good generalization performance. The option of training many middle layers works ok as well. Overall, the degradation in generalization performance does not exceed $2\%$ for aforementioned option (a) ($k=3,4$), (c), and (d).

When dealing with CIFAR-10, since the difficulty of the problem increases, training the top $k$ layer(s) alone does not work well (purple bar in Figure~\ref{fig:vgg5_mnist_retrain} (b)). At the same time, more bottom layers need to be involved in order to achieve a reasonably good generalization performance for a LWS model only relying on the bottom few layers. Training all the middle layers still does an acceptable job with less than $5\%$ performance drop. The combination that works the best is option (c), training top-$k$ and bottom-$q$ layers. With a suitable choice of $k$ and $q$,  the joint training of both the top and the bottom layers can almost maintain the generalization of the fully trained model.

{\bf Discussion.} The static LWS-SGD we explore is different from gradient sparsification \citep{Aji2017,Alistarh2018,stich2018} used in distributed SGD. In gradient sparsification, the sparse structure of the gradient is obtained by dropping a majority of small elements of the gradient. Since it requires the computation of full-size gradient first, any gradient sparsification step imposes an extra computational overhead. The static LWS-SGD, on the other hand, decides which layer(s) to update before the training starts and never computes the full gradient. The effective dimension of the resulting gradient depends on which layer(s) static LWS-SGD  employs and is usually much smaller than the full model dimension $p$. The reduction in gradient dimension implies that LWS-SGD  can be made computationally efficient, even though errors have to be back-propagated all the way to the bottom. Further, LWS-SGD can have considerably less memory I/O.

\begin{figure*}[th]
    \centering
    \subfigure[VGG-5, MNIST, 100 Epochs.]{
    \includegraphics[width=0.78\textwidth]{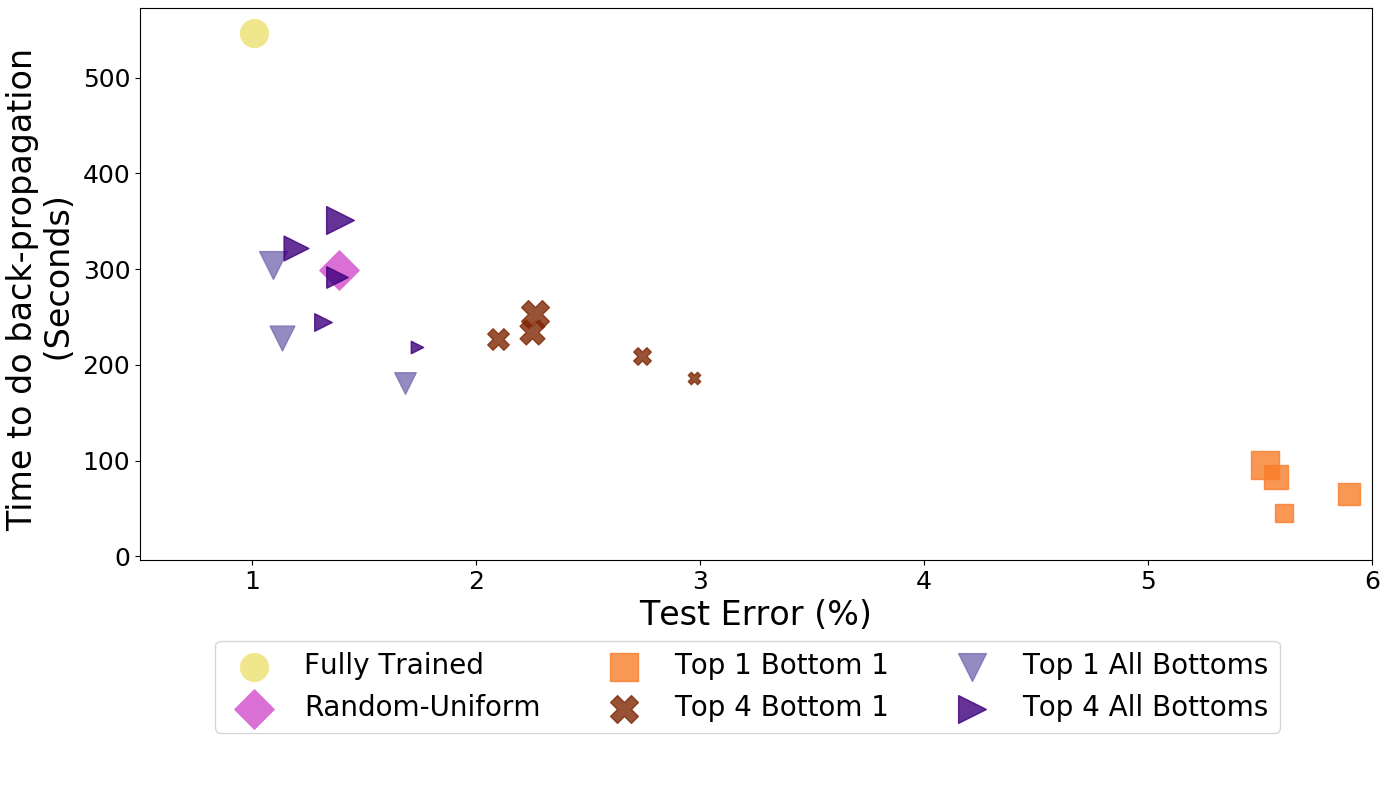}}
    \subfigure[VGG-11, CIFAR-10, 180 Epochs.]{
    \includegraphics[width=0.78\textwidth]{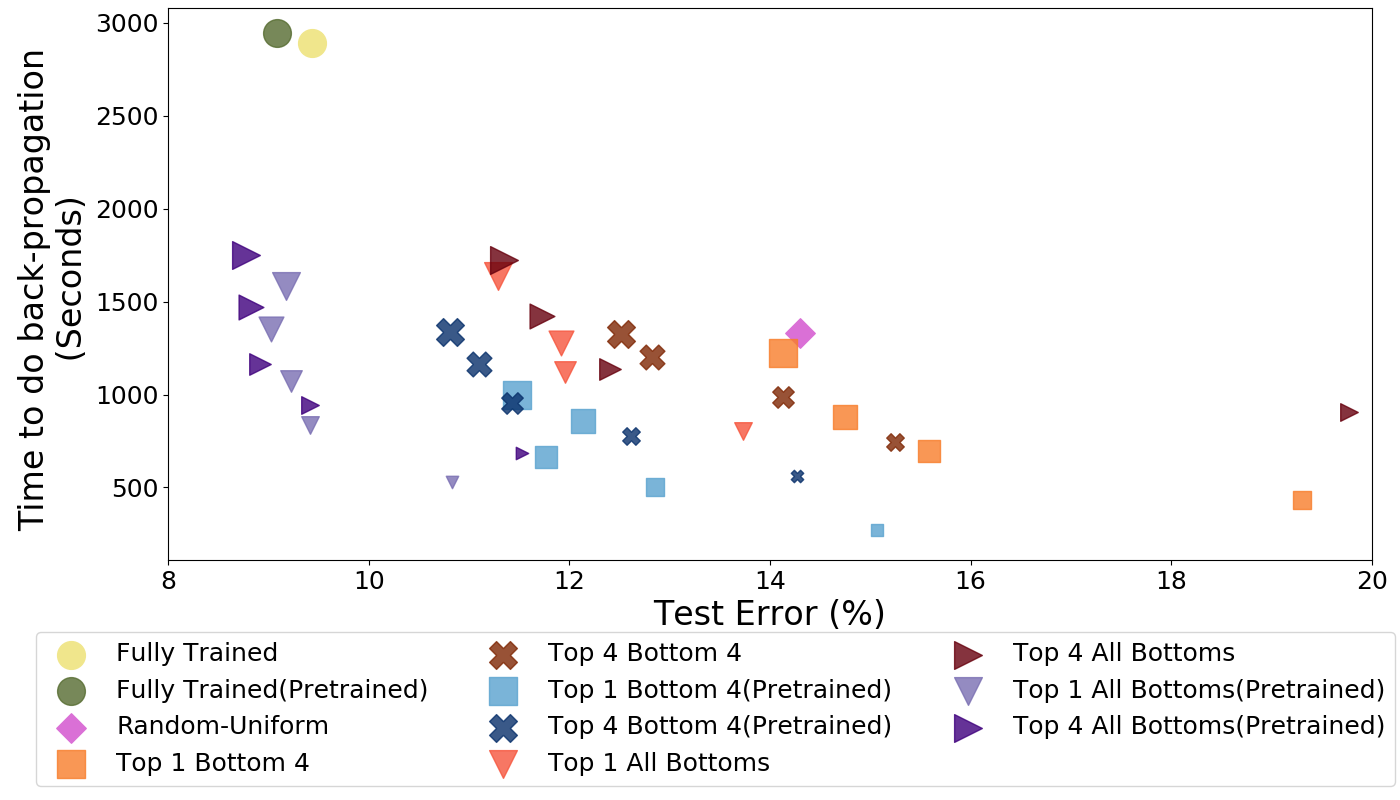}}
    \caption{Test Error($\%$) versus Time (in seconds) to perform back-propagation for networks trained by the following probabilistic LWS-SGD variants: (a) Top $k$ Bottom $q$, (b) Top $k$ All Bottoms, and (c) Random-Uniform. VGG-5 is initialized with Xavier initialization and VGG-11 is initialized using both Xavier initialization and pre-trained weights from ImageNet. The size of the marker indicates the probability $\rho\in [0.1,0.2,0.3,0.4,0.5]$ of updating the bottom layers. The larger the marker is, the more frequent probabilistic LWS-SGD updates the bottom layers. An ideal algorithm should be as close to the origin as possible, such that not only it generalizes well but also is very efficient. Among all SGD-type algorithms, Top $k$ All Bottoms (VGG-5 with Xavier initialization and VGG-11 with pre-trained weights) is always the one presented in the lower left corner. It only needs a small probability, e.g., $\rho=0.1$, to achieve similar generalization performance of the fully trained model, yet approximately 2 to 5 times faster than doing full back-propagation. }
    \label{fig:vgg11_cifar_adaptive_err_vs_time}
\end{figure*}
\clearpage

\begin{figure*}[t]
    \centering
    \vspace*{-2mm}
    \subfigure[VGG-5, MNIST.]{
    \includegraphics[width=0.6\textwidth]{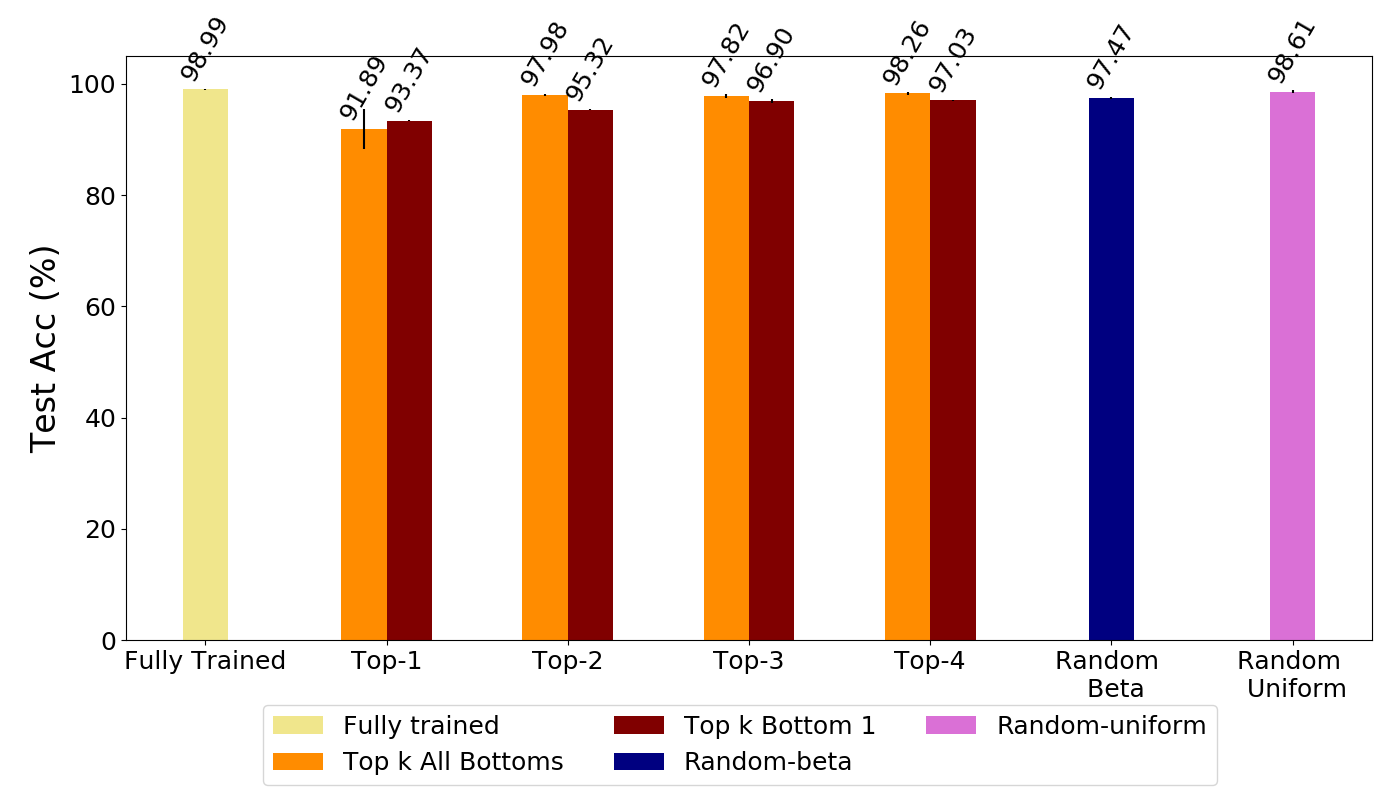}}
    \subfigure[VGG-11, CIFAR-10.]{
    \includegraphics[width=0.6\textwidth]{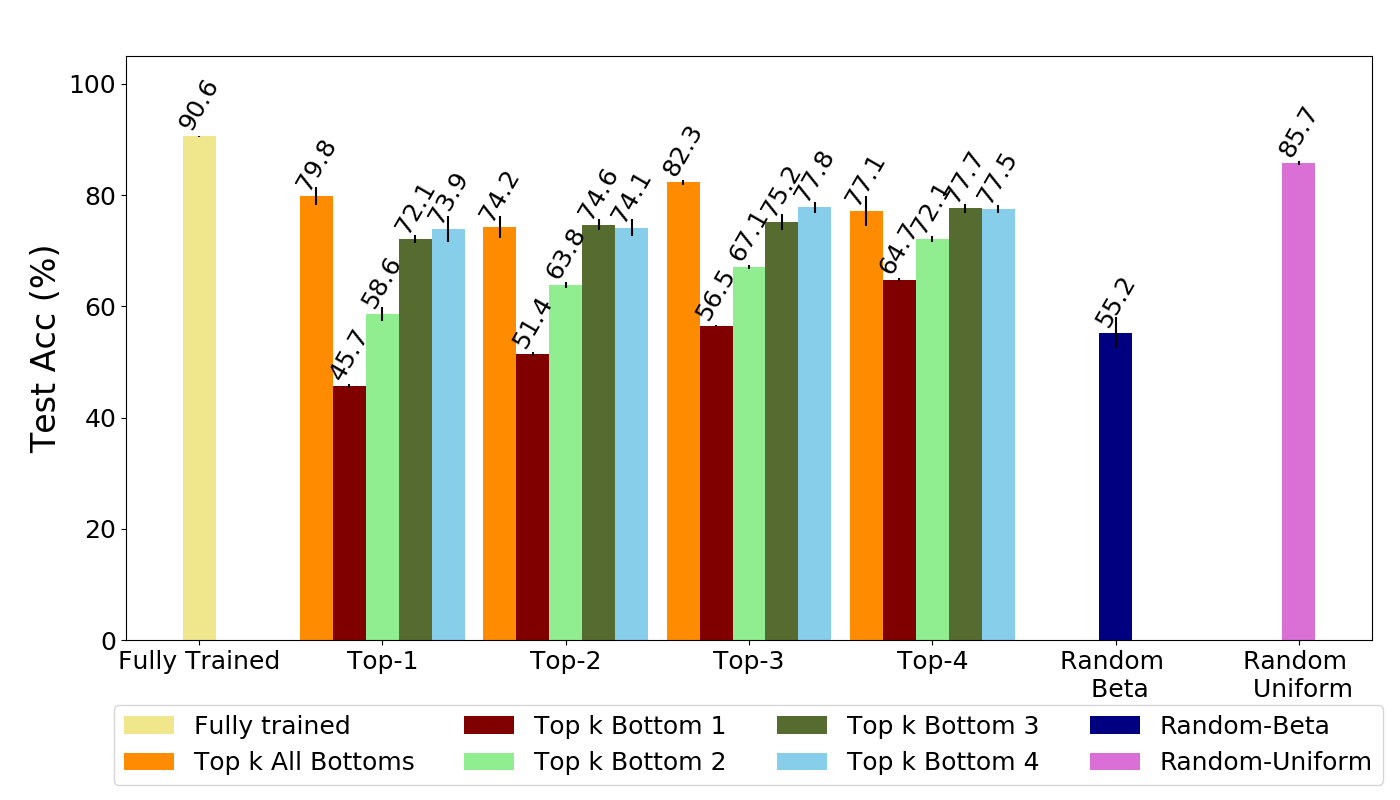}}
    \vspace{-3mm}
    \caption{Test accuracy ($\%$) of networks trained by the following probabilistic LWS-SGD variants: (a) Top $k$ Bottom $q$ ($\rho=0.1$), (b) Top $k$ All Bottoms ($\rho=0.1$), (c) Random-Beta, and (d) Random-Uniform. Networks are initialized with Xavier initialization. Both training with (a) Top $k$ All Bottoms (with sufficiently large $k$) and (d) Random-Uniform works fairly well on preserving the generalization.}
    \vspace{-3mm}
     \label{fig:vgg5_mnist_adaptive}
\end{figure*}

\begin{figure*}[t]
    \centering
    \subfigure[VGG-5, MNIST, 100 Epochs.]{
    \includegraphics[width=0.6\textwidth]{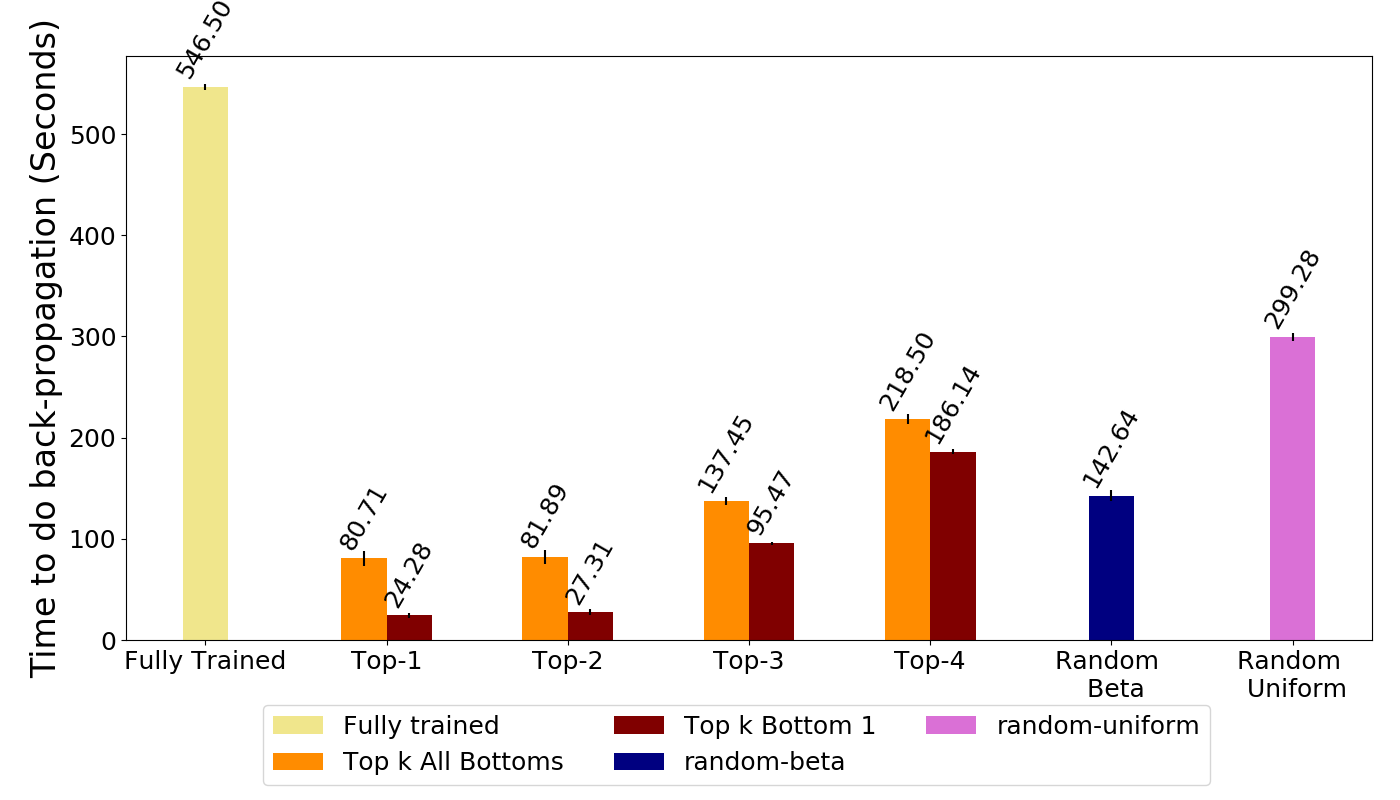}}
    \subfigure[VGG-11, CIFAR-10, 180 Epochs.]{\includegraphics[width=0.6\textwidth]{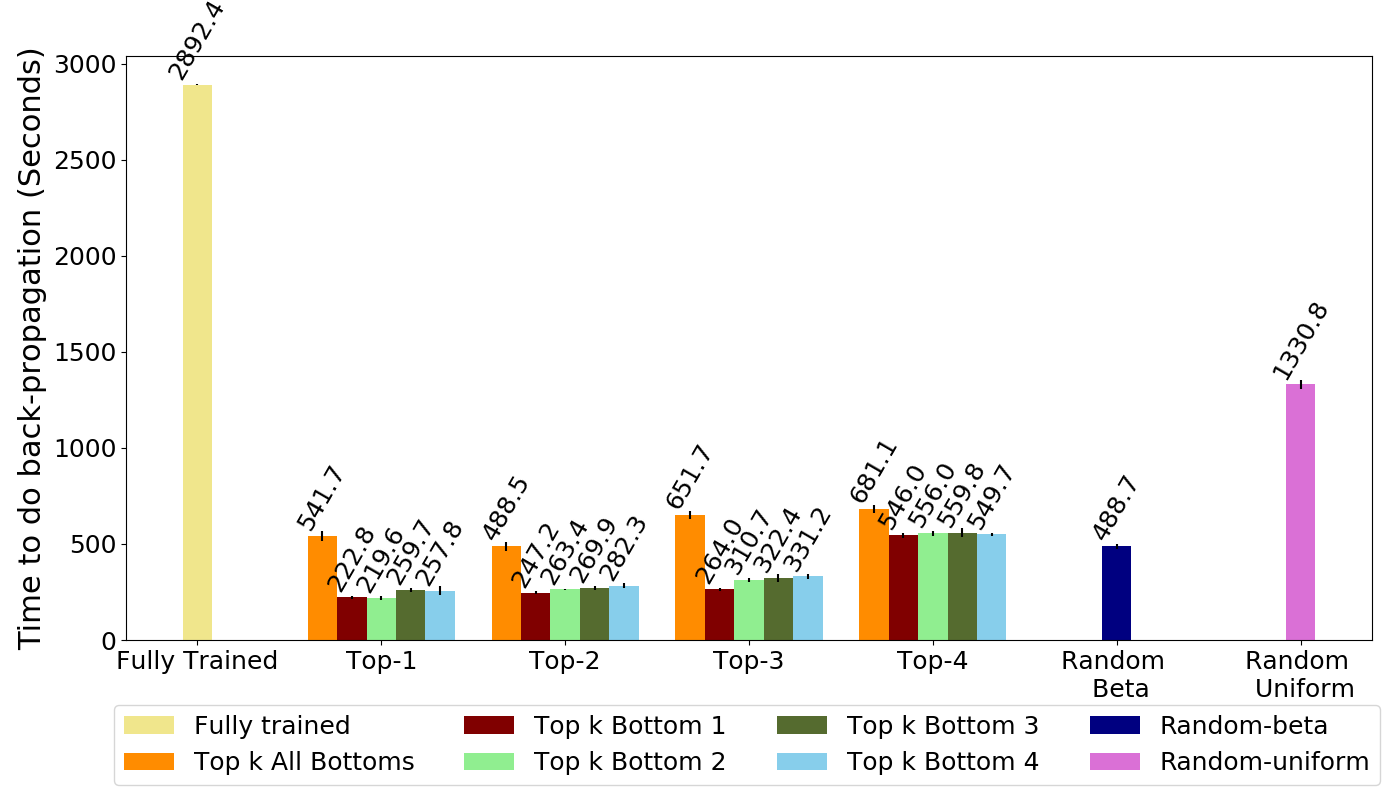}}
    \vspace{-3mm}
    \caption{Time (in seconds) to perform back-propagation for networks trained by the following probabilistic LWS-SGD variants: (a) Top $k$ Bottom $q$ ($\rho=0.1$), (b) Top $k$ All Bottoms ($\rho=0.1$), (c) Random-Beta, and (d) Random-Uniform. Networks are initialized with Xavier initialization. All four strategies can substantially reduce the computation.}
     \label{fig:vgg5_mnist_adaptive_time}
\end{figure*}

Figure \ref{fig:vgg5_mnist_retrain_time} demonstrates the exact time, measured in seconds, to perform back-propagation for each aforementioned LWS-SGD variant. Training the full model is the slowest overall and any LWS-SGD, to some extent, can reduce the computational cost. Training only the top $k$ layers is, as expected, the most efficient among all methods we explore, since it only requires the error to be propagated through the top $k$ layers and has the shortest computational graph. Both training the bottom $q$ layers only and training both the top and the bottom layers spend similar amount of time when doing back-propagation since both need to pass the error to the bottom layers. Training middle layers only is usually the second slowest, not only because it has to access up  to the last but one bottom layer but also the dimension of the resulting gradient, leaving the very top and very bottom out, is almost the same as the full network.

\subsection{Probabilistic Layer-Wise Sparse SGD} 
To take advantage of both the good generalization performance that training both the top and the bottom few layers exhibits and the efficiency that training only the top $k$ layers enjoys, we investigate the feasibility of combining the two through probabilistic LWS-SGD. Since training only the top $k$ layers can be fast and it is essential to access the bottom layers for deep networks to generalize well, a balanced approach may rely on an infrequent visit to the bottom layers. In particular, we propose and evaluate the following probabilistic LWS training strategies: 
\vspace*{-3mm}
\begin{enumerate}[(a)]
    \setlength\itemsep{0em}
    \item Top $k$ Bottom $q$: In each epoch, the training updates the top $k$ layers and with Bernoulli probability $\rho$ updates the bottom $q$ layers as well. 
    \item Top $k$ All Bottoms: Similar to Top $k$ Bottom $q$, but with probability $\rho$ we train all the bottom layers, so the full network gets trained with probability $\rho$.
    \item Random-Uniform: In each epoch, we use a discrete uniform distribution over $\{1,\ldots,d\}$ (depth) to sample an integer $k$ and train the top-$k$ layers.
    \item Random-Beta: In each epoch, we map a (skewed) beta distribution, e.g., $\alpha=2,\beta=5$, over the interval $[1,d]$ to sample an integer $k$ (by rounding) and train the top-$k$ layers.
\vspace*{-3mm}
\end{enumerate}

\begin{figure*}[th]
    \centering
    \subfigure[VGG-5, MNIST, 100 Epochs.]{
    \includegraphics[width=0.78\textwidth]{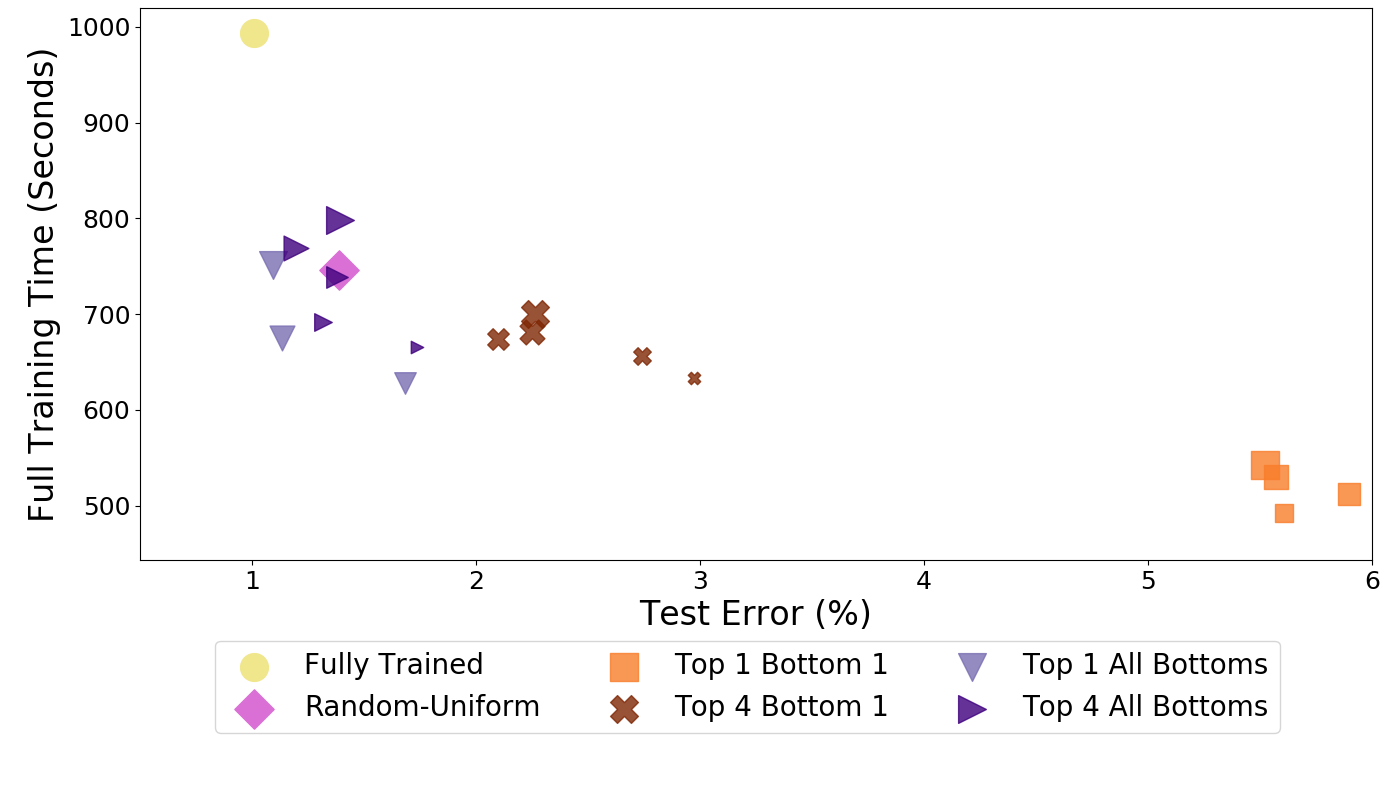}}
    \subfigure[VGG-11, CIFAR-10, 180 Epochs.]{
    \includegraphics[width=0.78\textwidth]{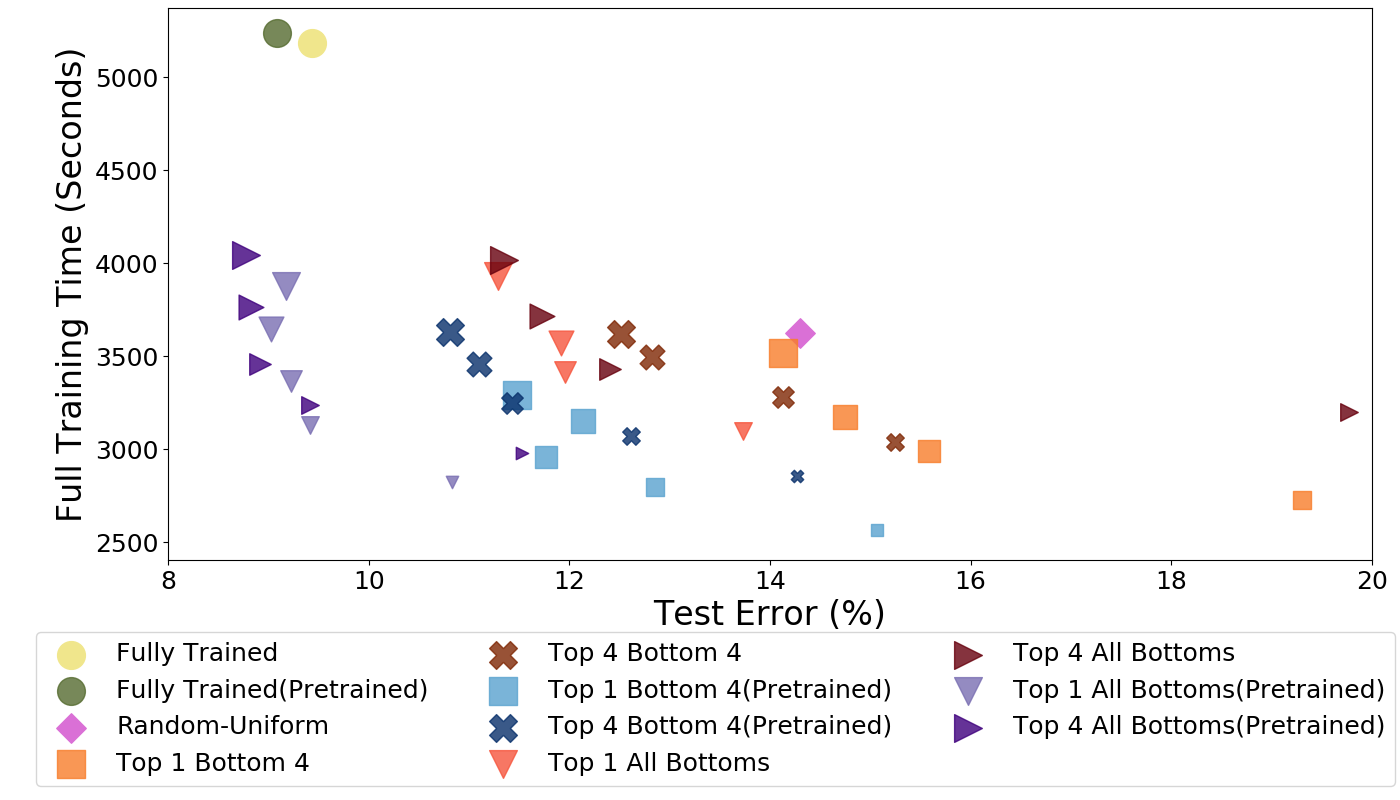}}
    \caption{Test Error($\%$) versus Full Training Time (in seconds) for networks trained by the following probabilistic LWS-SGD variants: (a) Top $k$ Bottom $q$, (b) Top $k$ All Bottoms, and (c) Random-Uniform. The full training time consists of both the time to do back-propagation, and a constant part that includes loading the mini batch, doing the forward pass, computing the loss(accuracy), etc. Probabilistic LWS-SGD could approximately reduce the overall training time by \textbf{half} comparing with than doing full back-propagation.}
    \label{fig:vgg11_cifar_adaptive_err_vs_full_time}
\end{figure*}
\clearpage

We study the generalization performance and the time efficiency of doing back-propagation for all aforementioned probabilistic LWS-SGD variants and show the comparison with vanilla SGD in Figures \ref{fig:vgg5_mnist_adaptive}, \ref{fig:vgg5_mnist_adaptive_time} and \ref{fig:vgg11_cifar_adaptive_err_vs_time}. We initialize VGG-5 with Xaiver initialization \citep{xavier} only, and initialize VGG-11 using both Xavier initialization as well as the pre-trained weights learnt from ImageNet \citep{imagenet_cvpr09,imagenet_ILSVRC15}. For simple problem as MNIST, the generalization only takes a slight hit even if the
probabilistic LWS-SGD only employs the very bottom layer ($q=1$) and utilizes a very low frequency to update the bottom layer, e.g., $\rho=0.1$. As the difficulty of the problem increases, to maintain the good generalization, we have to involve more bottom layers and increase $\rho$ (see Figure \ref{fig:vgg5_mnist_adaptive}(a) and Figure~\ref{fig:vgg11_cifar_adaptive_err_vs_time}), which leads to a higher computational cost. In practice, one can initialize the network with pre-trained weights learnt from ImageNet and
get away with lower $\rho$, i.e., less frequent updates of the bottom layers.

Overall, probabilistic LWS-SGD improves the efficiency of doing back-propagation (Figures \ref{fig:vgg5_mnist_adaptive_time} and \ref{fig:vgg11_cifar_adaptive_err_vs_time} for details).
In particular, Top $k$ All Bottoms works the best with a suitable selection of $k$. It matches the generalization achieved by fully trained model, but is typically much faster. The benefit of the right probabilistic LWS-SGD training is that it reduces the time to perform back-propagation typically by 2-5 times, while keeping the generalization largely unchanged, and even slightly improving it in some cases (purple triangles in Figure~\ref{fig:vgg11_cifar_adaptive_err_vs_time} (b)).

{\bf Discussion.} Probabilistic LWS-SGD only directly reduce the time to perform back-propagation, not necessarily the overall training time. The overall training time not only consists of the time to do back-propagation, but also a constant part that includes loading the mini batch, doing the forward pass, computing the loss(accuracy), etc. As Figure~\ref{fig:vgg11_cifar_adaptive_err_vs_full_time} demonstrates, even though the back-propagation time can be up to 5x more efficient (Figure~\ref{fig:vgg11_cifar_adaptive_err_vs_time}), LWS-SGD could only reduce the overall training time by half. At a high level, our results indicate that one may be able to use the implicit bias of SGD-type algorithms towards sparse solutions~\citep{gunasekar2018conv,wogl2019}, especially the layer-wise structure of the sparsity, to develop faster training algorithms with sparse gradients which reach a similar solution, with similar generalization behavior.

\section{Conclusion}
In this work, we empirically investigate generalization and optimization behavior of deep networks trained in the rich regime. Using the notions of $\gamma$ Active-re-initialization and $\varepsilon$ Lazy-re-initialization, we illustrate that there are active parameters which move more substantially during training, and re-initializing such active parameters leads to significant reduction in generalization. The active parameters primarily live in the bottom layers for wider networks. Based on such observation, we study both static and probabilistic LWS-SGD algorithms which update only subsets of layers. Experimental results on MNIST and CIFAR-10 demonstrate that suitable probabilistic LWS-SGD not only matches the generalization performance of vanilla SGD but also substantially speeds up the back-propagation phase. Such results can potentially be used to design faster algorithms for training without adversely affecting generalization.

\label{sec:conclude}

\section*{Acknowledgement}
The research was supported by NSF grants IIS-1908104, OAC-1934634, IIS-1563950, IIS-1447566, IIS-1447574, IIS-1422557, CCF-1451986. The authors would like to thank Minnesota Supercomputing Institute (MSI) at the University of Minnesota for providing the computing support.

\bibliographystyle{plainnat}
\bibliography{reference}

\newpage
\appendix
\section{Additional Experimental Results}
\label{sec:app_exp}

\begin{figure*}[th]
    \centering
    \subfigure[Relu Network, d=1 (2 layers), Fashion-MNIST.]{
    \includegraphics[width=0.88\textwidth]{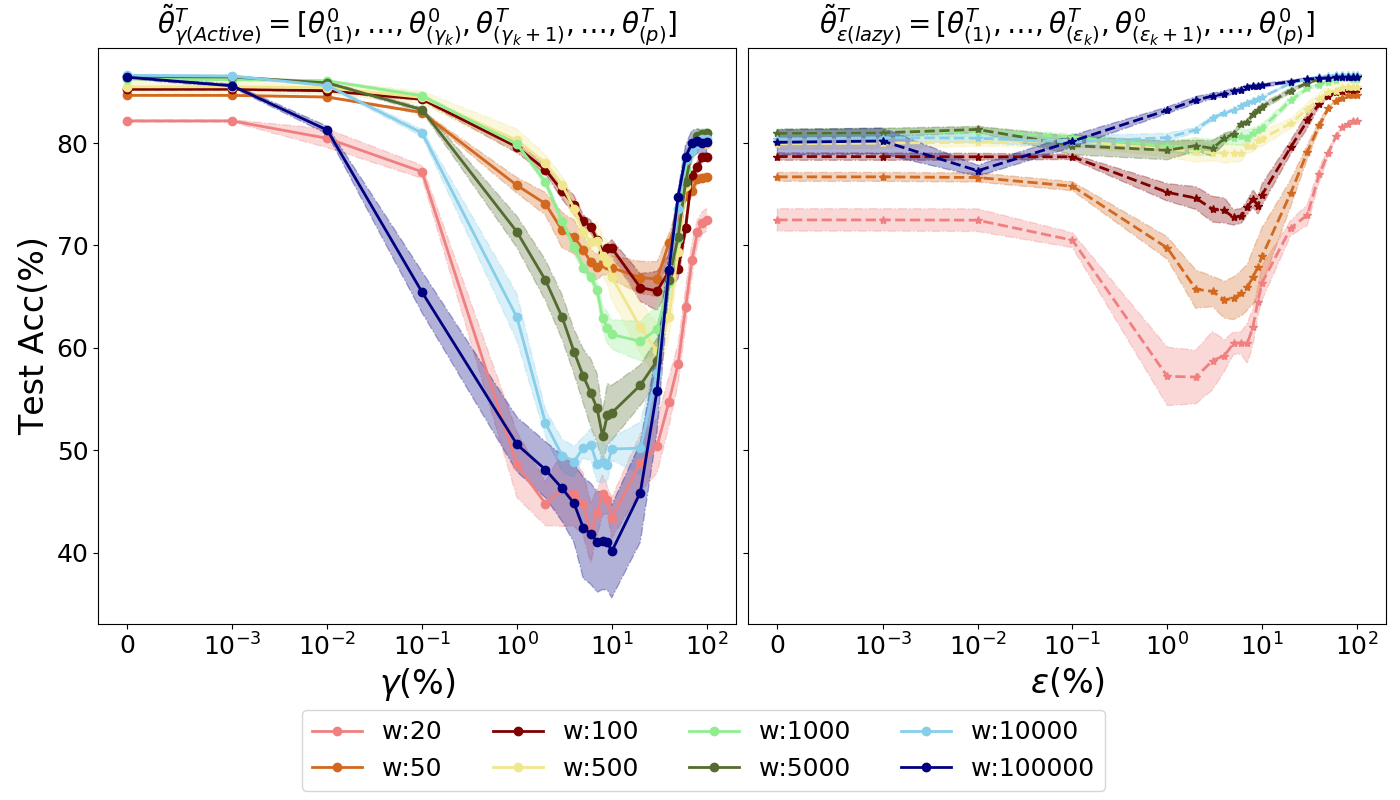}}
    \subfigure[Relu Network, d=4 (5 layers), Fashion-MNIST.]{
    \includegraphics[width=0.88\textwidth]{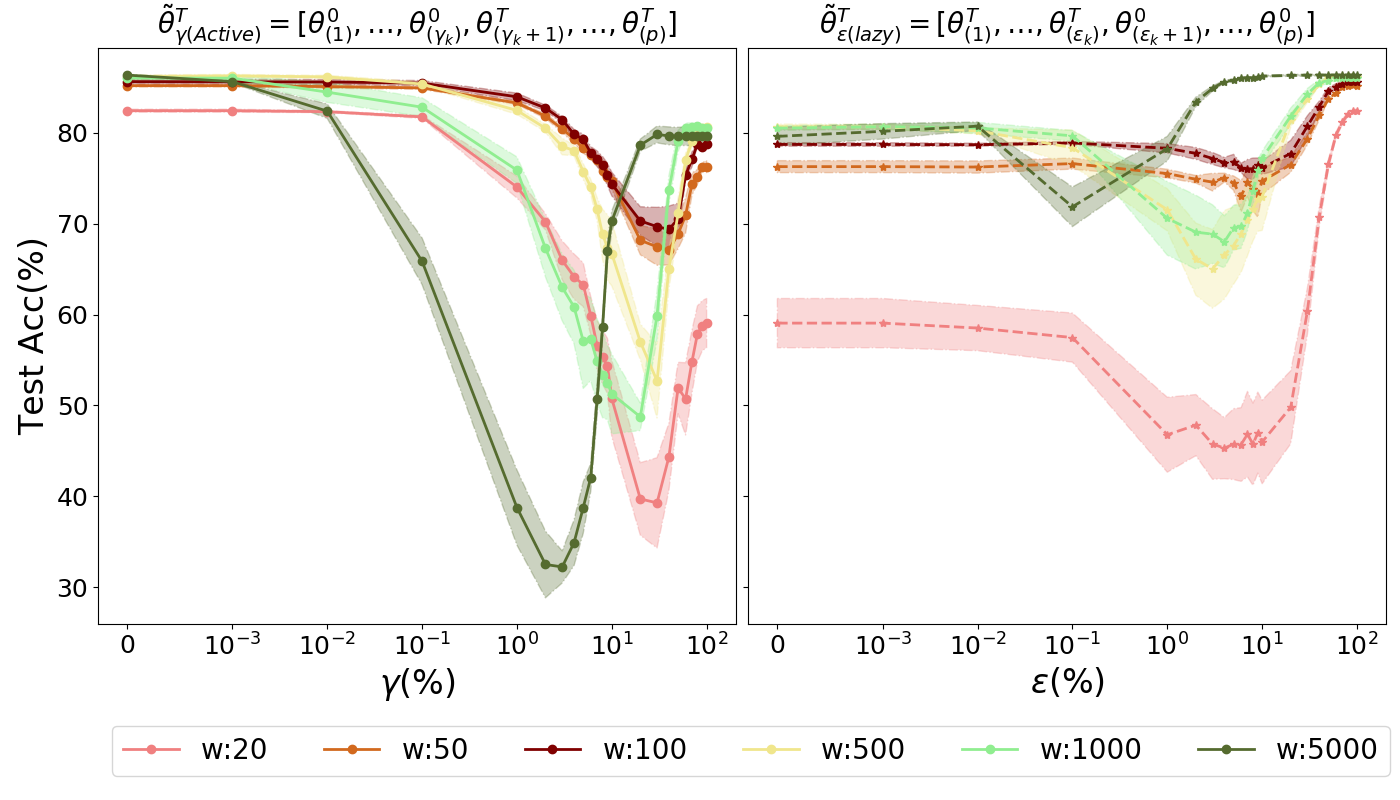}}
    \caption{Test accuracy of ReLU-Nets trained on \textbf{Fashin-MNIST} as a function of the choice of $\gamma$ and $\varepsilon$ after applying $\gamma$ Active-re-initialization (left) and $\varepsilon$ Lazy-re-initialization (right) . The x-axis is in log scale. Similar to the observations made on MNIST, as $\gamma$ increases, the test performance plunges at first then increases. The worse performance still happens when around $10\%$ of the active parameters are re-initialized, indicating those active parameters may be of greater importance. As $\varepsilon$ increases, the test performance gradually improves. With $w=100,000$, keeping approximately $10\%$ of parameters at their trained value is sufficient to recover the performance of the fully trained network($\varepsilon= 100\%$).
    }
     \label{fig:relu_fashion_reinit}
\end{figure*}

\begin{figure*}[t]
    \centering
    \subfigure[Relu Network, d=1 (2 layers), Fashion-MNIST.]{
    \includegraphics[width=0.88\textwidth]{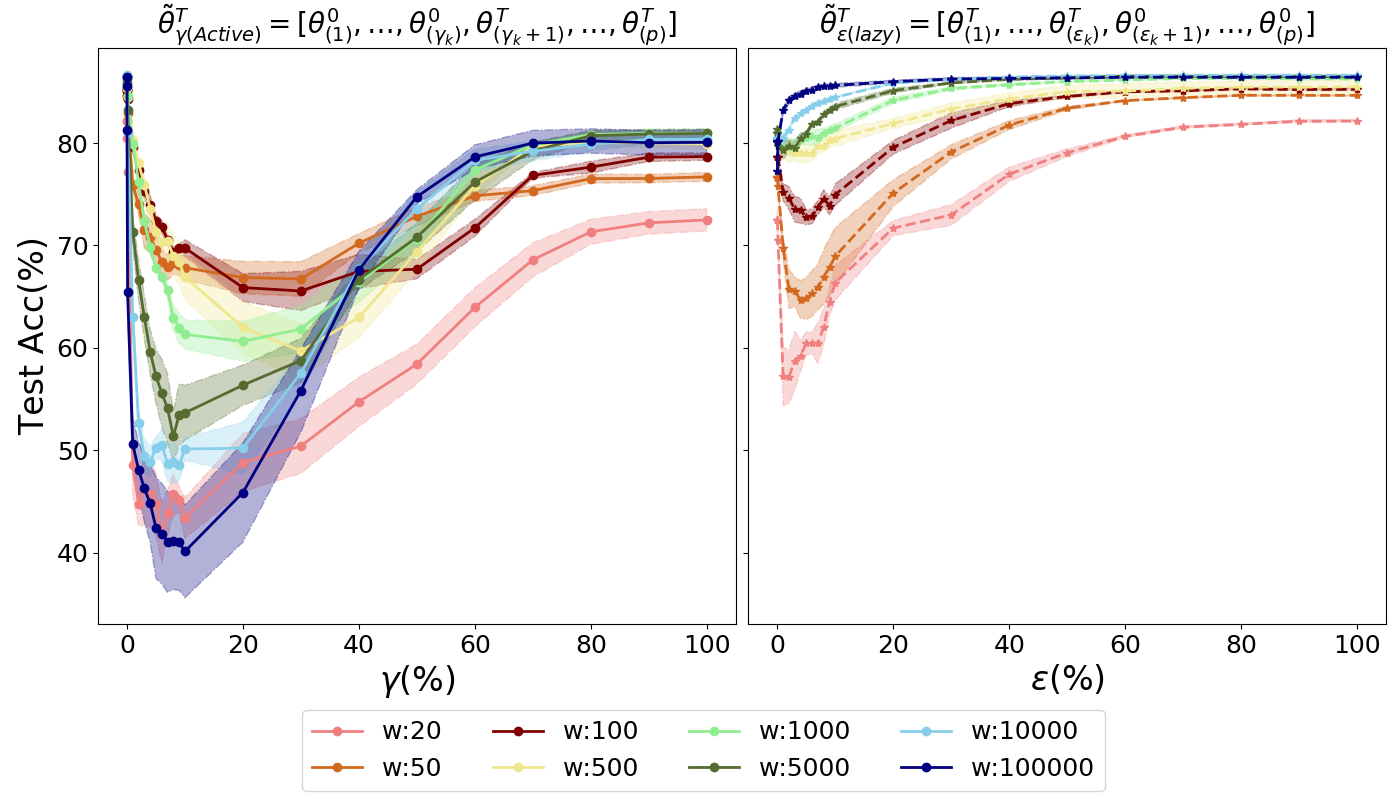}}
    \subfigure[Relu Network, d=4 (5 layers), Fashion-MNIST.]{
    \includegraphics[width=0.88\textwidth]{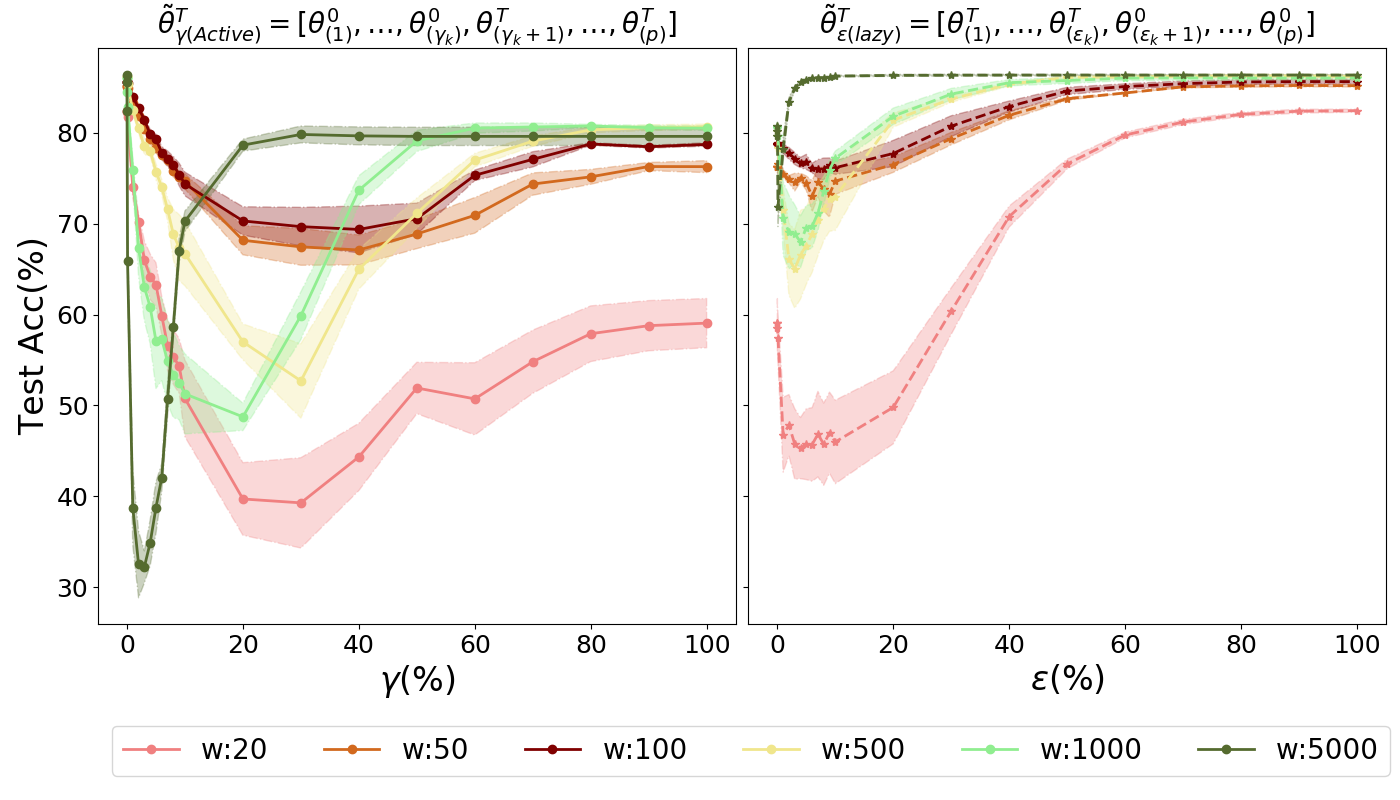}}
    \caption{The effect of width on the test accuracy after $\gamma$ Active-re-initialization (left) and $\varepsilon$ Lazy-re-initialization (right) for ReLU-Nets with $d=1$ and $d=4$, trained on \textbf{Fashion-MNIST}. The x-axis ($\gamma$ and $\varepsilon$) is in linear scale.}
     \label{fig:relu_fashion_reinit_d1_linx}
     \vspace{-5mm}
\end{figure*}

\begin{figure*}[t]
    \centering
    \subfigure[ReLU-Net, d=1 (2 layers), Fashion-MNIST.]{
    \includegraphics[width=0.88\textwidth]{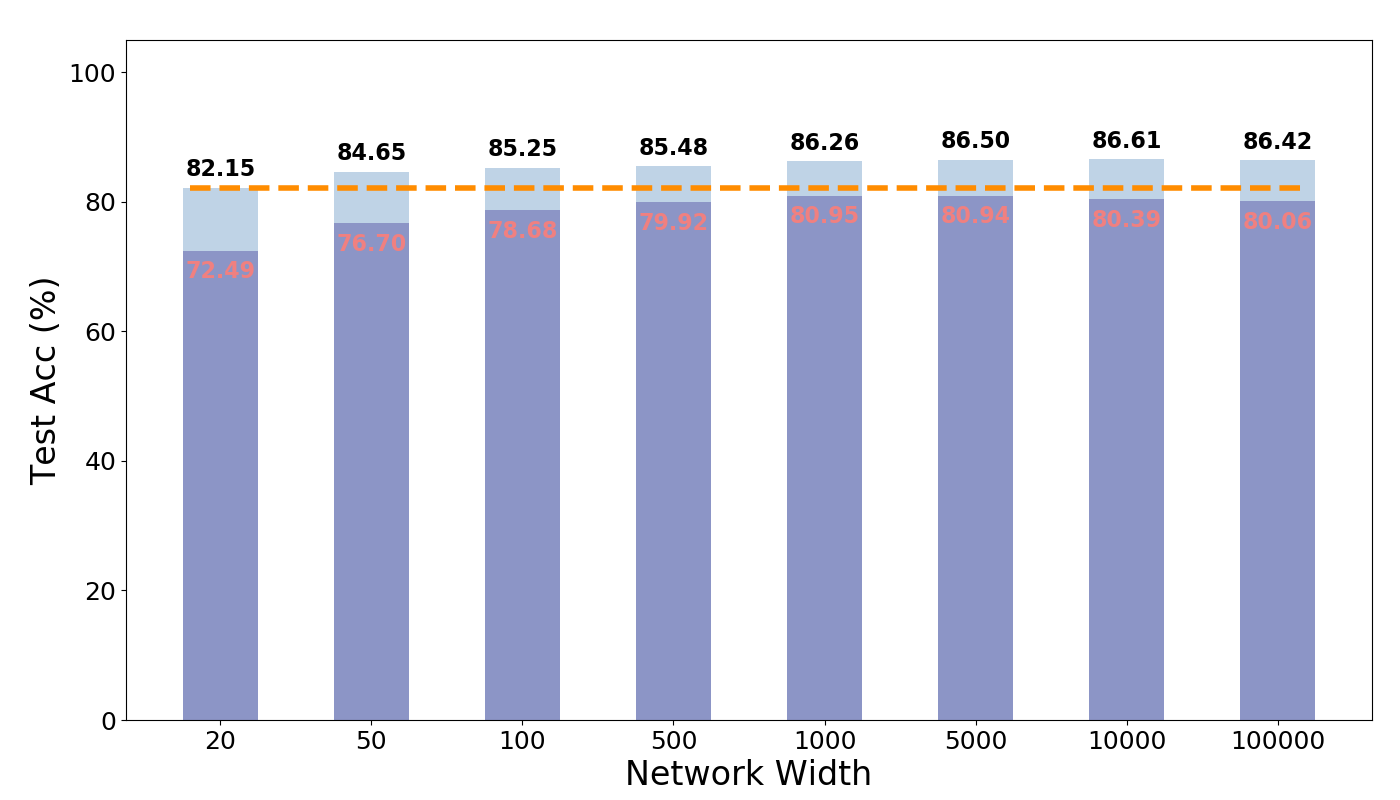}}
    \subfigure[ReLU-Net, d=4 (5 layers), Fashion-MNIST.]{
    \includegraphics[width=0.88\textwidth]{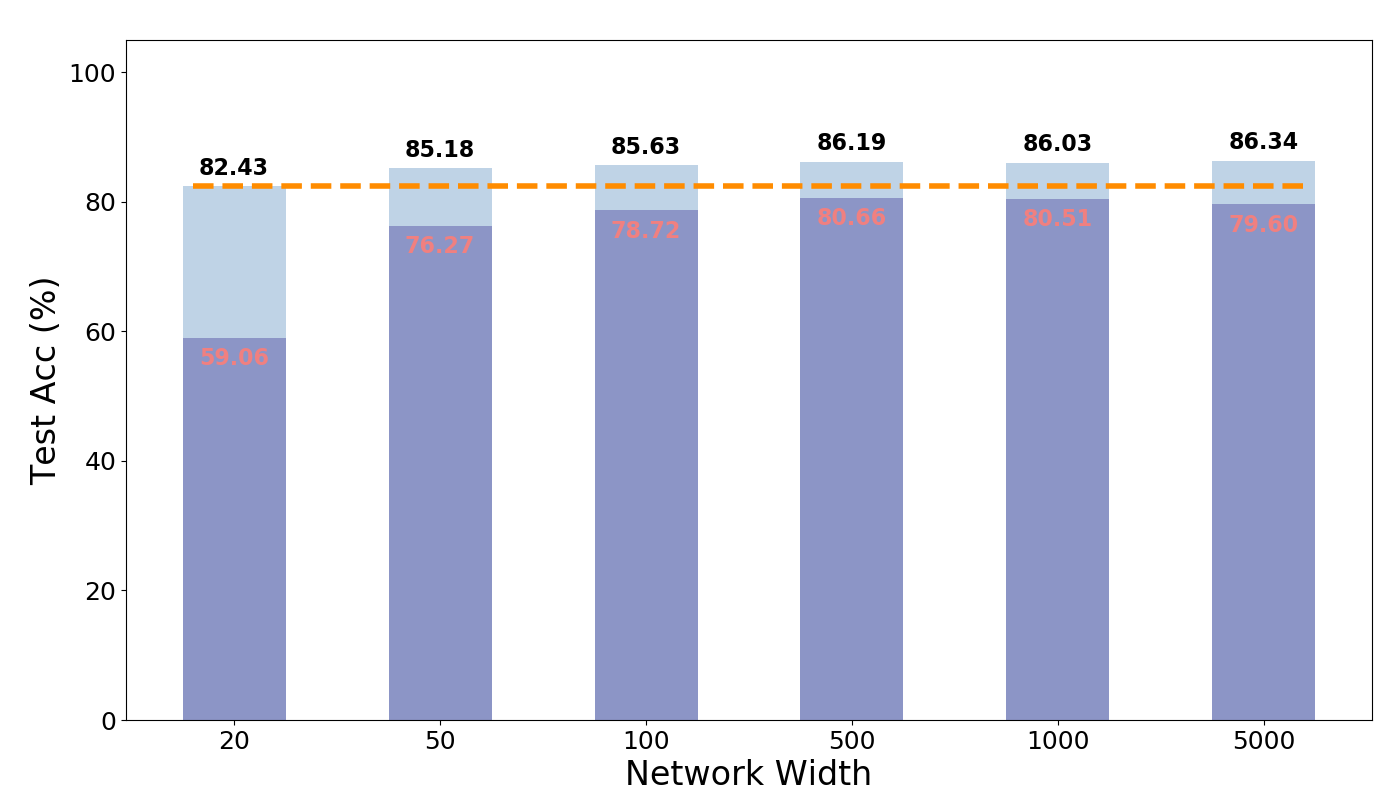}}
    \vspace{-5mm}
    \caption{Test accuracy ($\%$) at Xavier random initialization and convergence as we increase the width. ReLU-Nets are trained on \textbf{Fashion-MNIST}. The dark blue bar is the test accuracy at Xavier random initialization of which the value is shown in red. The light blue bar shows the performance difference/gap between initialization and convergence. The final test accuracy at convergence is shown in black. Increasing width improves the generalization at random initialization. However, training always helps. A trained network with w=20 can outperform a random network with w=100,000. }
     \label{fig:relu_fashion_init_vs_final}
\end{figure*}

\begin{figure*}[t]
    \centering
    \subfigure[$d=4, w=10, \alpha=1\%$.]{
    \includegraphics[width=0.48\textwidth]{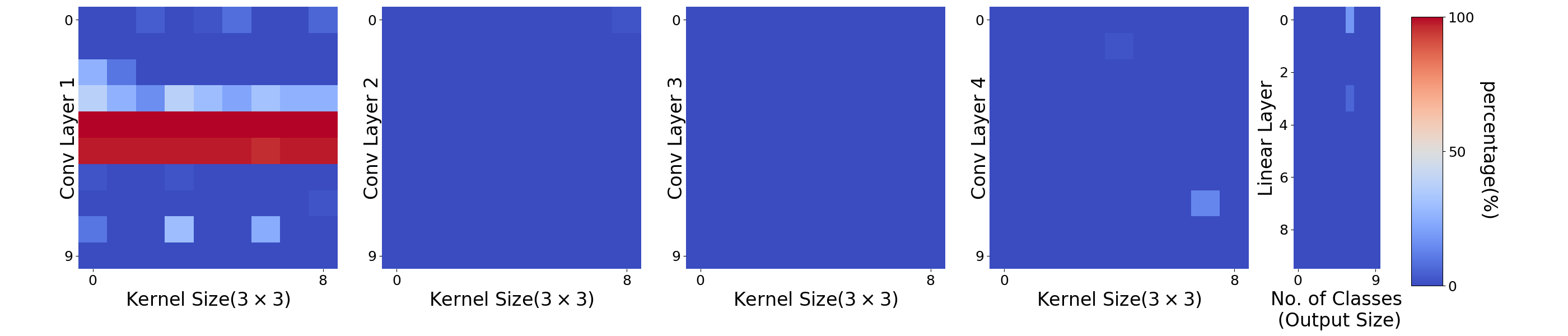}}
    \subfigure[$d=4, w=100, \alpha=1\%$.]{
    \includegraphics[width=0.48\textwidth]{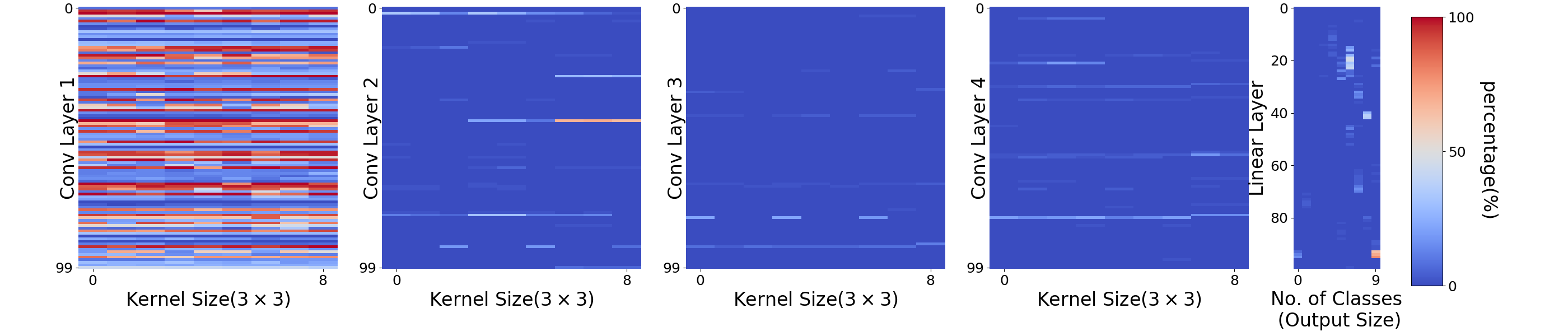}}
    \subfigure[$d=4, w=10, \alpha=10\%$.]{
    \includegraphics[width=0.48\textwidth]{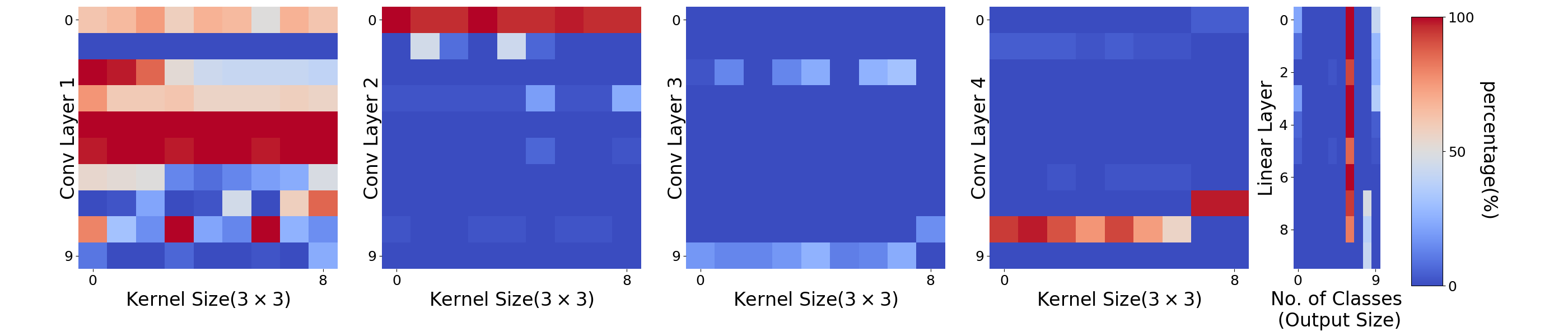}}
    \subfigure[$d=4, w=100, \alpha=10\%$.]{
    \includegraphics[width=0.48\textwidth]{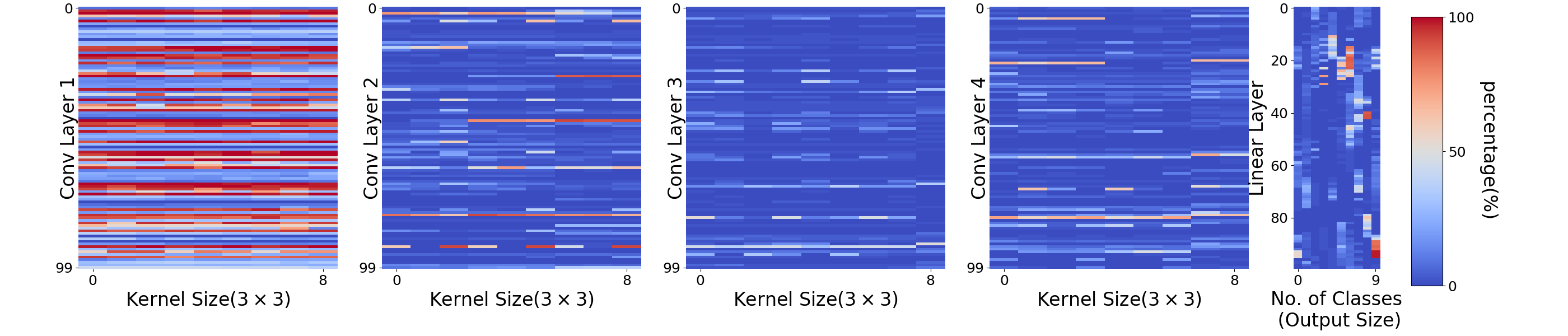}}
    \subfigure[$d=4, w=10, \alpha=30\%$.]{
    \includegraphics[width=0.48\textwidth]{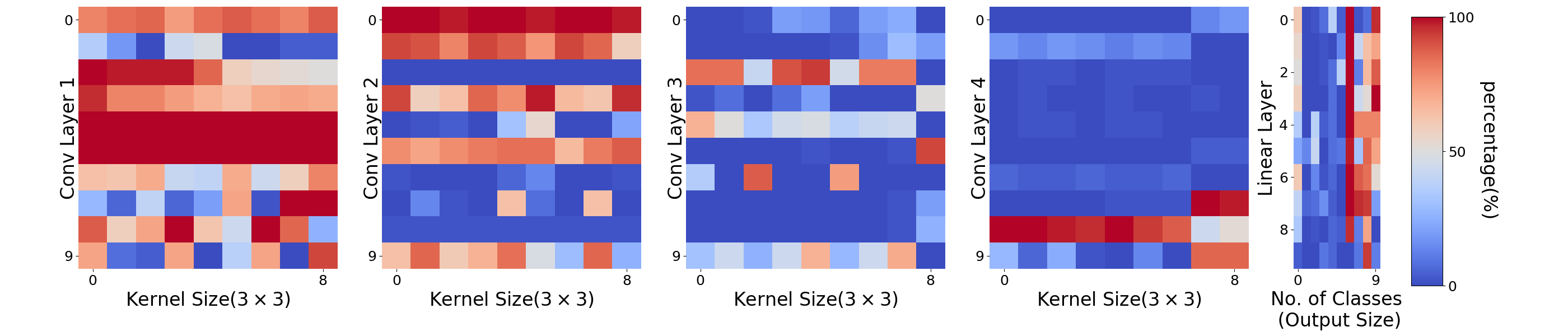}} 
    \subfigure[$d=4, w=100, \alpha=30\%$.]{
    \includegraphics[width=0.48\textwidth]{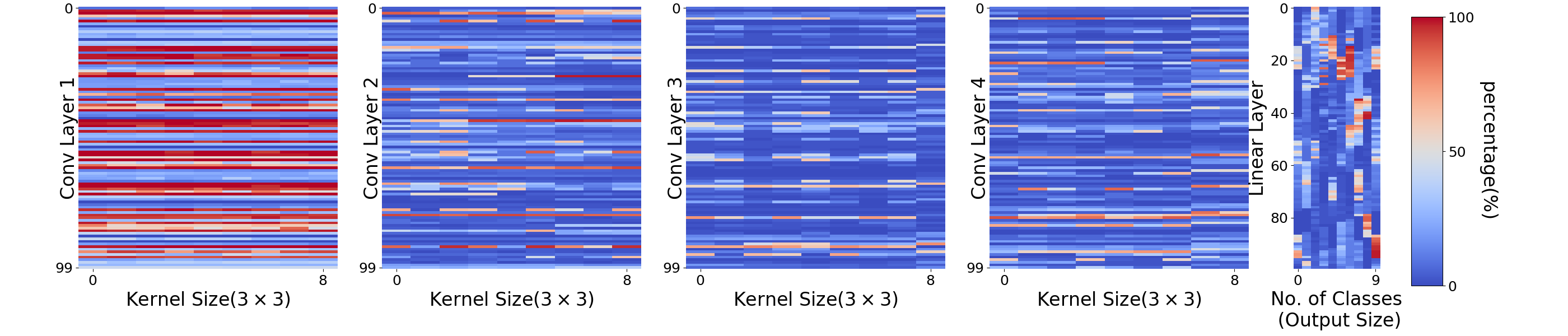}}
    \vspace{-2mm}
    \caption{Frequency of $\theta_i$ been selected as active during training for different choice of $\alpha$ for Conv-Nets ($d=4$) trained on \textbf{MNIST}. The results are the average over 5 repetitive runs.
    Each rectangle consists of parameters of a $3\times3$ kernel at layer $l_i$. 
    Red indicates high frequency, close to $100\%$ and blue means low frequency, close to $0$. Similar to what we have observed on Conv-Nets with $d=2$, as we increase the width, the active parameters in a Conv-Net concentrate at the very bottom layer.
    }
    \vspace{-4mm}
     \label{fig:vgg4_mnist_active}
\end{figure*}

\begin{figure*}[t]
    \centering
    \subfigure[$d=4, w=100, \alpha=1\%$.]{
    \includegraphics[width=0.48\textwidth]{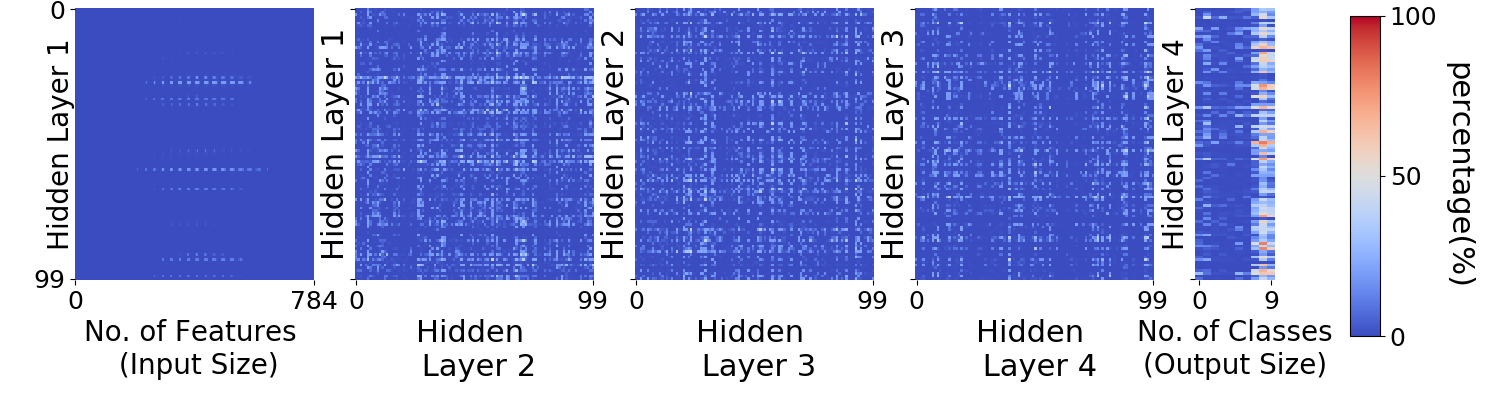}}
    \subfigure[$d=4, w=1000, \alpha=1\%$.]{
    \includegraphics[width=0.48\textwidth]{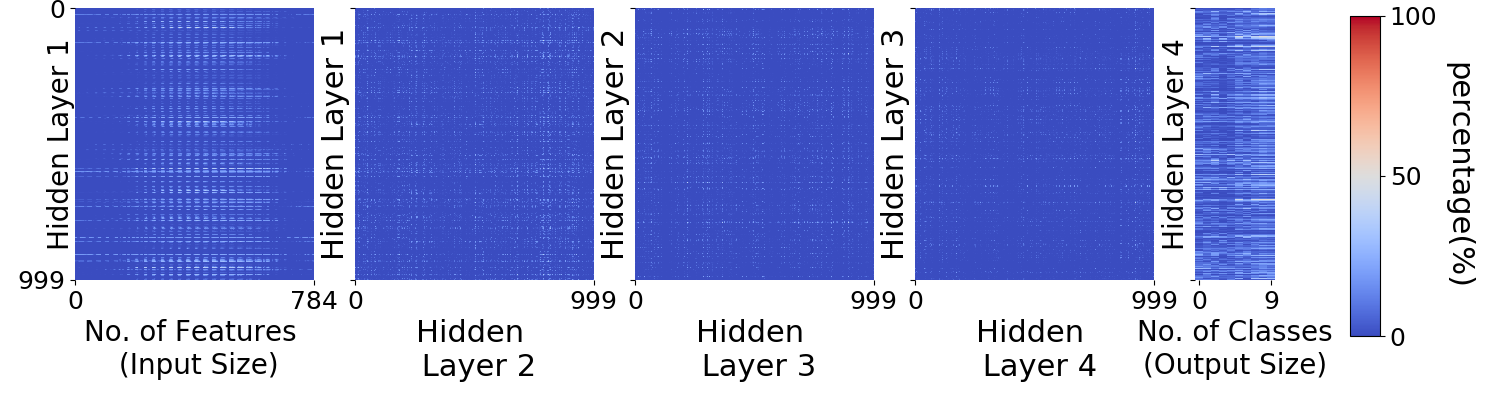}}
   \subfigure[$d=4, w=100, \alpha=10\%$.]{
    \includegraphics[width=0.48\textwidth]{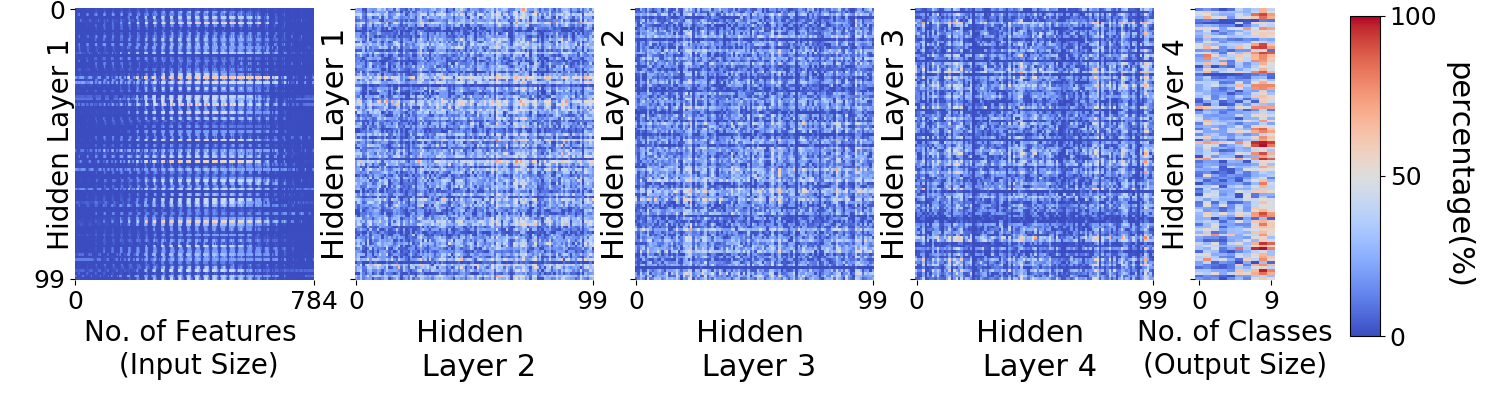}}
    \subfigure[$d=4, w=1000, \alpha=10\%$.]{
    \includegraphics[width=0.48\textwidth]{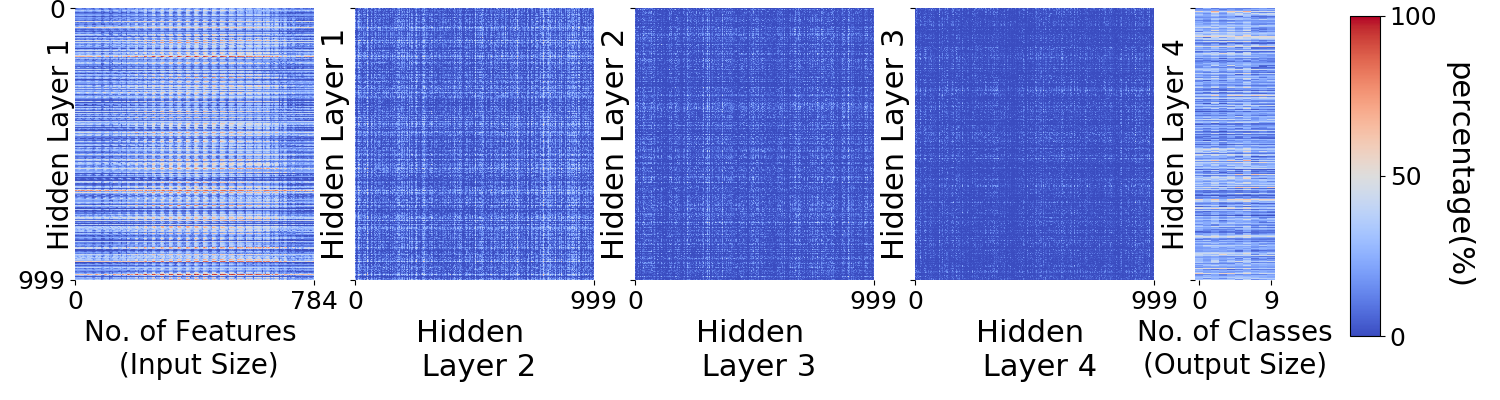}}
    \subfigure[$d=4, w=100, \alpha=30\%$.]{
    \includegraphics[width=0.48\textwidth]{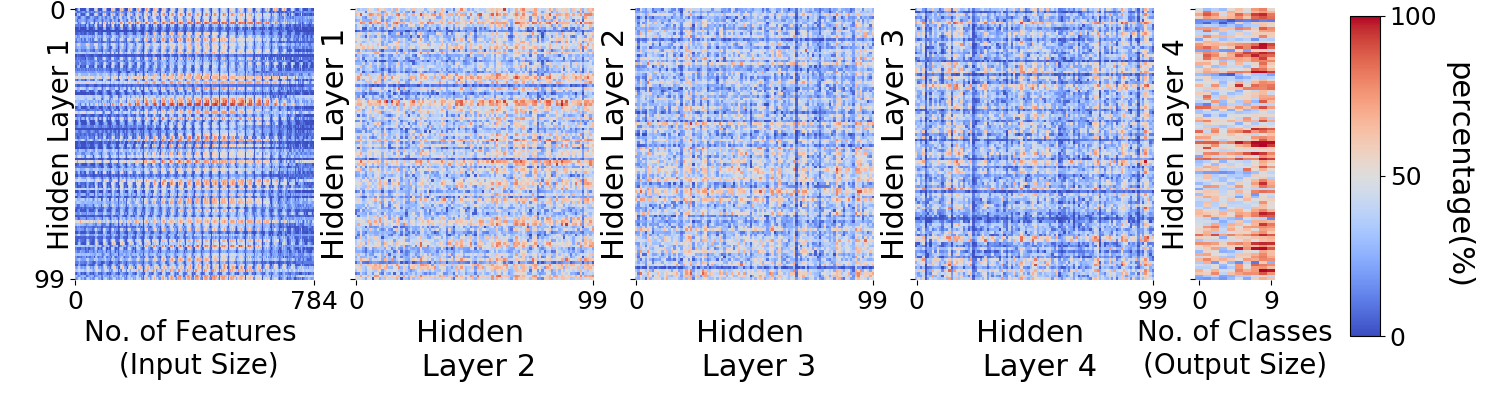}}   
    \subfigure[$d=4, w=1000, \alpha=30\%$.]{
    \includegraphics[width=0.48\textwidth]{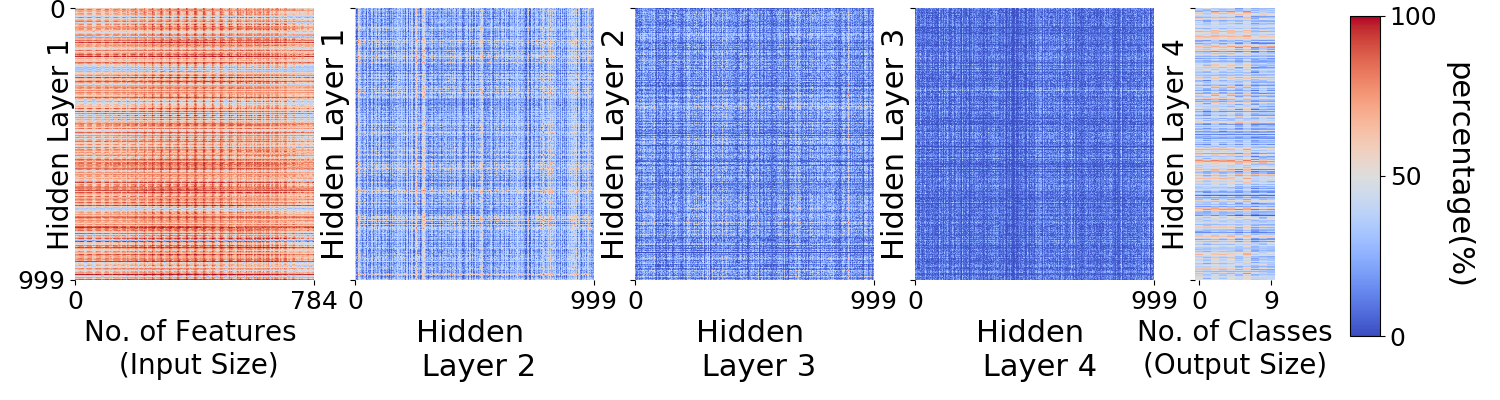}}
    \vspace{-2mm}
    \caption{Frequency of $\theta_i$ been selected as active during training with different choice of $\alpha$ for 5-Layer ReLU-Nets ($d=4$) trained on \textbf{Fashion-MNIST}. The results are the average over 5 repetitive runs. Each rectangle consists of parameters connecting layer $l_i$ and $l_{i-1}$. Red indicates a high frequency, close to $100\%$ and blue means low frequency, close to $0$. When $w$ is small, active parameters are spread across all layers. Similar to what has been observed on MNIST, as we increase the width, active parameters in the ReLU-Net are concentrated to both the very top and the very bottom layer.
    }
    \vspace{-3mm}
     \label{fig:relu_fashion_active}
\end{figure*}

\begin{figure*}[t]
    \centering
    \subfigure[$d=2, w=10, \alpha=1\%$.]{
    \includegraphics[width=0.48\textwidth]{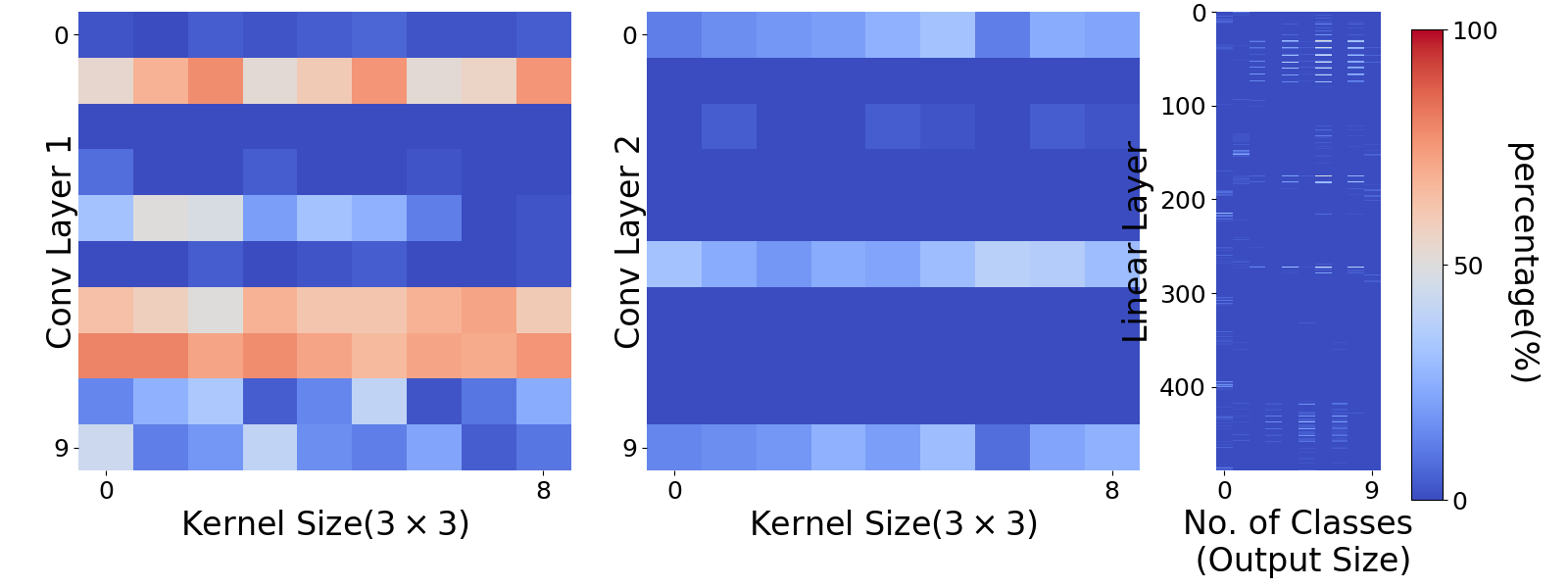}}
    \subfigure[$d=2, w=100, \alpha=1\%$.]{
    \includegraphics[width=0.48\textwidth]{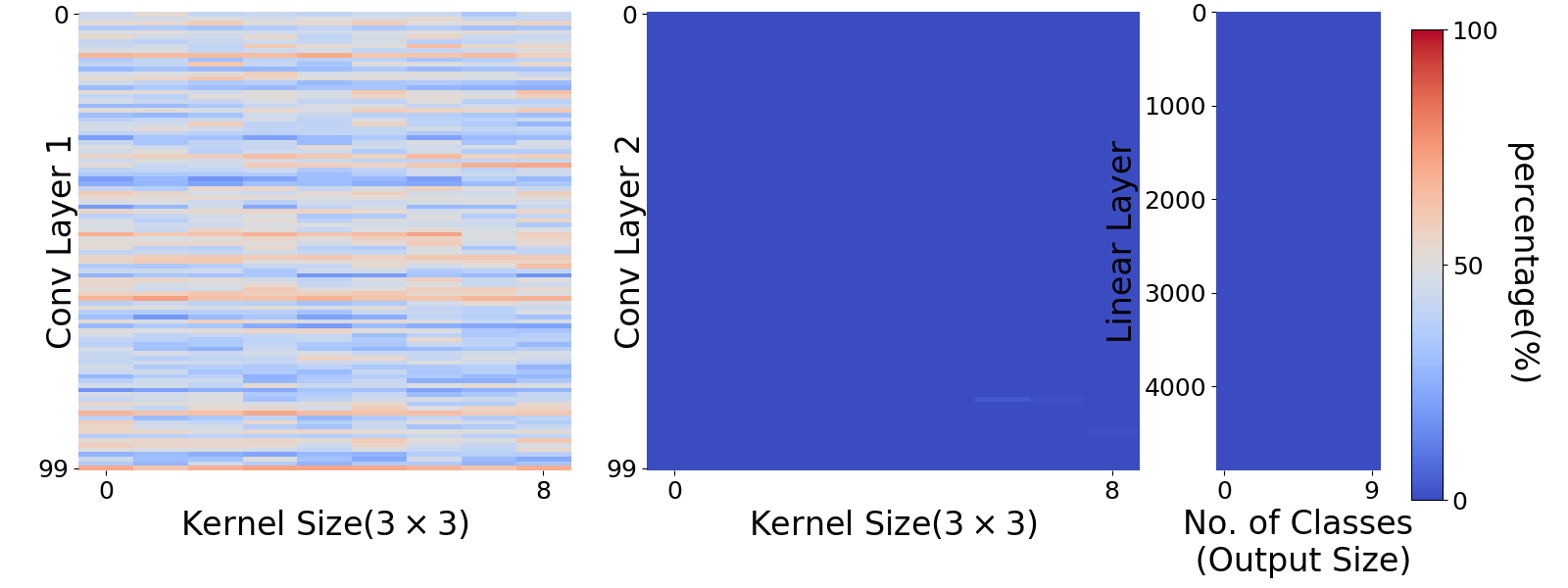}}
    \subfigure[$d=2, w=10, \alpha=10\%$.]{
    \includegraphics[width=0.48\textwidth]{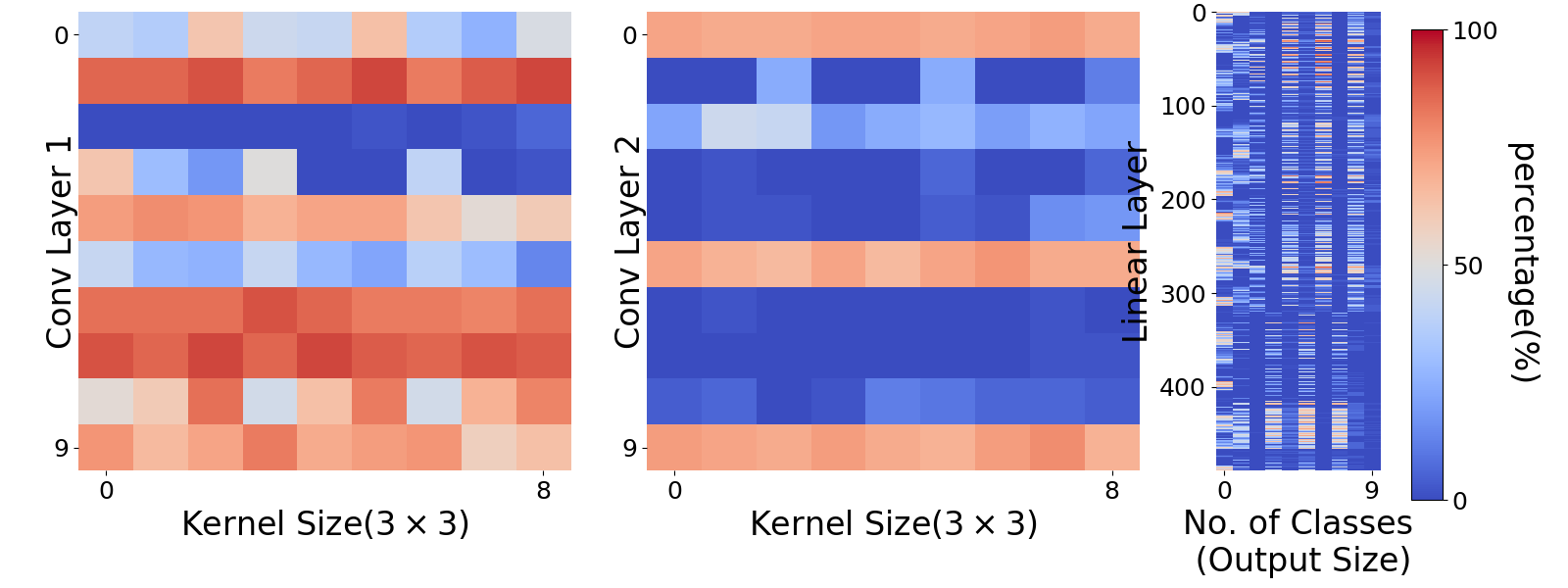}}
    \subfigure[$d=2, w=100, \alpha=10\%$.]{
    \includegraphics[width=0.48\textwidth]{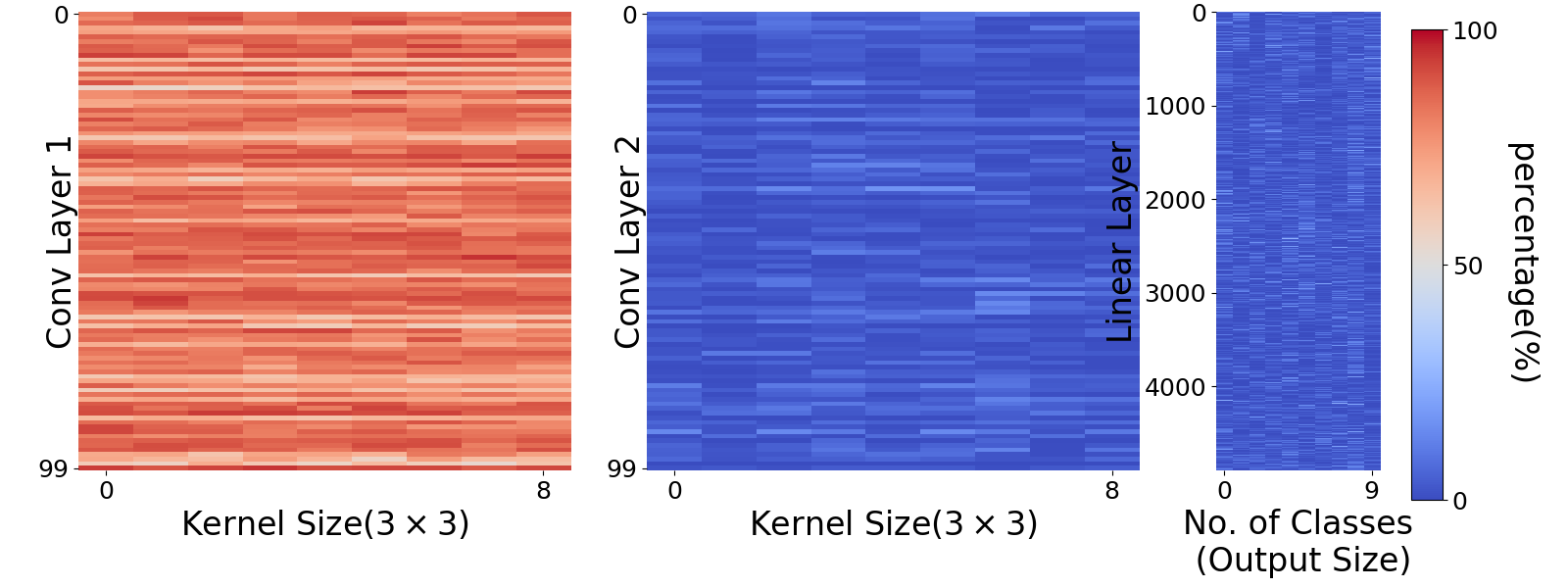}}
    \subfigure[$d=2, w=10, \alpha=30\%$.]{
    \includegraphics[width=0.48\textwidth]{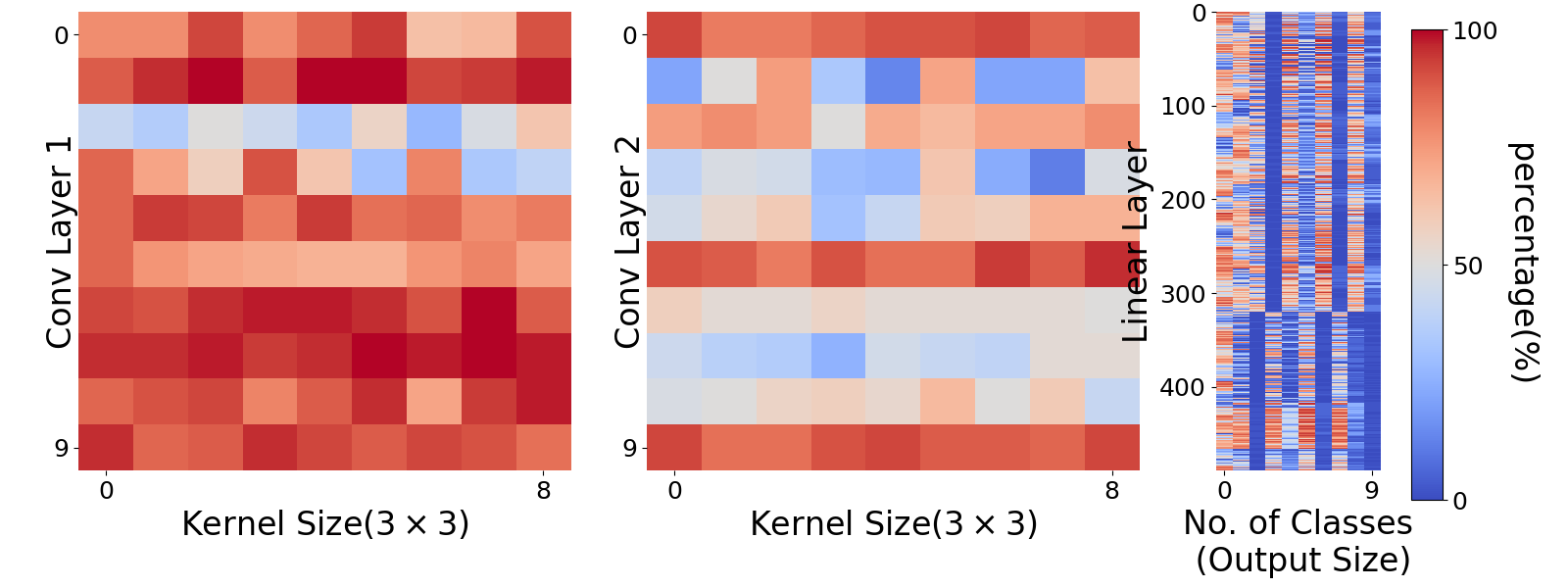}} 
    \subfigure[$d=2, w=100, \alpha=30\%$.]{
    \includegraphics[width=0.48\textwidth]{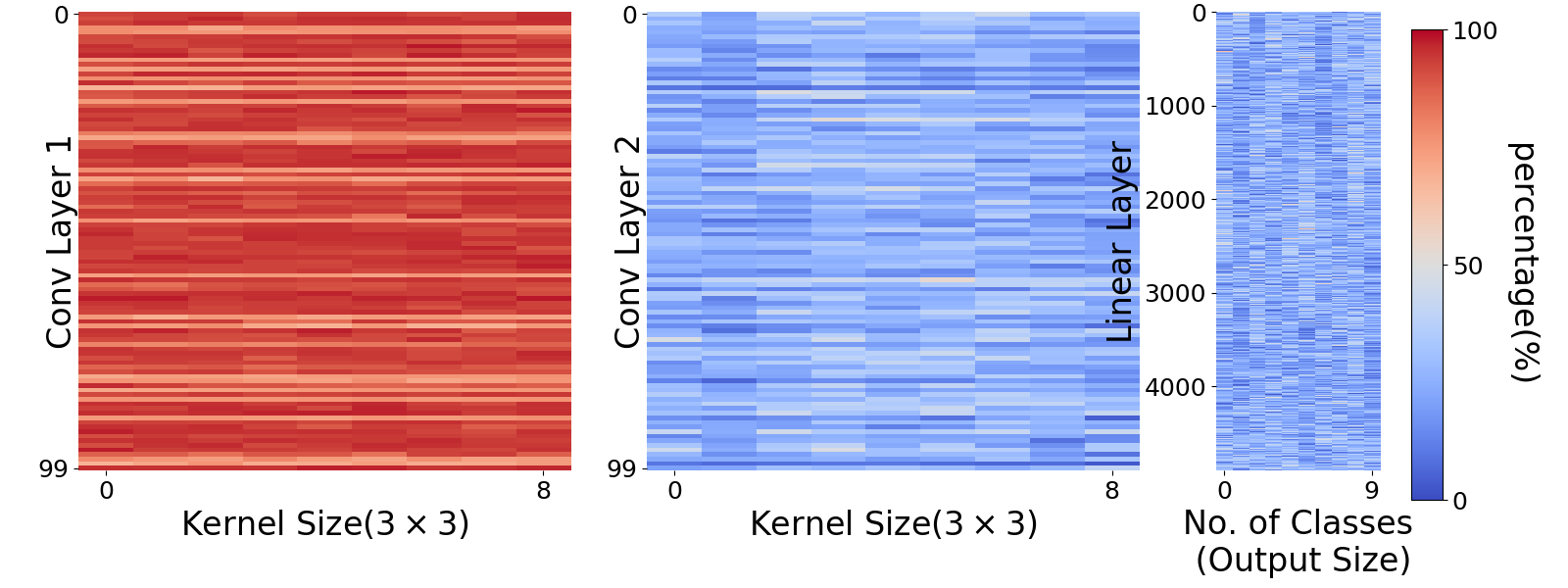}}
    \caption{Frequency of $\theta_i$ been selected as active during training for different choice of $\alpha$ for 3-Layer Conv-Nets ($d=2$ ) trained on \textbf{Fashion-MNIST}. The results are the average over 5 repetitive runs.
    Each rectangle consists of parameters of a $3\times3$ kernel at layer $l_i$. Red indicates high frequency, close to $100\%$ and blue means low frequency, close to $0$. Similar to the observations made on MNIST, as we increase the width, the active parameters in a Conv-Net concentrate at the very bottom layer.
    }
    \vspace{-4mm}
     \label{fig:vgg2_fashion_active}
\end{figure*}

\begin{figure*}[t]
    \centering
    \subfigure[$d=4, w=10, \alpha=1\%$.]{
    \includegraphics[width=0.48\textwidth]{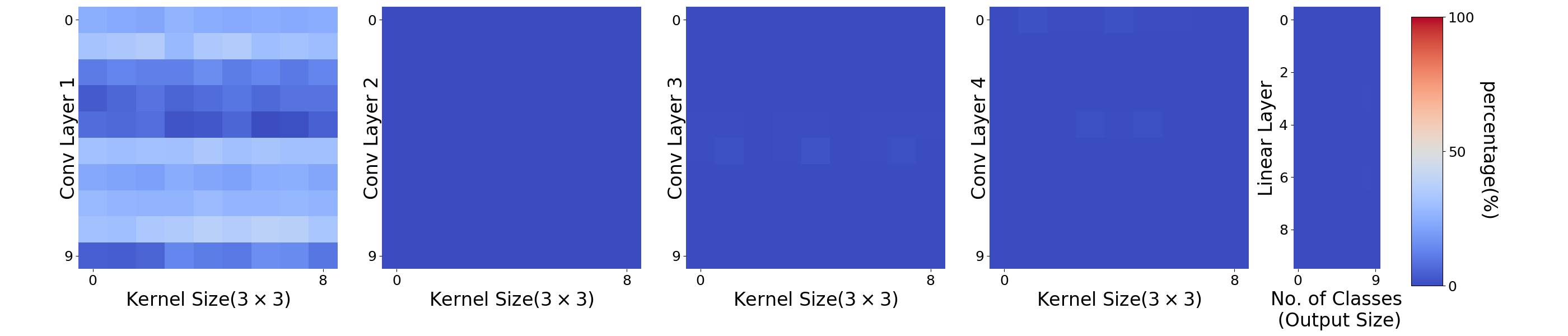}}
    \subfigure[$d=4, w=100, \alpha=1\%$.]{
    \includegraphics[width=0.48\textwidth]{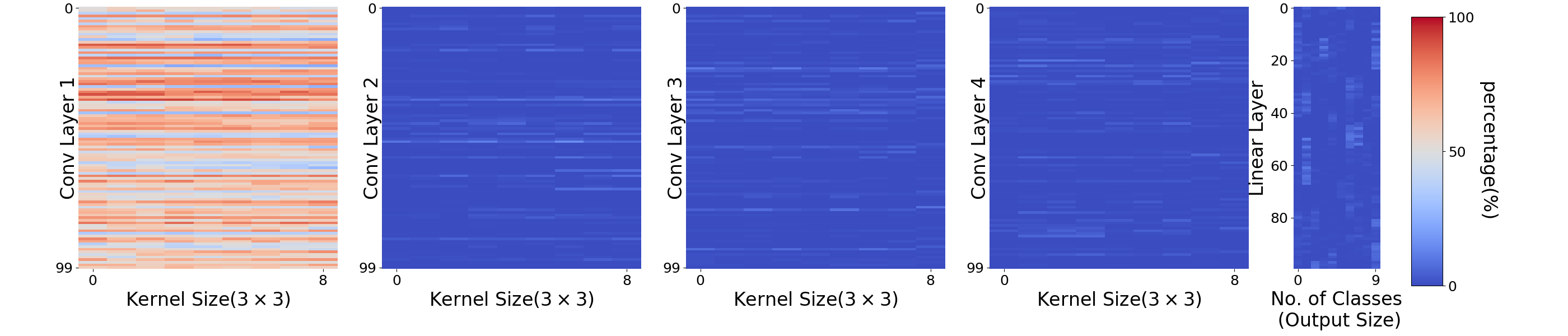}}
    \subfigure[$d=4, w=10, \alpha=10\%$.]{
    \includegraphics[width=0.48\textwidth]{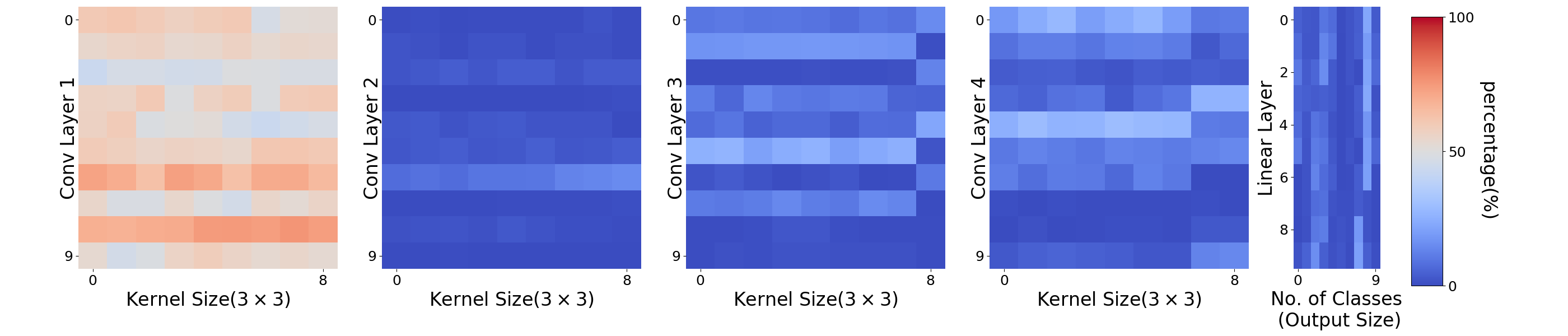}}
    \subfigure[$d=4, w=100, \alpha=10\%$.]{
    \includegraphics[width=0.48\textwidth]{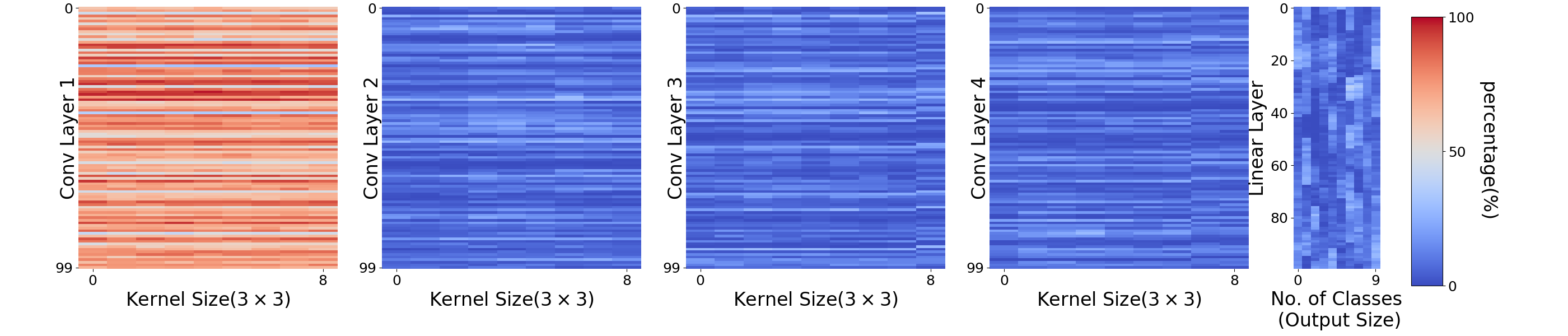}}
    \subfigure[$d=4, w=10, \alpha=30\%$.]{
    \includegraphics[width=0.48\textwidth]{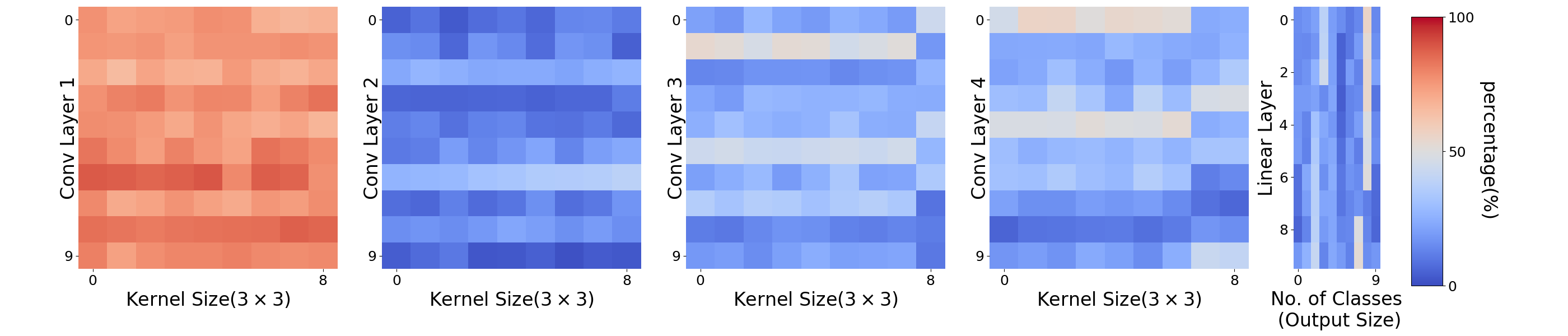}} 
    \subfigure[$d=4, w=100, \alpha=30\%$.]{
    \includegraphics[width=0.48\textwidth]{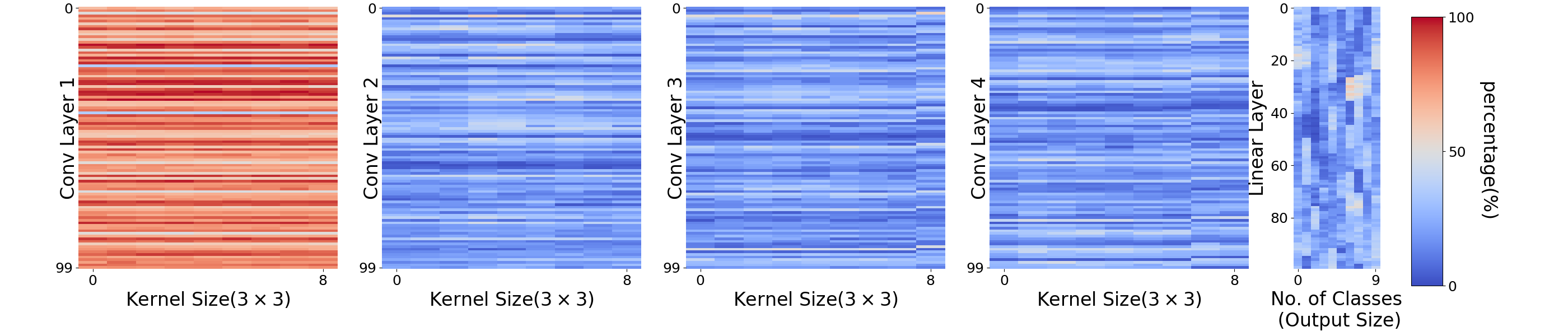}}
    \vspace{-2mm}
    \caption{Frequency of $\theta_i$ been selected as active during training for different choice of $\alpha$ for 5-Layer Conv-Nets ($d=4$) trained on \textbf{Fashion-MNIST}. The results are the average over 5 repetitive runs.
    Each rectangle consists of parameters of a $3\times3$ kernel at layer $l_i$. 
    Red indicates high frequency, close to $100\%$ and blue means low frequency, close to $0$. Similar to the observations made on MNIST, as we increase the width, the active parameters in a Conv-Net concentrate at the very bottom layer.
    }
     \label{fig:vgg4_fashion_active}
\end{figure*}

\begin{figure*}[t]
    \centering
    \subfigure[Test Acc.]{
    \includegraphics[width=0.98\textwidth]{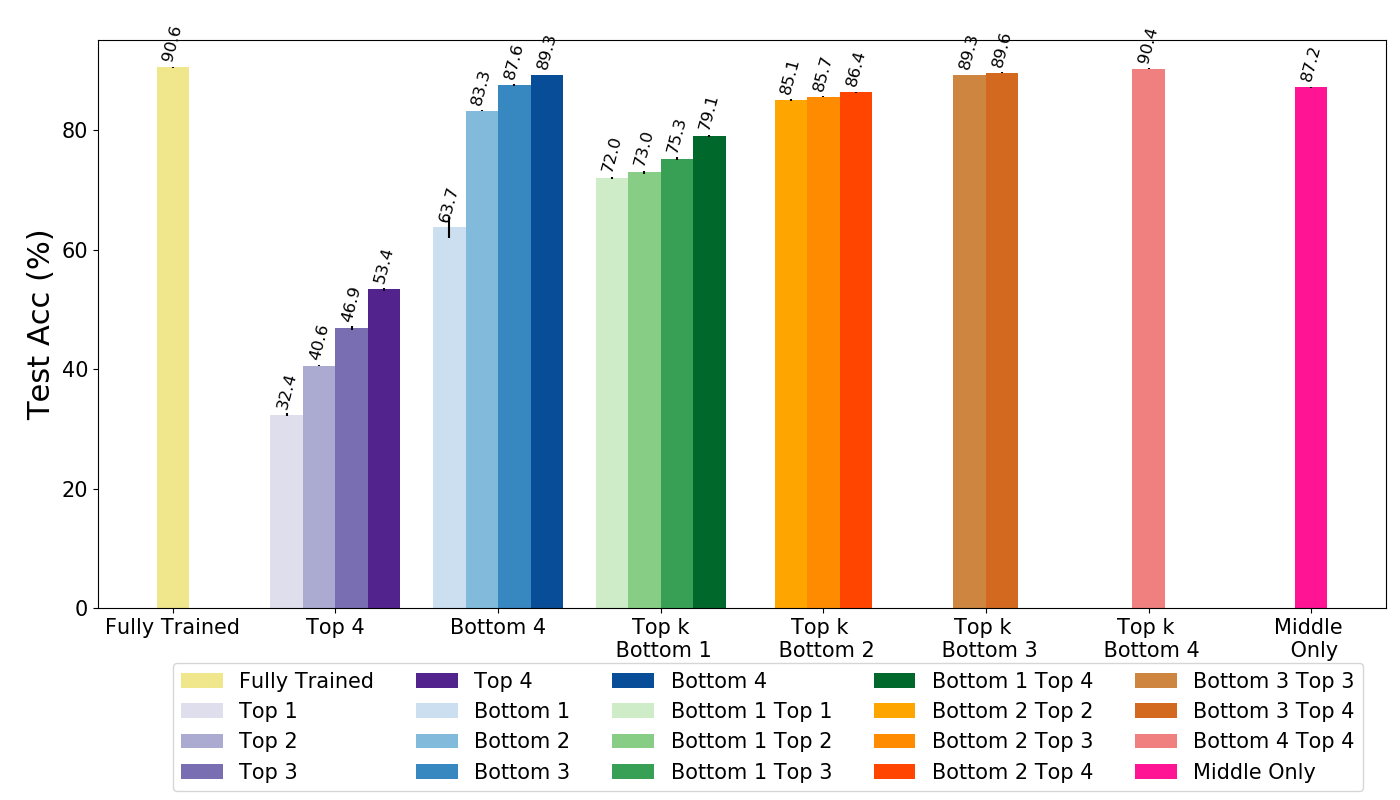}}
    \subfigure[Time.]{
    \includegraphics[width=0.98\textwidth]{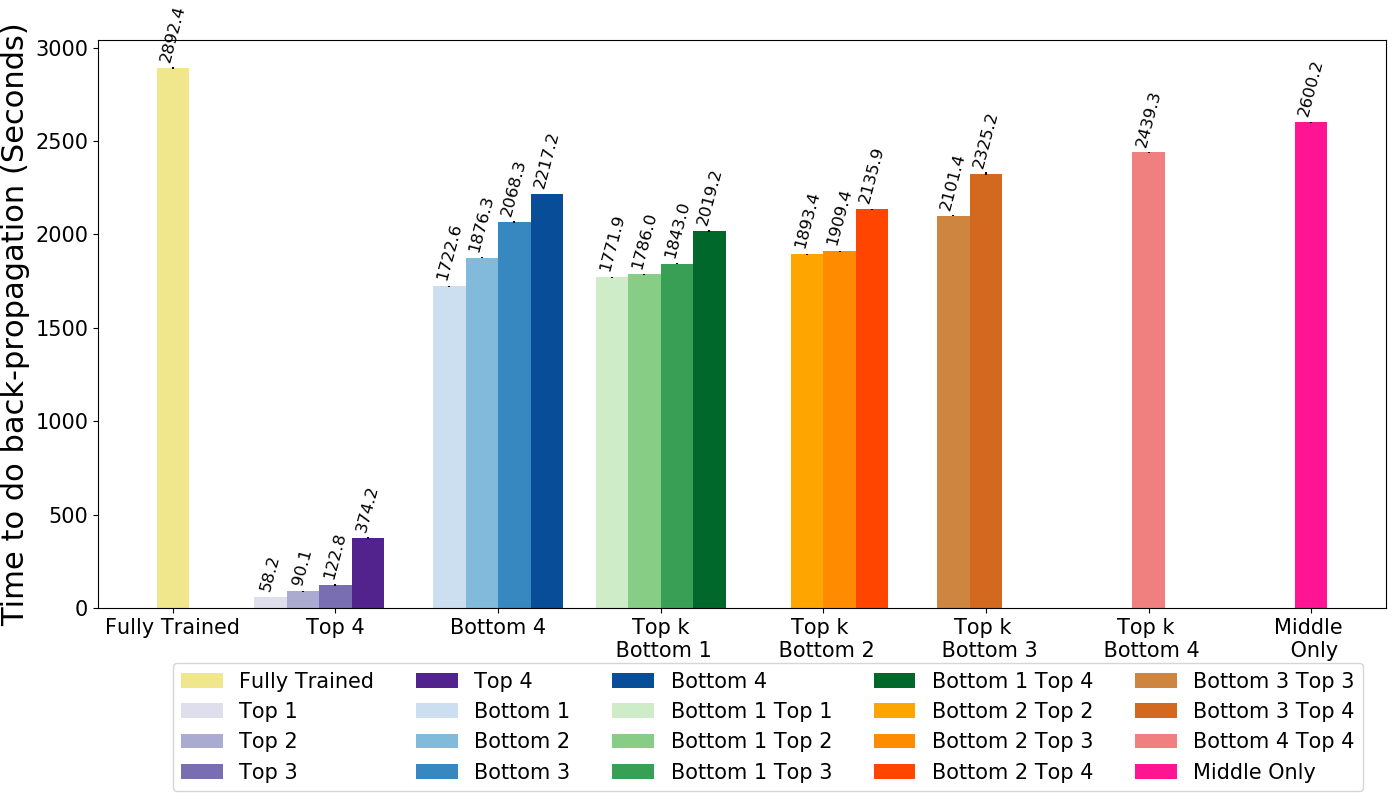}}
    \caption{Test accuracy ($\%$) and the corresponding time (in seconds) to perform back-propagation for VGG-11 trained CIFAR-10 using the following LWS-SGD variants: (a) the top $k$ layer(s) only, (b) the bottom $q$ layer(s) only, (c) the combination of both top and bottom layer(s), and (d) the middle layer(s) only. The parameters in other layers are frozen during the training. The bar shows the average test accuracy at convergence, computed over 5 repetitive runs, and the error bar shows the corresponding one standard error. In general, training an appropriate combination of top $k$ and bottom $q$ layers works the best with little or no adversarial effect on generalization.}
     \label{fig:vgg11_cifar10_retrain}
\end{figure*}

\begin{figure*}[t]
    \centering
    \subfigure[VGG-11, CIFAR-10.]{
    \includegraphics[width=0.97\textwidth]{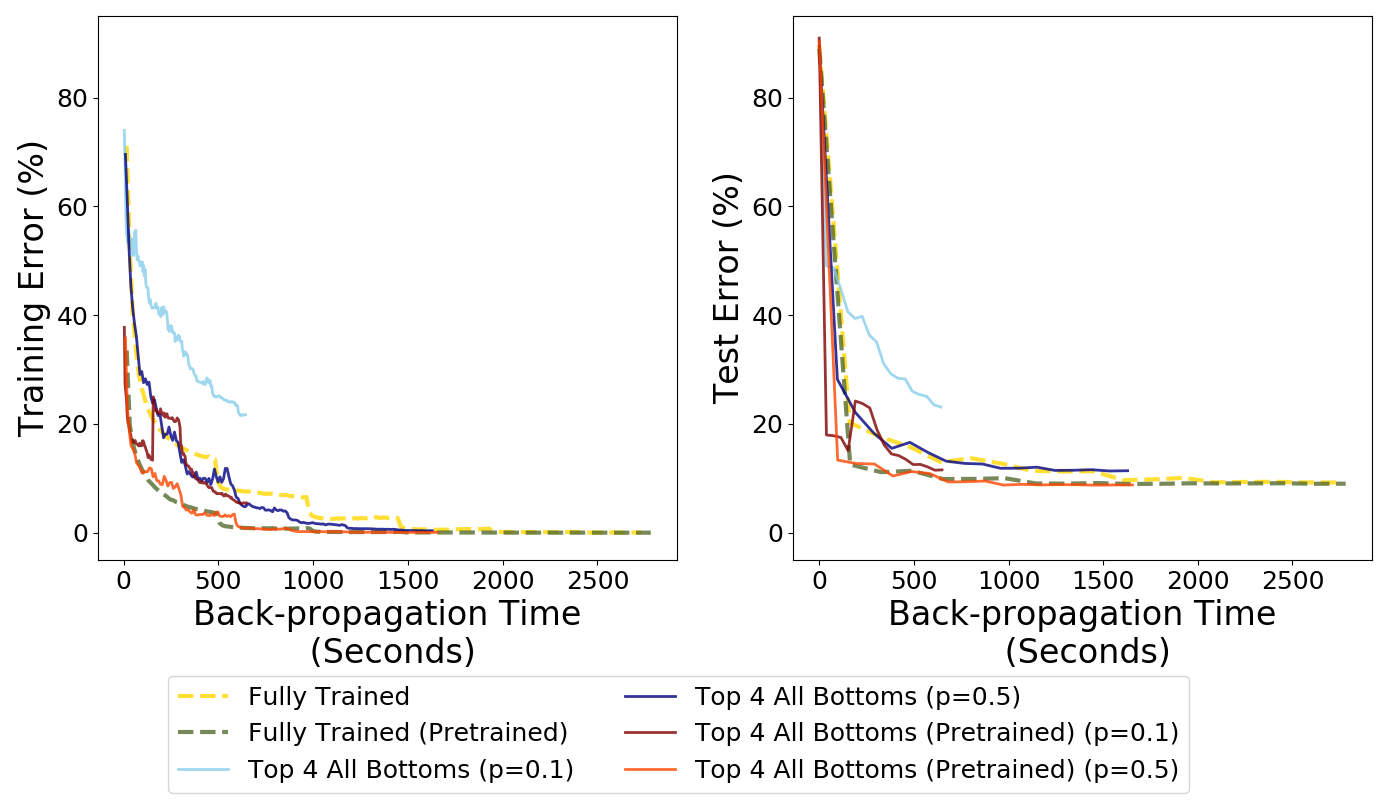}}
    \caption{Dynamics of training and test Error(\%) as a function of actual time to perform back-propagation (in seconds) for VGG-11 trained on CIFAR-10. We compare performance among fully trained model and the model trained using the best-performed probabilistic LWS-SGD variant, i.e., Top $4$ All Bottoms. VGG-11 is initialized using both Xavier random initialization and pre-trained values learnt from ImageNet. Networks initialized with pre-trained weights not only have better generalization but also converge fast.}
    \label{fig:vgg11_cifar_train_test_err_dym}
\end{figure*}

In this section, we first show additional results of $\gamma$ Active-re-initialization and $\varepsilon$ Lazy-re-initialization performed on Fashion-MNIST dataset \citep{fashionmnist} 
(Figures~\ref{fig:relu_fashion_reinit}-\ref{fig:relu_fashion_init_vs_final}). Observations we made on Fashion-MNIST are consistent with what has been observed on MNIST. Next, we present layer-wise distribution of active(lazy) parameters of (1) 5-Layer Conv-Nets ($d=4$) trained on MNIST (Figure~\ref{fig:vgg4_mnist_active}); (2) 5-Layer ReLU-Nets ($d=4$) trained on Fashion-MNIST (Figure~\ref{fig:relu_fashion_active}); and (3) Conv-Nets (both $d=2$ and $d=4$) trained on Fashion-MNIST (Figure~\ref{fig:vgg2_fashion_active} and \ref{fig:vgg4_fashion_active}). Additional results on both MNIST and Fashion-MNIST with ReLU-Nets and Conv-Nets with various depth and width confirm that, as width increase, active parameters lean toward concentration at bottom. Finally, we provide experimental results of both static and probabilistic LWS-SGD with a broader choice of $k$ and $q$ (Figures~\ref{fig:vgg11_cifar10_retrain}-\ref{fig:vgg11_cifar_train_test_err_dym} and Table~\ref{tab:vgg11_cifar_adaptive_acc_and_time}).

\begin{figure*}[h]
    \centering
    \subfigure[VGG-5.]{
    \includegraphics[width=0.25\textwidth]{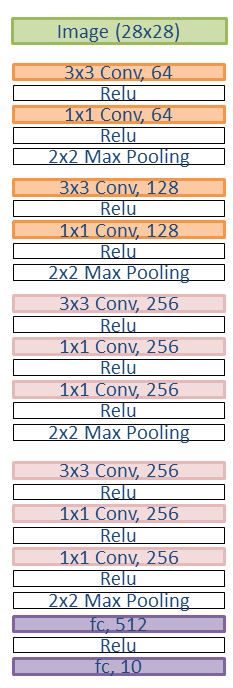}}
    \hspace{25mm}
    \subfigure[VGG-11.]{
    \includegraphics[width=0.25\textwidth]{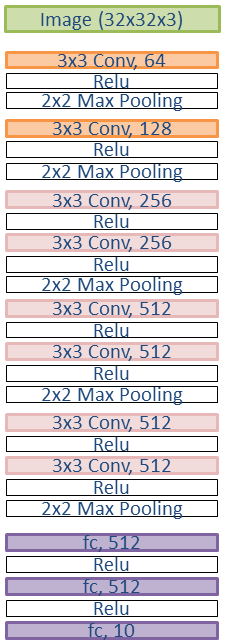}}    
    \caption{The architecture of (a) VGG-5 used for MNIST and (b) VGG-11 used for CIFAR-10. Both the batch normalization and dropout layer has been excluded, since our focuses are purely on the convolutional layer and the fully-connected layer.}
     \label{fig:vgg5}
\end{figure*}

\clearpage

\begin{figure*}[t]
    \centering
    \subfigure[VGG-5, MNIST.]{
    \includegraphics[width=0.97\textwidth]{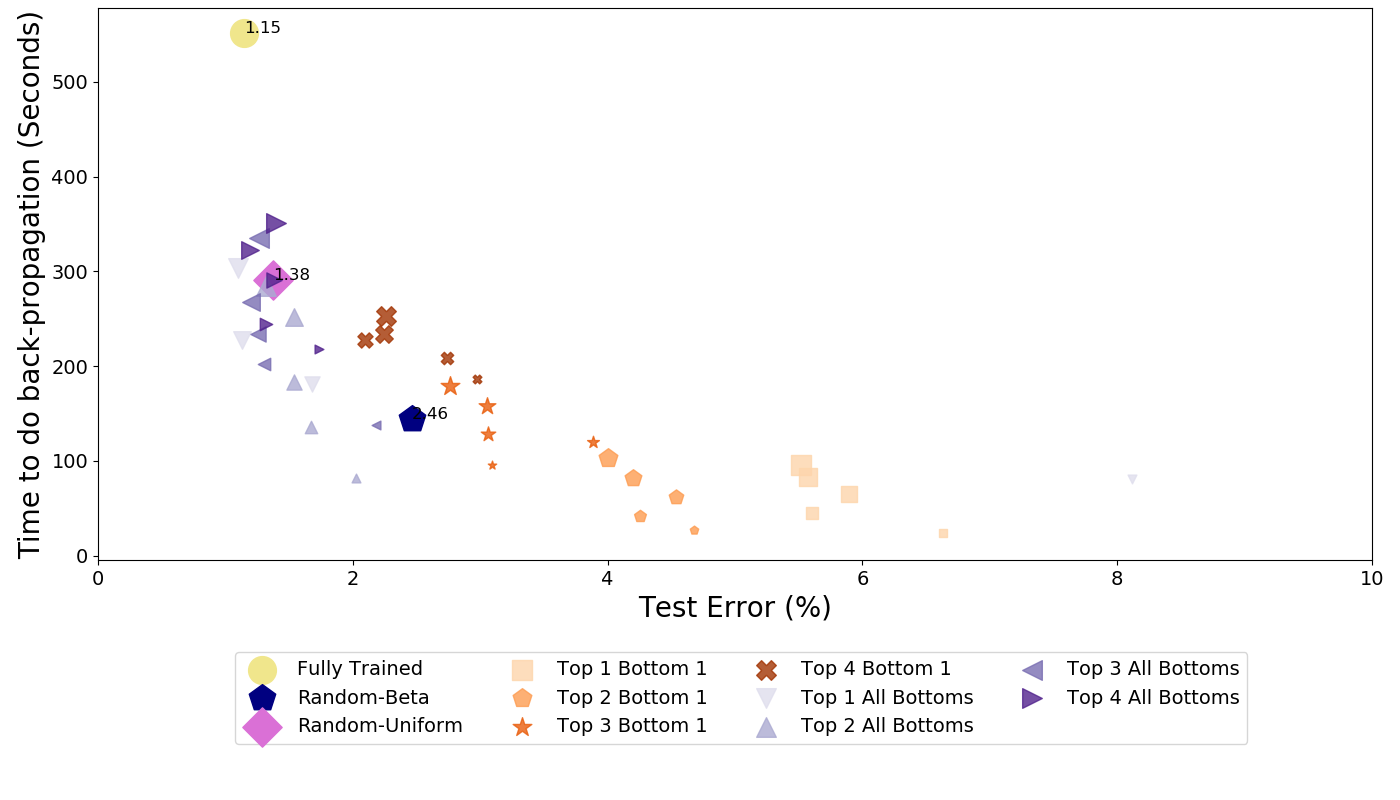}}
    \subfigure[VGG-11, CIFAR-10.]{
    \includegraphics[width=0.97\textwidth]{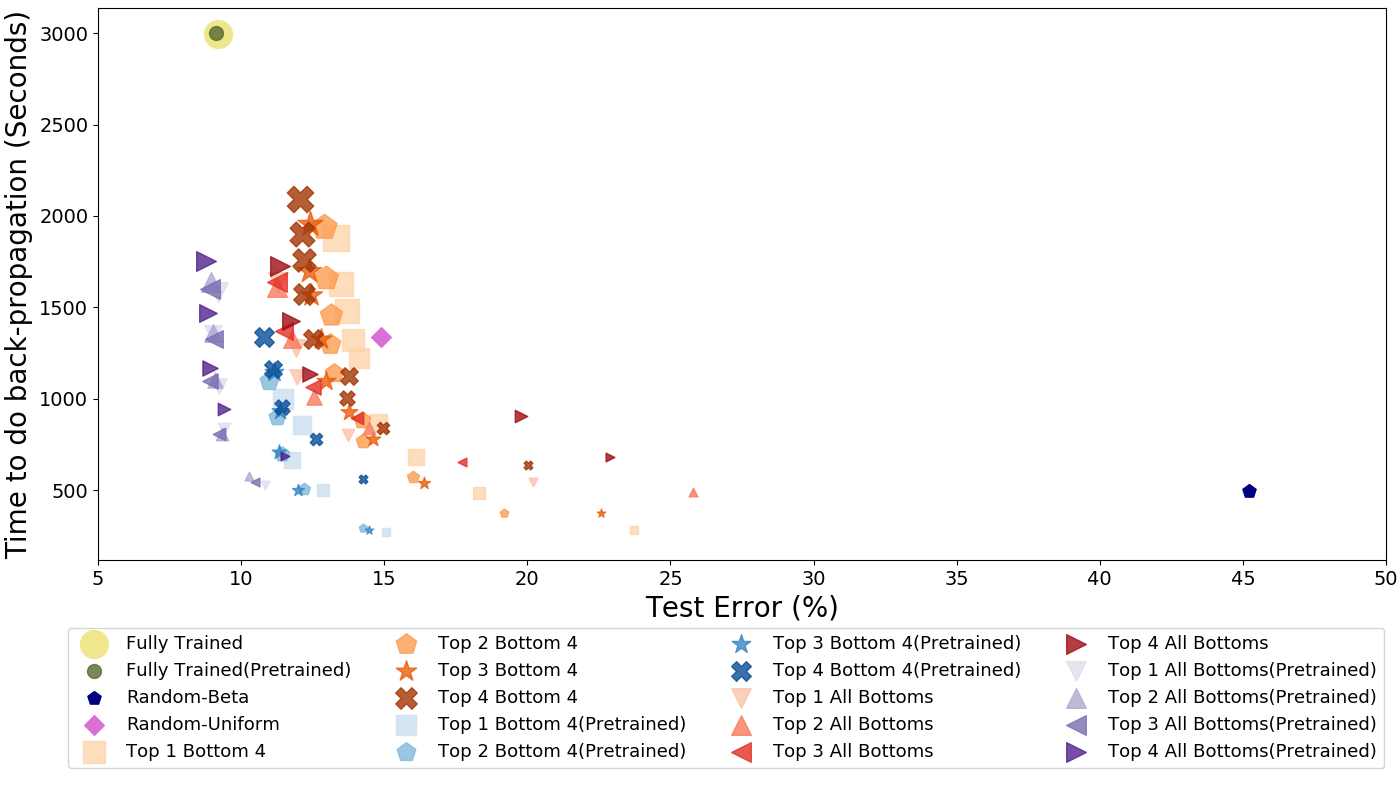}}
    \caption{ Test Error($\%$) versus Time (in seconds) to perform back-propagation for networks trained by the following probabilistic LWS SGD variants: (a) Top $k$ Bottom $q$($q=1$ for MNIST and $q=4$ for CIFAR-10), (b) Top $k$ All Bottoms, (c) Random-Uniform, and (d) Random-Beta. VGG-5 is initialized using random values and VGG-11 is initialized use both random values and pre-trained values learnt from ImageNet. The size of the marker indicates the probability $\rho\in [0.1,0.2,0.3,0.4,0.5]$ of accessing the selected bottom layers, the larger the marker is, the more frequent it updates the bottom layers. Top $k$ All Bottoms (VGG-5 with random initialization and VGG-11 with pre-trained weights) only needs a small $\rho$ value (e.g., $\rho=0.1$) to achieve similar generalization performance of the fully trained model, yet it approximately 2 to 5 times faster than doing back-propagation on a full model.}
\end{figure*}

\clearpage

\begin{table*}[h]
  \caption{Test accuracy ($\%$) and the corresponding back-propagation time (seconds) of VGG-11 trained on CIFAR-10. We compare the performance among vanilla SGD and probabilistic LWS-SGD variants, including (a) Random-Uniform, (b)Random-Betta, (c) Top $k$ Bottom $q$, and (d) Top $k$ All Bottoms. For demonstration purpose, we only show the results for $k\in [1,4]$, $q=4$ and $\rho \in [0.1,0.2,0.3,0.4,0.5]$. The parentheses behind test accuracy show the performance difference ($\%$) compared with fully trained model (vanilla SGD). The positive number indicates an improvement while the negative number means the performance has been deteriorated. The parentheses behind back-propagation time includes the ratio between the time spend for current method and the time used to train the full model. The smaller the number is the more efficient the training algorithm is.}
 \label{tab:vgg11_cifar_adaptive_acc_and_time}
\centering
\resizebox{0.98\textwidth}{!}{
\begin{tabular}{c|c|c|c|c|c}
\hline
 & \textbf{Fully trained} & \textbf{Fully trained (pre-train)} & \textbf{Random-Uniform} & \textbf{Random-Beta} &  \\
 \hline
 Test accuracy (Difference)  & 90.80(0.00) & 90.89(+0.09) & 85.10(-5.70)& 54.80(-36.01) & \\
 Time (Ratio) & 2995.86(1) & 3000.17($\approx 1$)& 1336.77(0.45) &493.75(0.16)& \\
 \hline
 \hline
  $\rho$ & 0.1 & 0.2& 0.3 & 0.4& 0.5 \\
  \hline 
 & \multicolumn{5}{c}{\textbf{Top 1 Bottom 4}}\\
 \hline
 Test accuracy (Difference)  & 76.28(-14.53) & 81.68(-9.12) & 83.91(-6.90)& 85.20(-5.60) & 85.87(-4.94)\\
 Time (Ratio) & 283.42(0.09) & 484.14(0.16) & 681.99(0.23) & 864.28(0.29) & 1221.86(0.41)\\
 \hline
 & \multicolumn{5}{c}{\textbf{Top 1 Bottom 4 (pre-train)}}\\
 \hline
 Test accuracy (Difference)  & 84.93(-5.87) & 87.14(-3.66) & 88.23(-2.57)& 87.87(-2.94) & 88.52(-2.28)\\
 Time (Ratio) & 269.95(0.09) & 501.89($0.17$) & 666.70($0.22$) & 856.71($0.29$) & 996.81($0.33$)\\
 \hline
 & \multicolumn{5}{c}{\textbf{Top 4 Bottom 4}}\\
 \hline
 Test accuracy (Difference)  & 79.98(-10.83) & 85.04(-5.77) & 86.31(-4.50)& 86.23(-4.58) & 87.48(-3.32)\\
 Time (Ratio) & 634.58 (0.21)& 836.79(0.28) & 1002.64(0.33) & 1126.26(0.38) &1324.76(0.44)\\
 \hline
 & \multicolumn{5}{c}{\textbf{Top 4 Bottom 4 (pre-train)}}\\
 \hline
 Test accuracy (Difference)  & 85.73(-5.08) & 87.38(-3.42) & 88.57(-2.24)& 88.90(-1.91) & 89.19(-1.61)\\
 Time (Ratio) & 560.98($0.19$)& 778.76($0.26$)&953.81($0.32$)& 1162.25($0.39$)& 1336.65($0.45$)\\
 \hline
 & \multicolumn{5}{c}{\textbf{Top 1 All Bottoms}}\\
 \hline
 Test accuracy (Difference)  & 79.79(-11.01) & 86.27(-4.53) & 88.04(-2.76)& 88.08(-2.72) & 88.71(-2.10)\\
 Time (Ratio) & 541.73(0.18) & 802.80(0.27) & 1120.82(0.37) & 1276.80(0.43) & 1636.31(0.55)\\
 \hline
 &\multicolumn{5}{c}{\textbf{Top 1 All Bottoms (pre-train)}}\\
 \hline
 Test accuracy (Difference) & 89.17(-1.64) &90.59(-0.22)&90.77(-0.03)&90.98(+0.17)&90.82(+0.02)\\
 Time (Ratio) & 528.19(0.18) & 835.34(0.28) & 1070.72(0.36) & 1355.49(0.45) & 1585.46(0.53)\\
 \hline
  & \multicolumn{5}{c}{\textbf{Top 4 All Bottoms}}\\
 \hline
 Test accuracy (Difference)  & 77.12(-13.69) & 80.23(-10.58) & 87.60(-3.21)& 88.27(-2.53) & 88.65(-2.15)\\
 Time (Ratio) & 681.12(0.23)&904.54(0.30)&1134.99(0.38)&1424.47(0.48)&1725.12(0.58)\\
 \hline
  & \multicolumn{5}{c}{\textbf{Top 4 All Bottoms (pre-train)}}\\
 \hline
 Test accuracy (Difference)  & 88.47(-2.34) & 90.59(-0.22) & 91.08(+0.28)& 91.17(+0.36) & 91.23(+0.42)\\
 Time (Ratio) & 687.06(0.23)&945.95(0.32)&1166.41(0.39)&1470.47(0.49)&1752.37(0.58)\\
 \hline
\end{tabular}
}
\end{table*}

\end{document}